\pdfoutput=1

\documentclass[11pt]{article}

\usepackage[preprint]{acl}

\usepackage{times}
\usepackage{latexsym}
\usepackage{amssymb}
\usepackage{amsmath}
\usepackage{subfigure}
\usepackage{booktabs}

\usepackage[T1]{fontenc}

\usepackage[utf8]{inputenc}

\usepackage{microtype}

\usepackage{inconsolata}

\usepackage{multirow}
\usepackage{diagbox}
\usepackage{makecell}
\usepackage{graphicx}
\usepackage{enumitem}
\usepackage[toc,page,header]{appendix}
\usepackage{minitoc}
\setlist{leftmargin=*}

\title{TARDIS: Mitigating Temporal Misalignment via Representation Steering}

\author{%
  \textbf{Changho Shin, Xinya Yan, Suenggwan Jo, Sungjun Cho, Shourjo Aditya Chaudhuri, Frederic Sala}\\
  Department of Computer Sciences \\
  University of Wisconsin-Madison \\
  \texttt{\{cshin23, xyan89, sjo32, cho266, sachaudhuri, fsala\}@wisc.edu}
}

\newcommand{\SYSNAME}{\textsc{TARDIS}}
\appto\appendix{\addtocontents{toc}{\protect\setcounter{tocdepth}{0}}}
\begin{document}
\doparttoc
\maketitle
\begin{abstract}
Language models often struggle with \textit{temporal misalignment}, performance degradation caused by shifts in the temporal distribution of data. Continuously updating models to avoid degradation is expensive. Can models be \emph{adapted} without updating model weights? We present $\SYSNAME$, an unsupervised representation editing method that addresses this challenge. $\SYSNAME$ extracts steering vectors from unlabeled data and adjusts the model's representations to better align with the target time period's distribution. Our experiments reveal that $\SYSNAME$ enhances downstream task performance without the need for fine-tuning, can mitigate temporal misalignment even when exact target time period data is unavailable, and remains efficient even when the temporal information of the target data points is unknown at inference time. 
\end{abstract}

\section{Introduction}
Language distributions continuously change, tracking the fact that people use language to represent ever-changing worlds. However, trained language models only capture the distribution of their training data. This results in \emph{temporal misalignment}--- performance degradation of language models induced by the temporal difference between training and test time periods. 

Existing methods attempting to resolve temporal misalignment require fine-tuning \cite{rottger2021temporal, luu2022time, nylund2023time}  or training in a time-aware way from the beginning \cite{son2023time,zhao2024set}.  However, collecting data with time annotations for the target time period or performing time-aware training is costly. Additionally, large model sizes imply that the computational requirements for fine-tuning are also non-trivial. 

Can we address temporal misalignment \emph{without additional training or time-specific data} for a target time period? While seemingly challenging, the distribution shifts in languages are often already captured by pretrained model representations. We exploit this notion to produce $\SYSNAME$ (Time Alignment via Representation DIstribution Shifting), an efficient unsupervised method. 

\begin{figure}[!t]
	\centering
\includegraphics[width=0.48\textwidth]{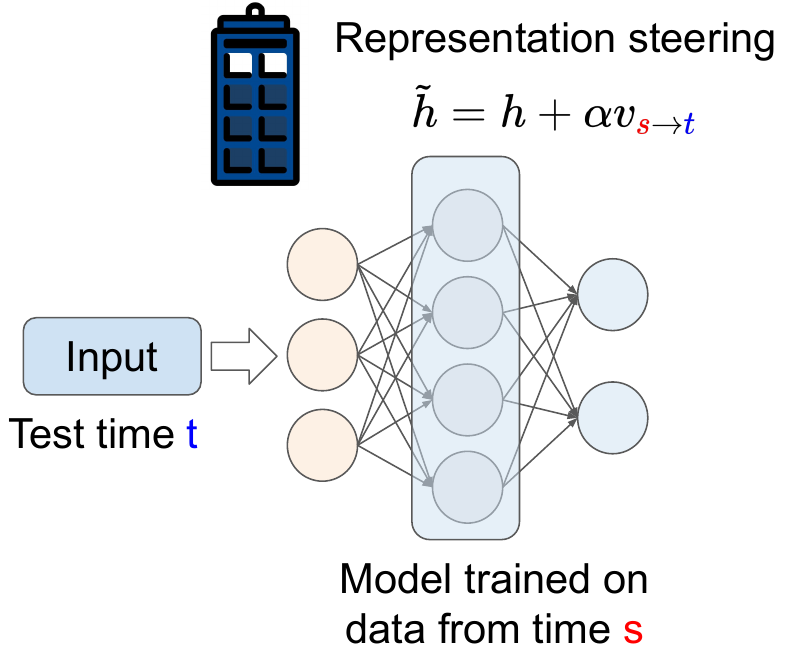}
\caption{TARDIS steers model representations to mitigate degradation caused by temporal misalignment.}
    \label{fig:framework}
\end{figure}

$\SYSNAME$ hinges on representation editing methods \cite{liu2023context, subramani2022extracting,tigges2023linear,turner2023activation,zou2023representation, li2024inference, lee2024mechanistic,uppaal2024detox, adila2024free}. These techniques have been used to reduce toxicity \cite{lee2024mechanistic,uppaal2024detox} and bias \cite{adila2024discovering}, improve trustfulness \cite{li2024inference}, and align models with human preference in general \cite{luo2024pace, adila2024free}. 

Representation editing methods typically use two steps. First, they extract \textbf{steering vectors} that represent desirable/undesirable directions in the hidden layer representation space. Next, during inference, they steer representations into desirable directions (or away from undesirable directions). In our setting, we obtain the steering vectors by subtracting the average representation in the training time period from the average representation in the test time period. The steering vectors are simply added to the target data representations.

Our experiments show that $\SYSNAME$ can mitigate temporal misalignment significantly, improving accuracy up to 19.2\%. Additionally, we investigate whether we can perform time vector arithmetic, showing that there is temporal structure in the representation spaces of LLMs. Finally, we examine a dynamic adaptation method based on time period estimation.


\section{Background} 
\paragraph{Temporal Misalignment.} Temporal changes in language have been widely reported \cite{altmann2009beyond, eisenstein2014diffusion}. Language model degradation as an outcome of such changes implies that temporal alignment is an important problem. To address temporal adaptation, many fine-tuning or time-aware training methods have been suggested. We highlight several. 

\citet{rottger2021temporal} shows that  temporal fine-tuning effectively mitigates temporal misalignments. \citet{luu2022time} analyzes temporal alignments in various tasks, revealing that label shift and vocabulary shift are crucial components of temporal misalignment. \citet{nylund2023time} proposes \emph{time vectors}---differences between the fine-tuned model on the target time period and the base model. Task vector arithmetic \cite{ortiz2024task} can be applied in the context of temporal misalignment. However, these methods require model weight updates and run  the risk of degradation in non-target time periods. In contrast, $\SYSNAME$ provides a lightweight way to adapt models without fine-tuning.

\paragraph{Representation Editing.} Representation editing seeks to modify model behavior without fine-tuning, mainly by transforming representations instead of updating weights. It has been widely applied for various purposes. 
\citet{li2024inference} seeks to improve trustfulness via linear probing on the embeddings. \citet{lee2024mechanistic,uppaal2024detox} show  that transforming representations can efficiently detoxify language model outputs at inference time. \citet{luo2024pace} constructs concept dictionaries  and aligns models with relevant concepts from these at  inference time. \citet{adila2024free} uses self-generated synthetic data to extract steering vectors for human alignment, mitigating the need for a dataset to find steering vectors.
These works, however, do not address temporal misalignment.
\section{TARDIS: Fixing Temporal Misalignment with Representation Steering}\label{sec:method}
$\SYSNAME$ steers representations to the target time period by intervening on them duing inference time via a two-step procedure: 

\paragraph{Extracting steering vectors.} Let the training time period dataset be $D_{s} = \{x_1^s, \ldots, x_{n_s}^s\}$ and the target time period test dataset be $D_{t} = \{x_1^t, \ldots, x_{n_t}^t\}$. Note that the training time period dataset \emph{does not need to be} the training dataset itself. Denote the corresponding $l$-th MLP layer outputs as $H_s^l=\{h^{s,l}_1, \ldots, h^{s,l}_{n_s}\}$ and $H_t^l=\{h^{t,l}_1, \ldots, h^{t,l}_{n_t}\}$. Then, we define the steering vectors for each $l$-th layer as
\[ v^l_{s\rightarrow t}=\frac{1}{n_t}\sum_{i=1}^{n_t}h^{t,l}_i - \frac{1}{n_s}\sum_{i=1}^{n_s}h^{s,l}_i.\]

\paragraph{Steering representations to the target time.} At inference time, let $h^l$ be the $l$-th MLP layer output when forwarding a data point from the target time period $t$. We transform $h^l$ by
\[\tilde{h^l}=h^l + \alpha v^l_{s\rightarrow t},\] where $\alpha \in \mathbb{R}$ is a hyperparameter that controls the strength of alignment. 

In practice, we only intervene on the last encoder- and last decoder-layer for encoder-decoder architecture models \cite{raffel2020t5}, or on the last 3 decoder-layers for decoder-only architecture models \cite{radford2019gpt2}.

\begin{figure*}[t!]
	\centering
	\subfigure [AIC]{
\includegraphics[width=0.33\textwidth]{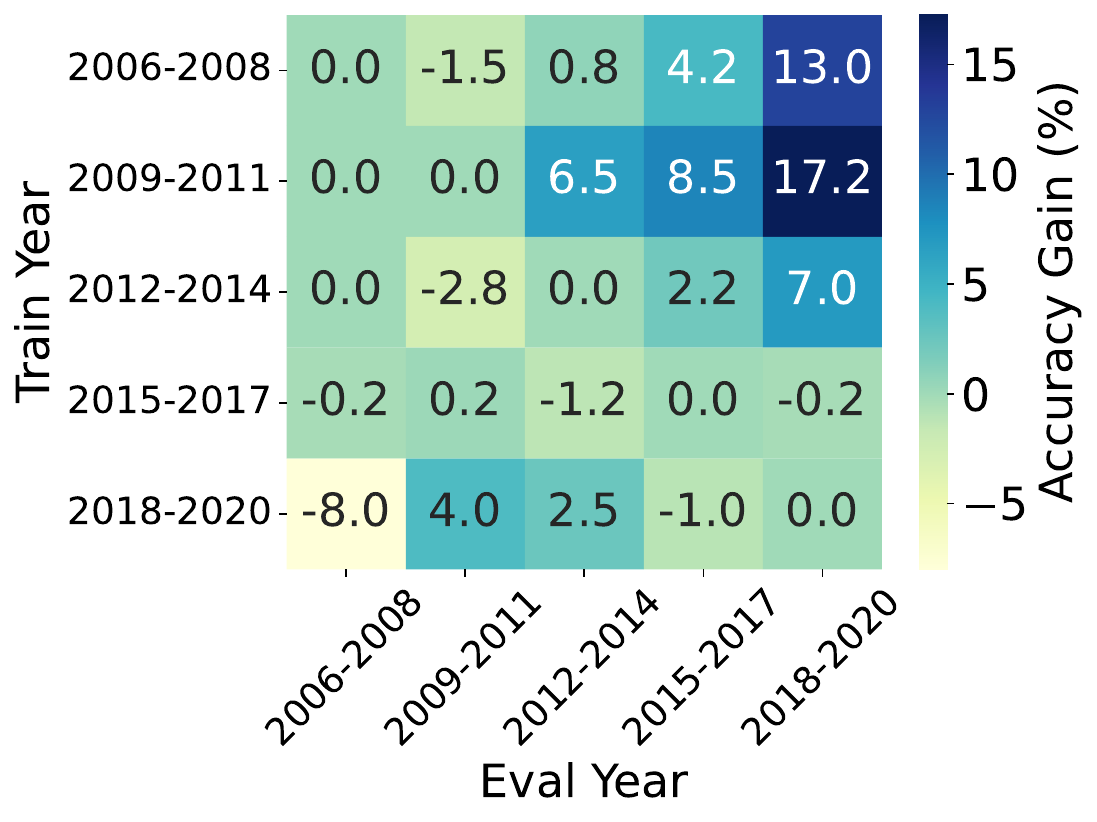}}
\subfigure [PoliAff]{
\includegraphics[width=0.33\textwidth]{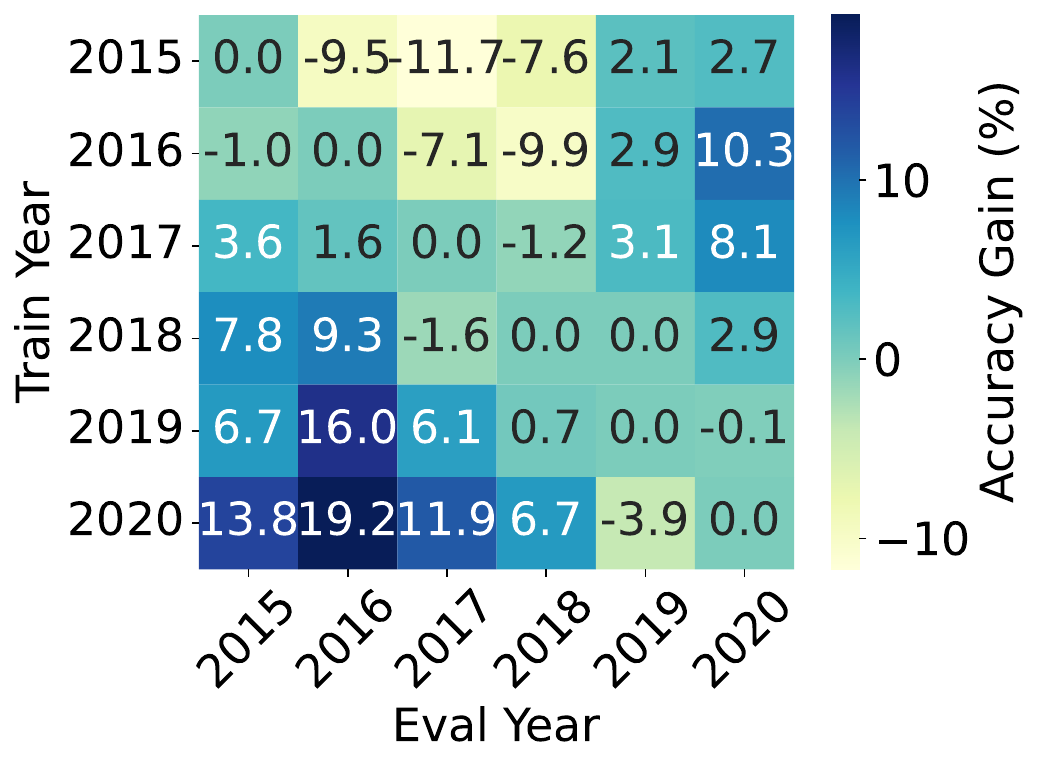}}
\subfigure [NewsCls]{
\includegraphics[width=0.31\textwidth]{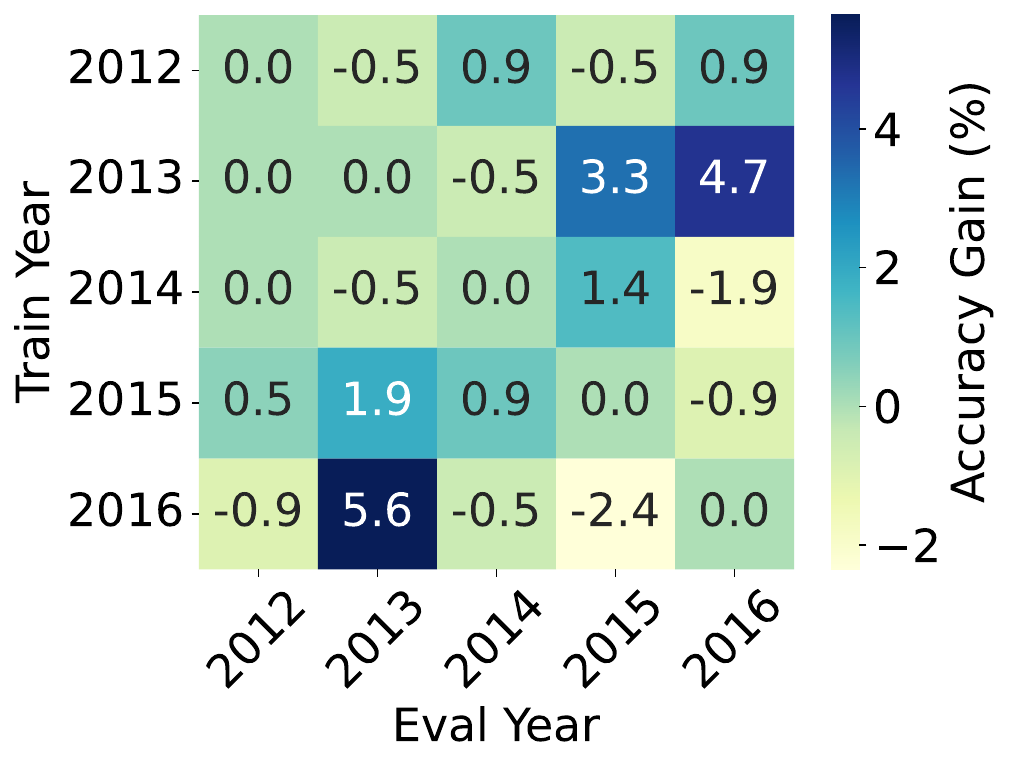}}

\caption{Performance gains when using $\SYSNAME$ . We observe that $\SYSNAME$ can improve accuracy up to 19.2\% without any fine-tuning.}
    \label{fig:steer_results}
\end{figure*}

\section{Experiments}
We evaluate the following claims about $\SYSNAME$.
\begin{itemize}
    \item \textbf{Mitigates temporal misalignment}: $\SYSNAME$ effectively mitigates degradation due to temporal distribution shifts.
    \item \textbf{Interpolating/Extrapolating steering vectors}: $\SYSNAME$ allows interpolation/extrapolation of steering vectors, reducing the need for the target period data.
    \item \textbf{Dynamic $\SYSNAME$ via time period estimation}: $\SYSNAME$ can be used even when the time period of the target data is \emph{unknown} by using time period estimation.
\end{itemize}

\paragraph{Datasets.} We use three classification datasets from \citet{luu2022time}: AIC (AI venue classification), PoliAff (Political affiliation classification), and NewsCls (Newsroom publisher classification). For dataset details, refer to Appendix \ref{appsub:dataset}.
    

    

Additionally, in Appendix \ref{app:additional_exp}, we provide further experiments on using the SVD approach for steering instead of the mean difference, along with an ablation study on the intervention layer and the number of samples.

\paragraph{Models.}  We use T5-60m \cite{raffel2020t5} fine-tuned on each year's data as in \citet{nylund2023time}\footnote{We use finetuned models from \href{https://huggingface.co/collections/KaiNylund/time-vector-models-66132ae0b5e6420d0380da7f}{https://huggingface.co/collections/KaiNylund/time-vector-models-66132ae0b5e6420d0380da7f} licensed under the MIT License.}. To show that conclusion does not change substantially based on the model size and model architecture, we include partial results with T5-770m and GPT2 in the Appendix.

\subsection{Mitigating Degradation by Temporal Misalignment}\label{subsec:exp1}
We first empirically evaluate how effective $\SYSNAME$ is in reducing performance degradation due to temporal misalignment.

\paragraph{Setup.} Given a model $\mathcal{M}_{s}$ fine-tuned on data points from time point $s$, we extract the steering vectors $v^l_{s \rightarrow t}$ for each year $t$ following the procedure in Section \ref{sec:method}. We choose the alignment weight $\alpha$ based on the validation with grid search on $[-5, -3, -2, -1, 1, 2, 3, 5]$. The selected $\alpha$'s are 1, 3, and -2 for AIC, PoliAff, and NewsCls, respectively.

\paragraph{Results.} Figure \ref{fig:steer_results} presents the performance gain when using $\SYSNAME$ over the yearly fine-tuned data. $\SYSNAME$ improves accuracy up to 19.2\% without any fine-tuning. $\SYSNAME$ tends to be more effective when the training year and the test year has a large gap. In Appendix \ref{appsec:exp1}, we provide further analysis and additional experiment results.

\subsection{Understanding Mechanism behind $\SYSNAME$}\label{subsec:exp2}
Figure \ref{fig:steer_results} brings up two questions: 1) Why is $\alpha>0$ beneficial in AIC, PoliAff while $\alpha<0$ yields better results in NewsCls? 2) What is the mechanism behind the accuracy improvements? 
\citet{luu2022time} identified two causes of temporal shifts: label shift and vocabulary shift, namely distribution shifts in label and input vocabulary. We hypothesize that steering to the target time period ($\alpha > 0$) reduces label shift, while steering to the training time period ($\alpha < 0$) mitigates the effect of vocabulary shift. Thus, if label shift is more significant than vocabulary shift (or semantic shift), $\alpha > 0$ is expected to be effective, and vice versa.

\paragraph{Setup.} We separate label shift and vocabulary shift. To simulate label shift only, we control the label distribution of the test data which is sampled in the training time period. We apply $\SYSNAME$ to the sampled data from the original data. Similarly, to simulate vocabulary shift, we keep the label distribution at test time to the same level of the training year by sampling, varying the test time period.

\begin{figure}[t!]
	\centering
	\subfigure [Label Shift]{
\includegraphics[width=0.225\textwidth]{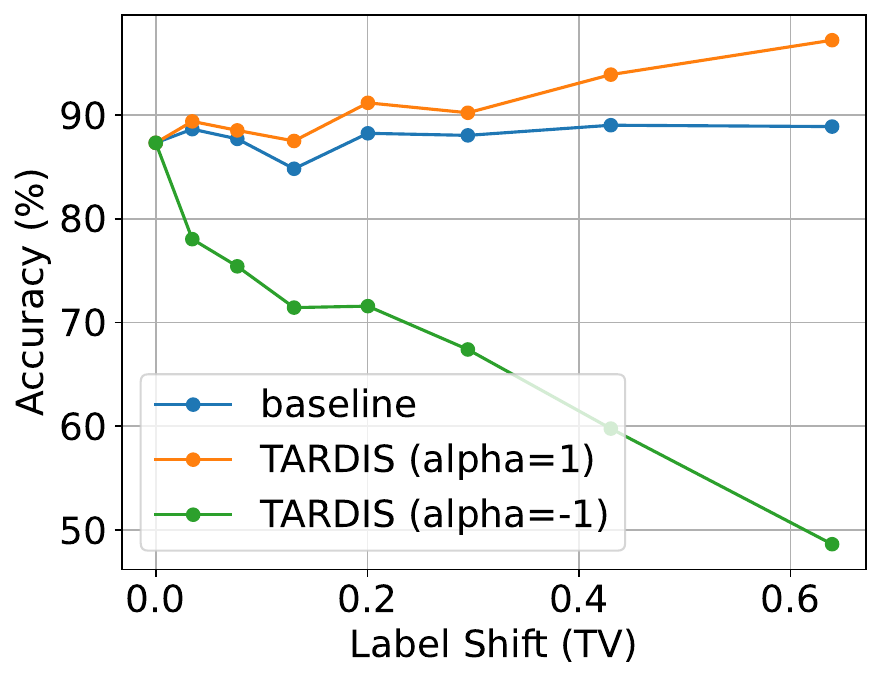}} 
\subfigure [Vocabulary Shift]{
\includegraphics[width=0.225\textwidth]{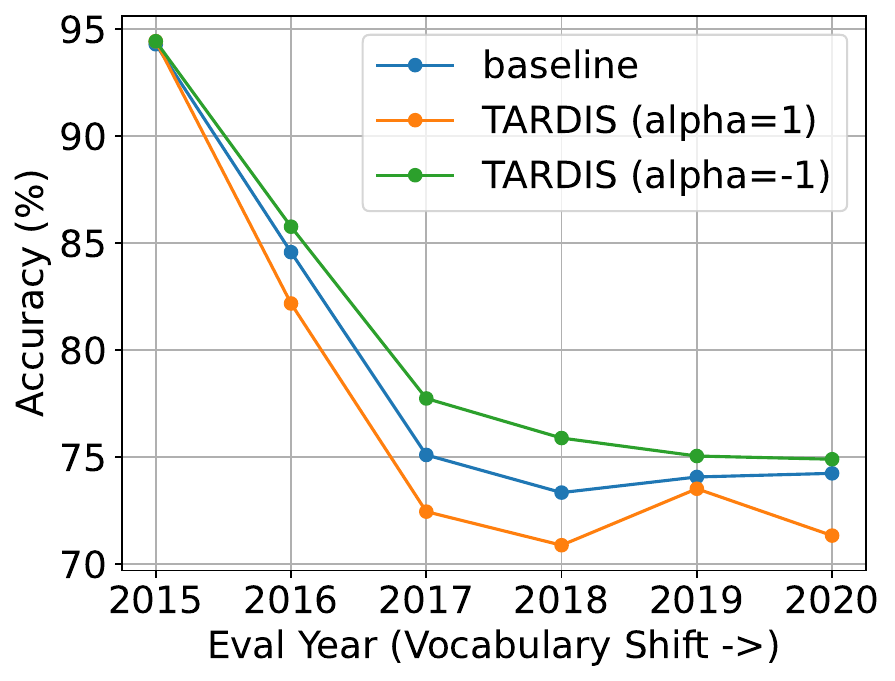}}
\caption{Semi-synthetic label shift experiment result with NewsCls task (a) and semi-synthetic vocabulary shift experiment result with PoliAff task (b). $\SYSNAME$ with $\alpha > 0$ yields more accuracy improvement as the level of the label shift increases, while $\SYSNAME$ with $\alpha < 0$ is effective to mitigate degradation due to  vocabulary shift. }
    \label{fig:semisynthetic}
\end{figure}

\paragraph{Results.} Figure \ref{fig:semisynthetic} (a) shows that $\SYSNAME$ with $\alpha>0$ can address label distribution shift effectively, by pushing representations to the target distribution. 
Figure \ref{fig:semisynthetic} (b) shows that $\SYSNAME$ with $\alpha<0$ can mitigate degradation from vocabulary shift by steering semantic information in input representations toward the training time distribution. 

As we anticipated, these results imply that if label shift is more significant than vocabulary shift, we can expect steering to the target year ($\alpha$ > 0) to be effective, and if vocabulary shift is more significant than label shift, then  steering to the training year ($\alpha$ < 0) is likely to be beneficial. We include the full results in Appendix \ref{appsec:exp2}

\subsection{Getting on the Timeline: Interpolating / Extrapolating Steering Vectors}\label{subsec:exp3}
\citet{nylund2023time} showed task arithmetic \cite{ortiz2024task} can be applied to time weight vectors, revealing that interpolating two weight vectors for two different time periods can be be used to adapt the model to an intermediate period. We hypothesize that steering vectors can also produce timelines, thus enabling interpolation and extrapolation by composing a steering vector in the middle of two time periods.

\paragraph{Setup.} Suppose we have data for the time periods $\{t, t+1, \ldots, t+d\}$. For the forward direction, interpolated steering vectors are formulated as $\hat{v}^l_{t \rightarrow t+j}=\frac{j}{d}v^l_{t \rightarrow t+d}$. Extrapolated steering vectors are composed as $\hat{v}^l_{t \rightarrow t+j}=j v^l_{t \rightarrow t+1}$. Steering vectors for backward direction are obtained similarly.

\begin{figure}[t!]
	\centering
	\subfigure [Forward (AIC)]{
\includegraphics[width=0.225\textwidth]{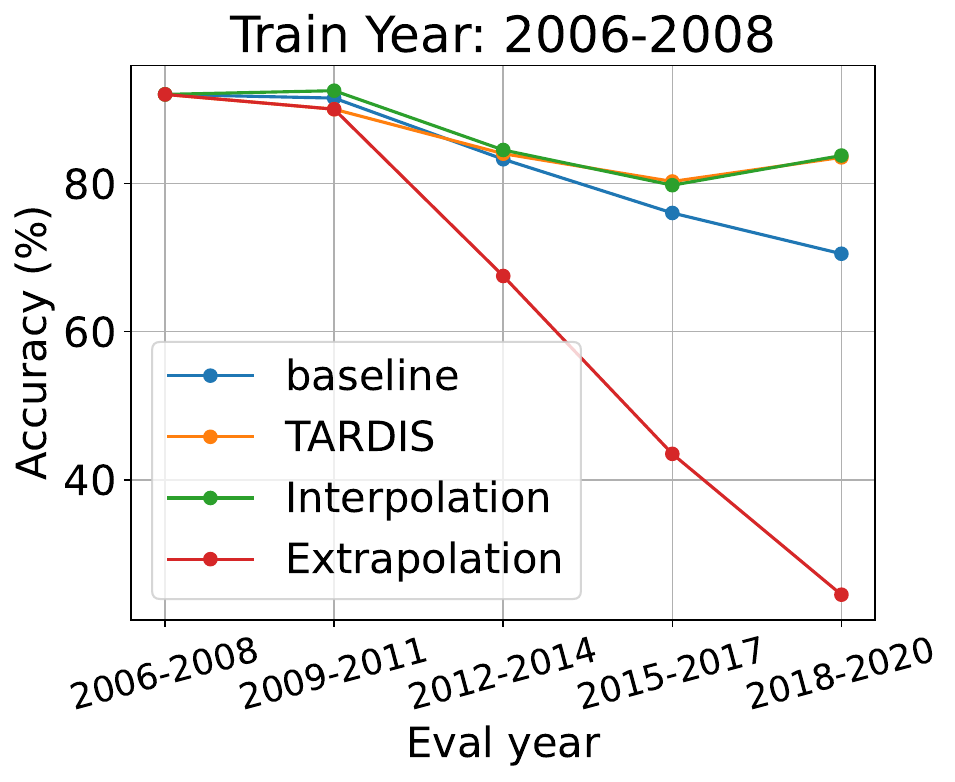}}
\subfigure [Backward (PoliAff)]{
\includegraphics[width=0.225\textwidth]{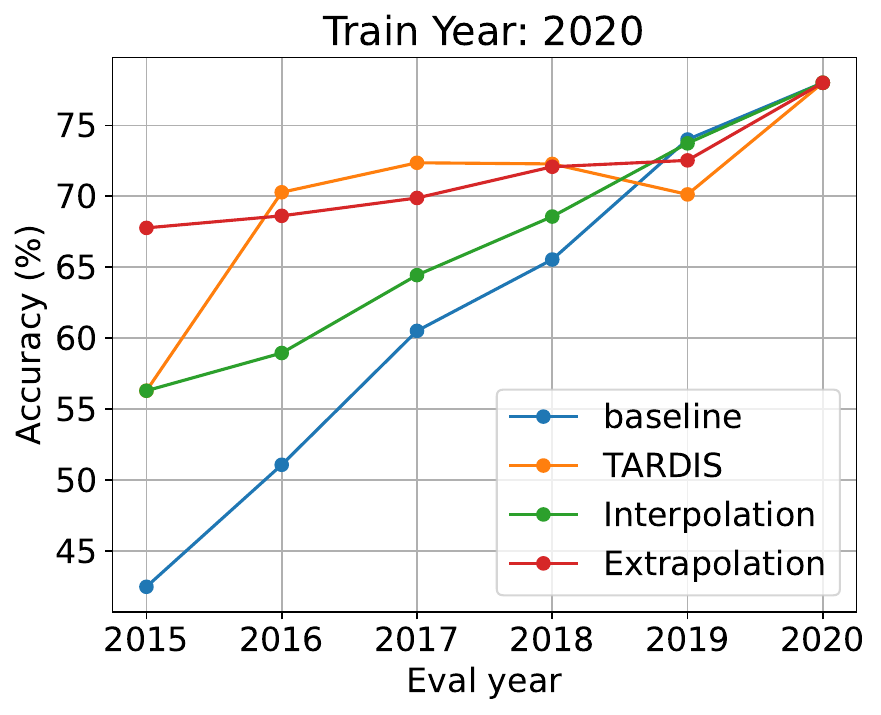}}
\caption{Interpolation/extrapolation of steering vectors. $\SYSNAME$ uses steering vectors taken at the target time period. We see that interpolated/extrapolated steering vectors can mitigate degradation similarly to the exact time steering vectors.}
    \label{fig:time_on_the_line}
\end{figure}

\paragraph{Results.} Figure \ref{fig:time_on_the_line} presents our results. We observe that interpolated steering vectors can mitigate the effect of temporal misalignment with in-between enhancement, while the extrapolation steering vectors occasionally collapse but sometimes yield even better results than the exact steering vectors. We include the full results in Appendix \ref{appsec:exp3}.

\subsection{Dynamic Steering via Target Time Period Estimation}\label{subsec:exp4}
Even when the steering vectors are given, the target time period might not be known. We hypothesize that dynamically combining steering vectors with the estimated time period can mitigate temporal misalignment, improving overall accuracy.

\paragraph{Setup.} We train a classifier to predict the time period on the data to construct steering vectors. We use DistillBERT \cite{sanh2019distilbert} as the base model. Suppose we have time periods $t_1, \ldots, t_d$ . Then, given a data point $x$, the time period classifier gives probability scores $p_1, \ldots, p_d$ such that $\sum_{i=1}^d p_i = 1$. The steering is conducted by \[\tilde{h}^l = h^l + \sum_{i=1}^d \alpha p_i v^l_{s \rightarrow t_i},\] where $s$ is the training time period. We use the same $\alpha$ as in Section \ref{subsec:exp1}. The test data is constructed by combining all years' test datasets.

\paragraph{Results.}
Table \ref{tab:dynamic} shows our results. Dynamic $\SYSNAME$ provides accuracy improvement comparable to or better than $\SYSNAME$ with the true time period, without any access to the time period of the target data points. We provide yearly results in Appendix \ref{appsec:exp4}.
\begin{table}[]
\centering
\begin{tabular}{l l l l}
\toprule
~ & Baseline & GT & Dynamic \\
\midrule
AIC & 83.81 & 85.86 & \textbf{86.58} \\
PoliAff &  69.38 & \textbf{71.65} & 70.89 \\
NewsCls & 78.54 & 79.00 & \textbf{79.30} \\
\bottomrule 
\end{tabular}
\caption{
	Average accuracy across the train years. $\SYSNAME$ (GT) uses the ground truth time stamp for representation steering, Dynamic $\SYSNAME$ uses the estimated time period. Dynamic $\SYSNAME$ yields accuracy enhancement comparable to or better than $\SYSNAME$ (GT).
}
\label{tab:dynamic}
\end{table}

\section{Conclusion} We introduced  $\SYSNAME$, a metho for efficient temporal adaptation with representation steering. We empirically show that $\SYSNAME$ can mitigate degradation from temporal misalignment without any fine-tuning. This work takes an initial step to adapt models to ever changing worlds without much cost. We also believe our work can provide clues for temporal reasoning tasks.

\paragraph{Limitations} Our work is limited to classification tasks and the model sizes are relatively small. Our future work will include more extensive tasks with large scaled models.

\bibliography{custom}

\onecolumn


\appendix
\addcontentsline{toc}{part}{Appendix} %
\part*{Appendix}
\parttoc %
\setcounter{figure}{0}
\renewcommand{\thefigure}{A\arabic{figure}}
\setcounter{table}{0}  
\renewcommand{\thetable}{A\arabic{table}}  

\section{Experiment details}\label{app:exp_details}

\subsection{Dataset Details.}\label{appsub:dataset}
\begin{itemize}
    \item AIC (AI venue classification): The task is to classify the academic venue (AAAI or ICML) given the abstract of the paper. The data is grouped into roughly equal-sized time periods (2009-2011, 2012-2014, 2015-2017, 2018-2020).
    
    \item PoliAff (Political affiliation classification): The task is to classify the writer's political affiliation into Democrat or Republican when given a tweet. We use uniformly-sized datasets for each year from 2015 to 2020, as in \citet{nylund2023time}.

    \item NewsCls (Newsroom publisher classification): The task is publisher classification (classes: FoxNews, NYTimes, WaPost) given an article. As in \citet{nylund2023time}, we use uniformly-sized datasets for each year from 2012 to 2016.
\end{itemize}

\begin{figure}[ht!]
	\centering
	\subfigure [AIC]{
\includegraphics[width=0.95\textwidth]{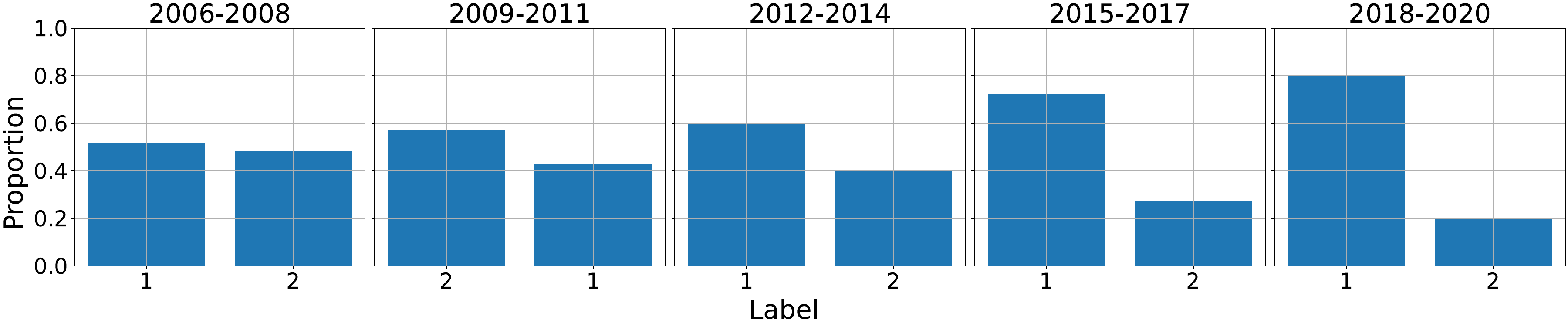}} 
\subfigure [PoliAff]{
\includegraphics[width=0.95\textwidth]{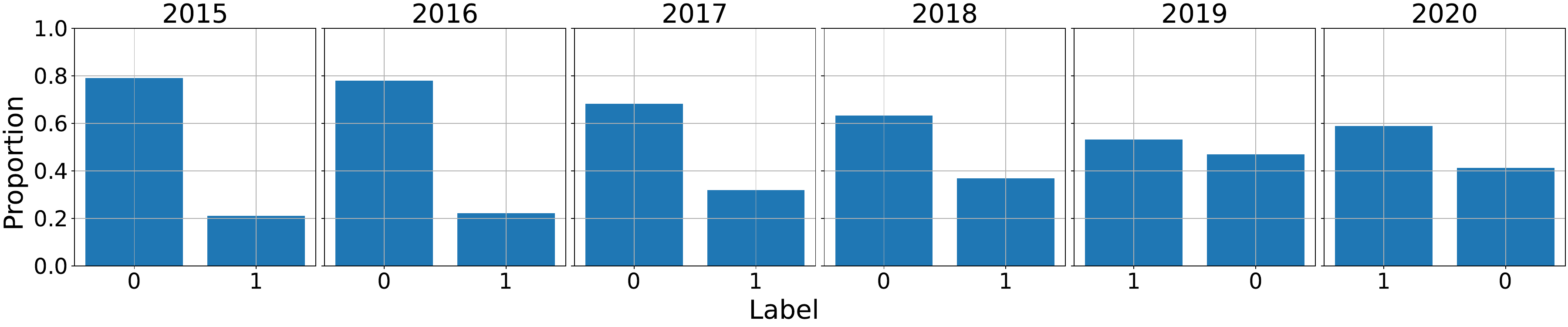}}
\subfigure [NewsCls]{
\includegraphics[width=0.95\textwidth]{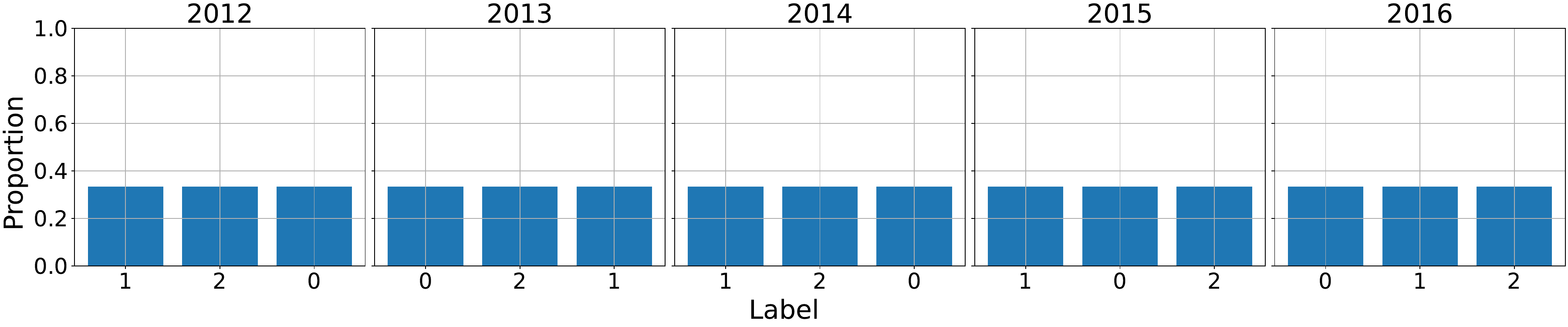}}
\caption{Label distribution of datasets}
    \label{fig:label_distribution}
\end{figure}

More descriptive statistics are given in Table \ref{tab:datasets}. Label distributions are visualized in \ref{fig:label_distribution}. Note that AIC and PoliAff have severe label shifts while NewsCls has a balanced label distribution across all years.

\begin{table}[h]
\centering
\begin{tabular}{l l l l l}
\toprule
~ & $n_{train}$ & $n_{val}$ & $n_{eval}$ & number of classes \\
\midrule
AIC & 2,400 & 400 & 400 & 2 \\
PoliAff & 10,000 & 1,500 & 1,500 & 2 \\
NewsCls & 2,124 & 195 & 213 & 3 \\
\bottomrule 
\end{tabular}
\caption{
	Descriptive statistics of datasets. Each time period datasets have roughly equal sizes as above.
}
\label{tab:datasets}
\end{table}

\newpage

\subsection{Detailed Experiment Results in Section \ref{subsec:exp1}} \label{appsec:exp1}
\begin{figure*}
    \centering
    \includegraphics[width=.9\textwidth]{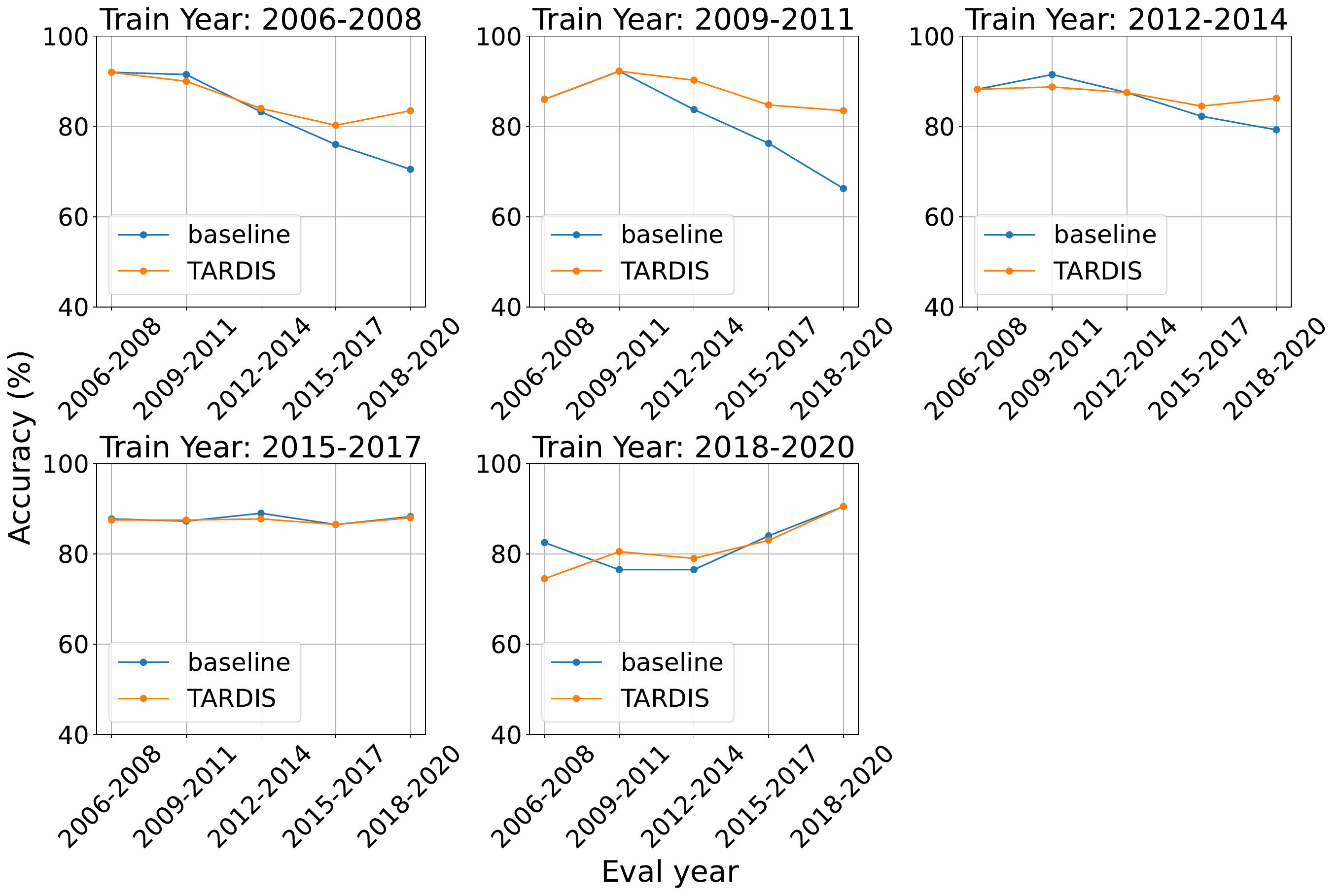}
    \caption{AIC Yearly Accuracy}
    \label{fig:aic_yearly}
\end{figure*}

\begin{figure*}
    \centering
    \includegraphics[width=.9\textwidth]{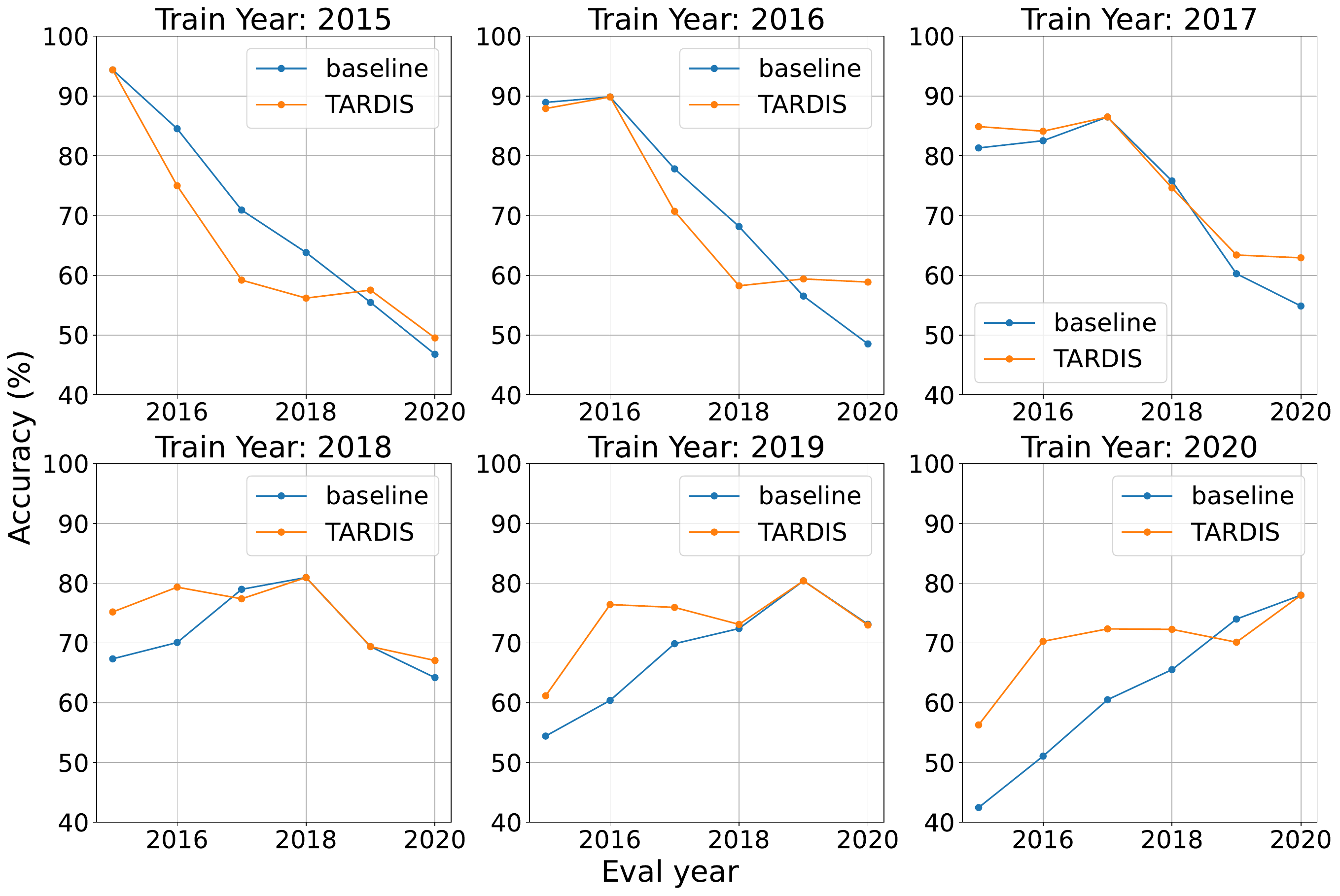}
    \caption{PoliAff Yearly Accuracy}
    \label{fig:poli_aff_yearly}
\end{figure*}

\begin{figure*}
    \centering
    \includegraphics[width=.9\textwidth]{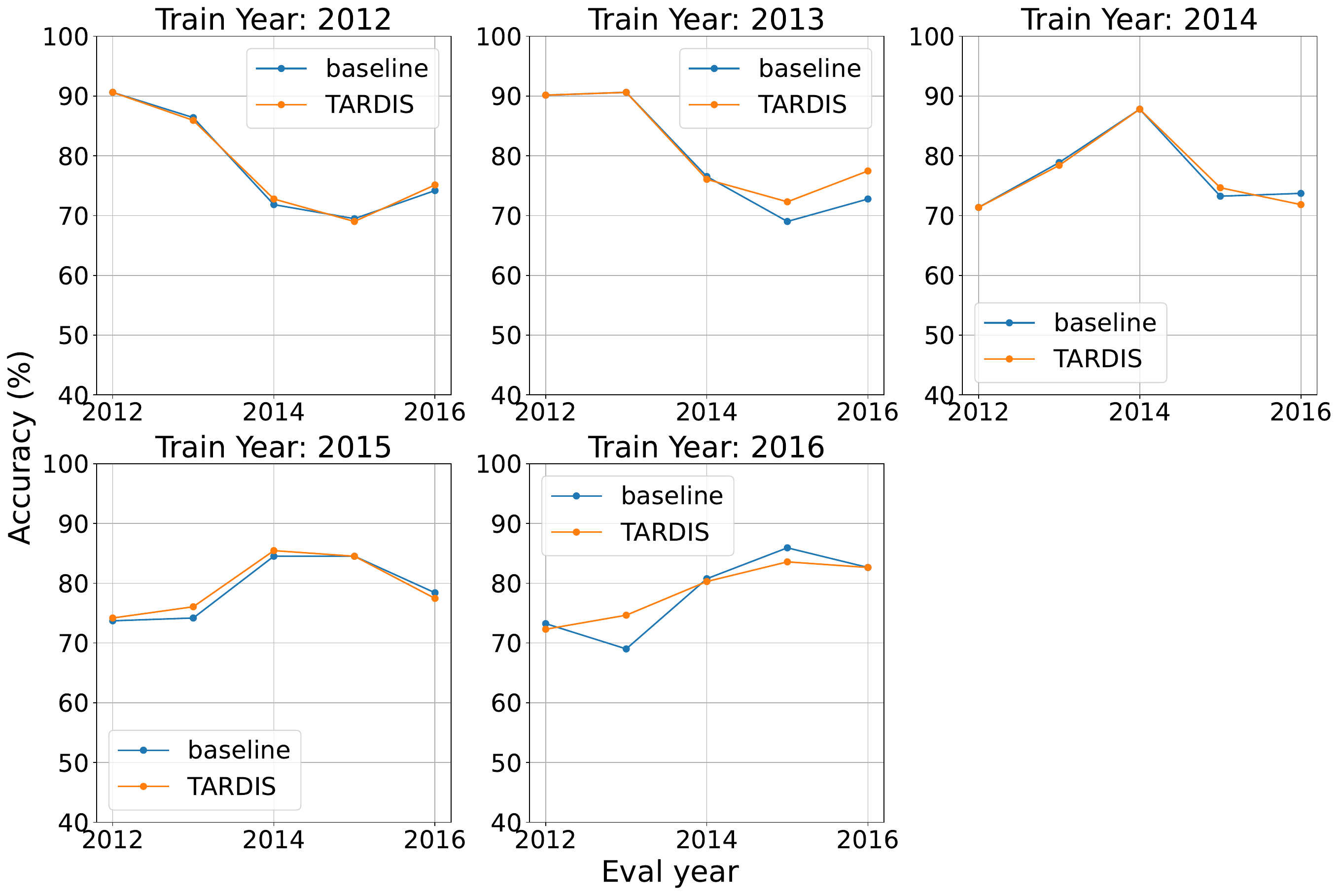}
    \caption{NewsCls Yearly Accuracy}
    \label{fig:news_cls_yearly}
\end{figure*}

\paragraph{Yearly Accuracy Comparison.} We report detailed yearly accuracy comparison in Figure \ref{fig:aic_yearly} ,\ref{fig:poli_aff_yearly}, \ref{fig:news_cls_yearly}.

\begin{figure*}[ht!]
\centering
\subfigure{\includegraphics[width=0.48\textwidth]{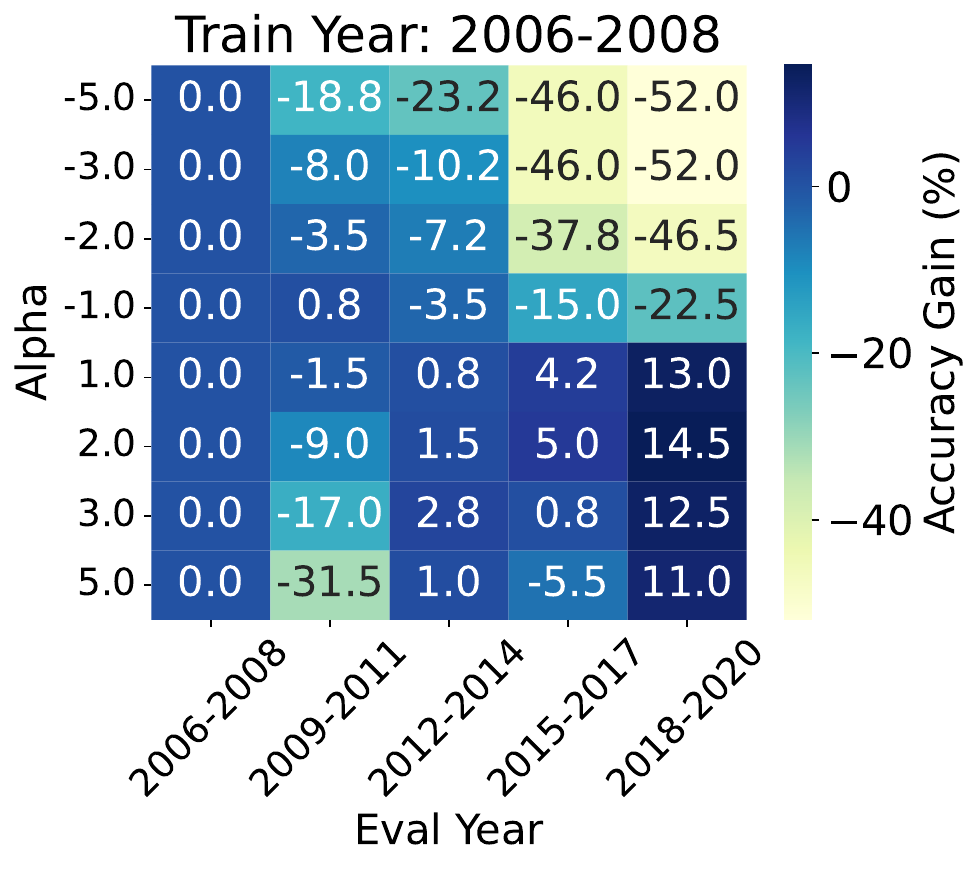}}
\subfigure{\includegraphics[width=0.48\textwidth]{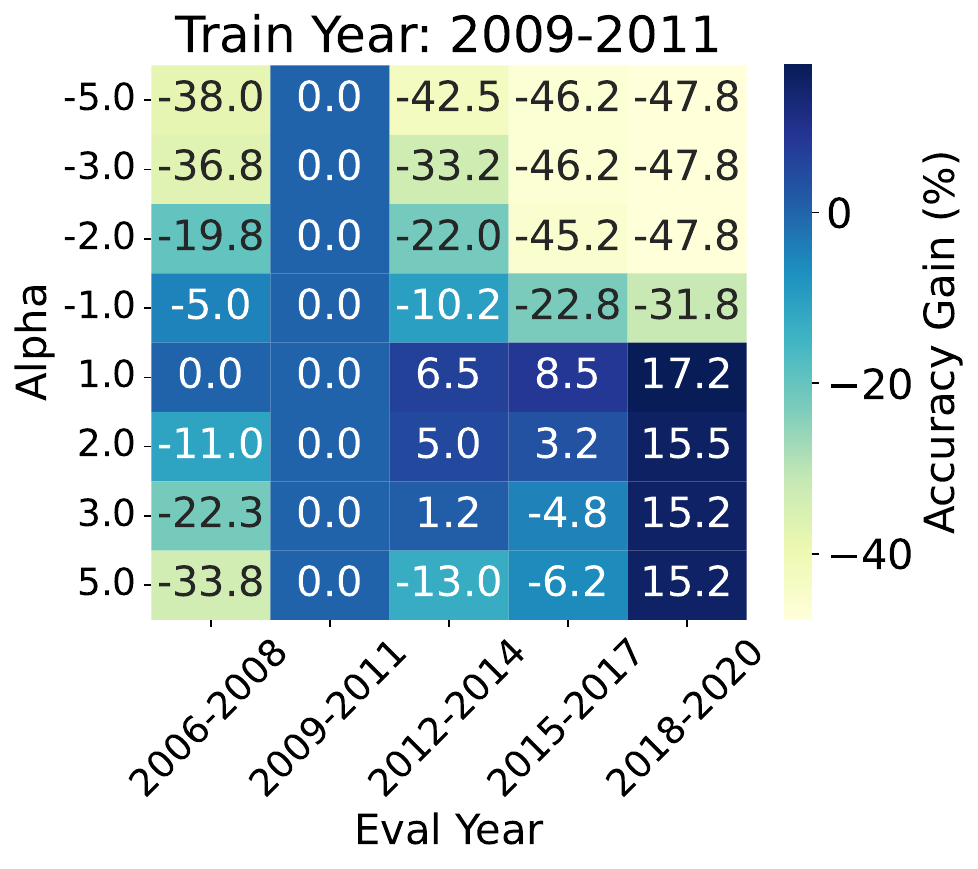}}
\subfigure{\includegraphics[width=0.48\textwidth]{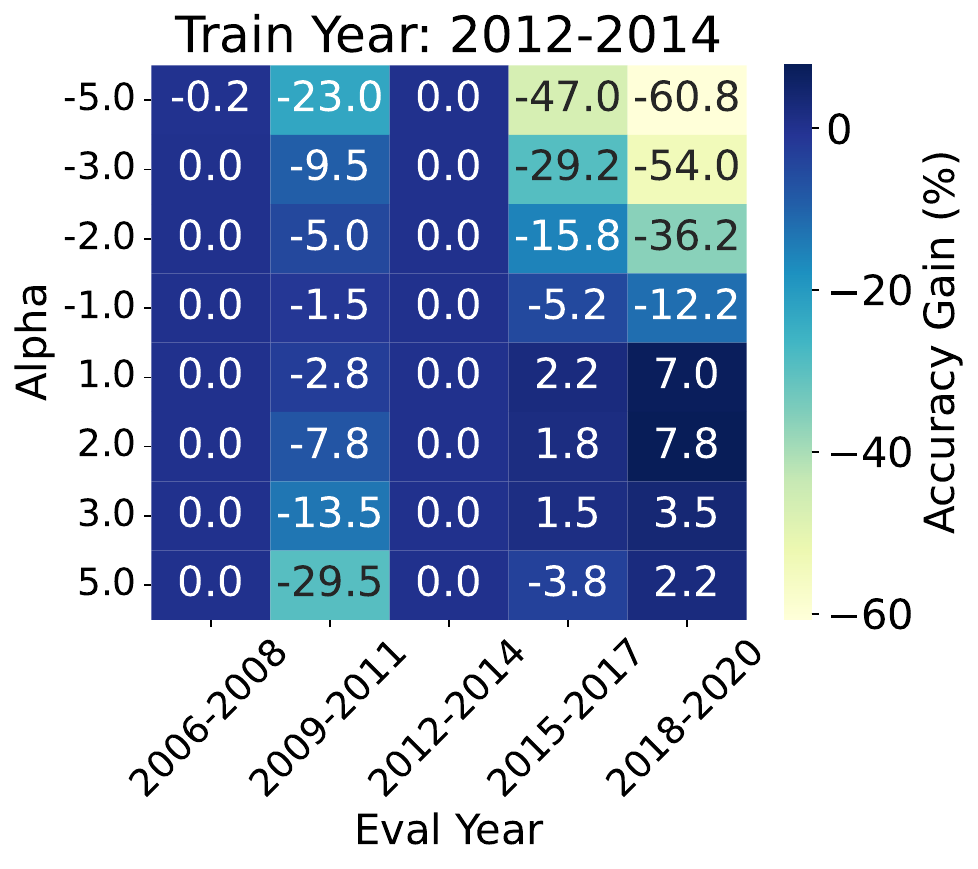}}
\subfigure{\includegraphics[width=0.48\textwidth]{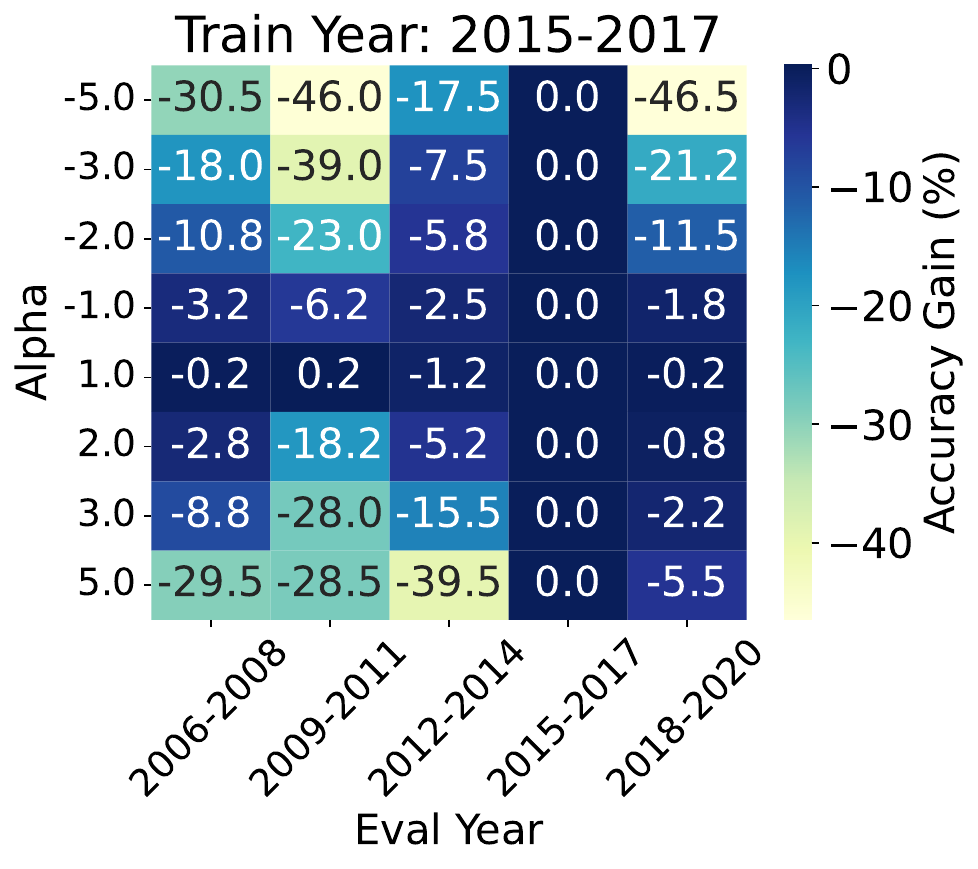}}
\subfigure{\includegraphics[width=0.48\textwidth]{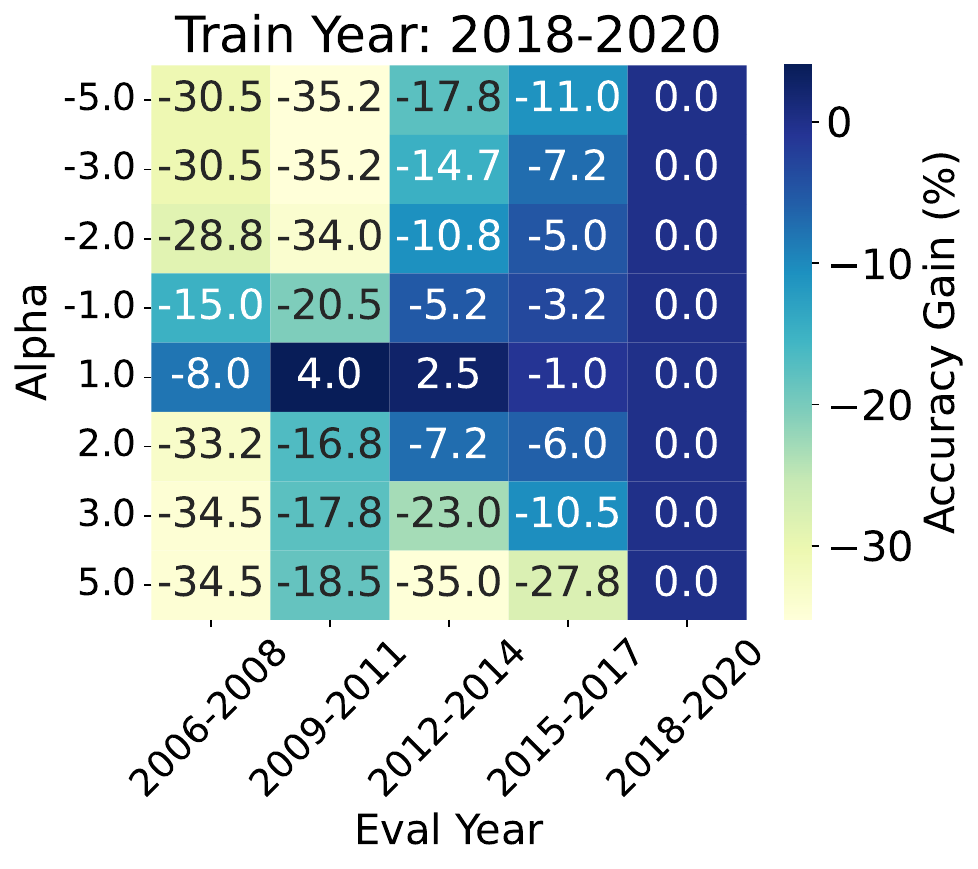}}

\caption{AIC: Performance gain by $\SYSNAME$ depending on $\alpha$.}
    \label{fig:steer_alpha_ablation_aic}
\end{figure*}

\begin{figure*}[ht!]
\centering
\subfigure{\includegraphics[width=0.48\textwidth]{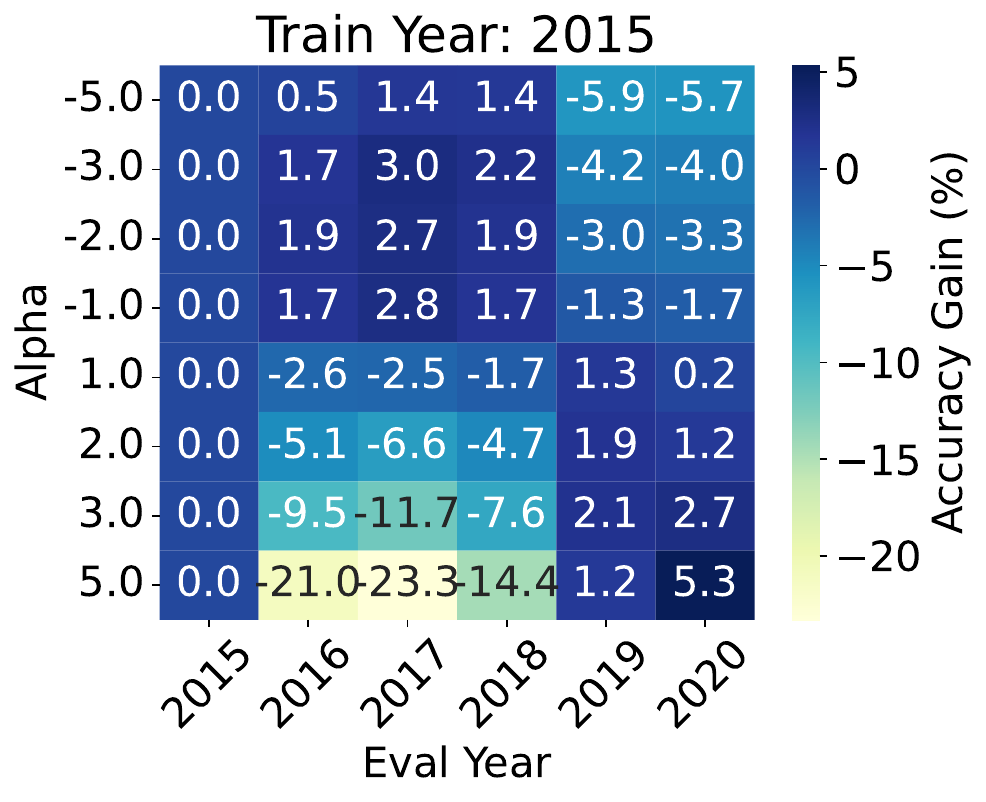}}
\subfigure{\includegraphics[width=0.48\textwidth]{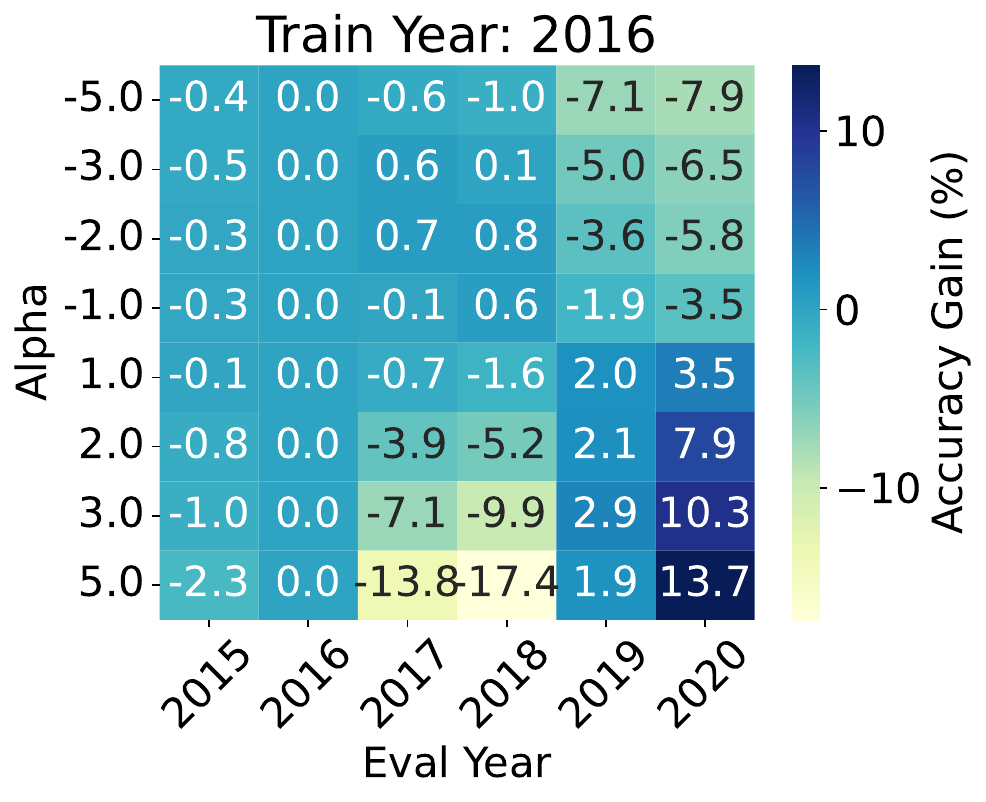}}
\subfigure{\includegraphics[width=0.48\textwidth]{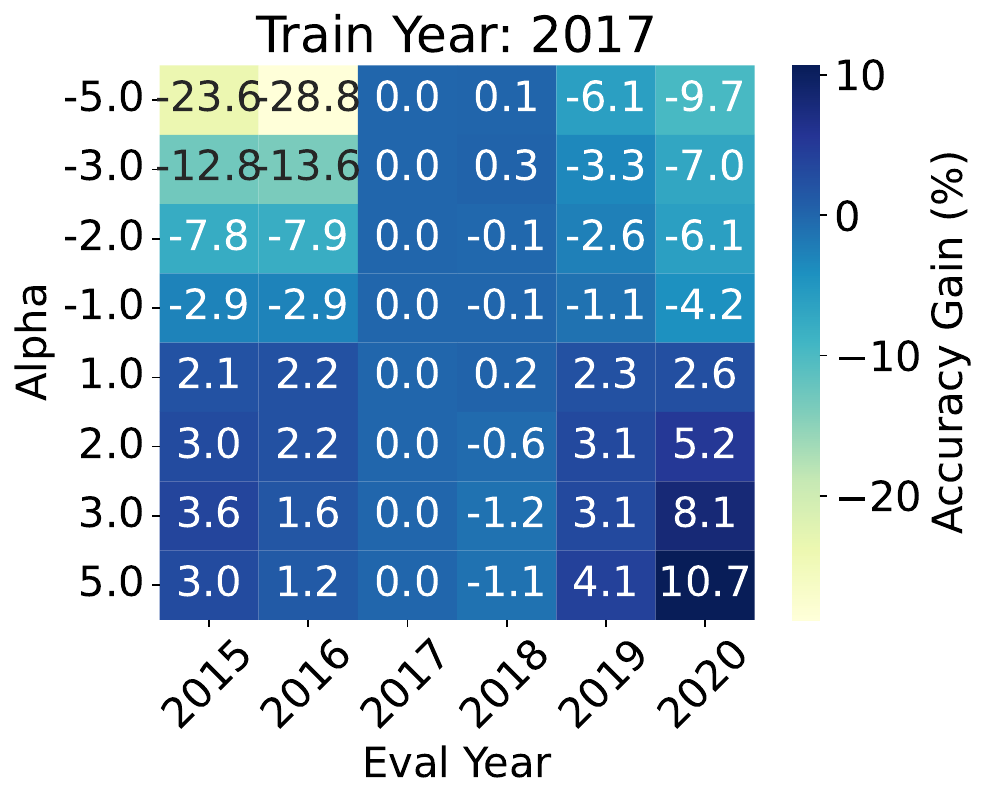}}
\subfigure{\includegraphics[width=0.48\textwidth]{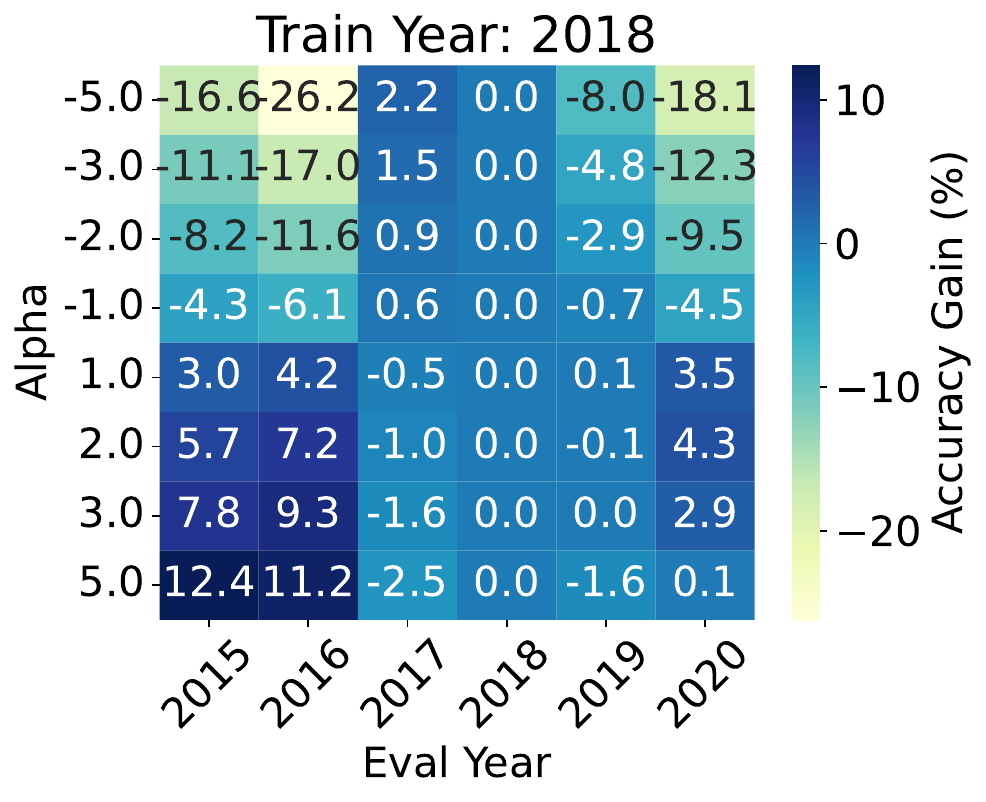}}
\subfigure{\includegraphics[width=0.48\textwidth]{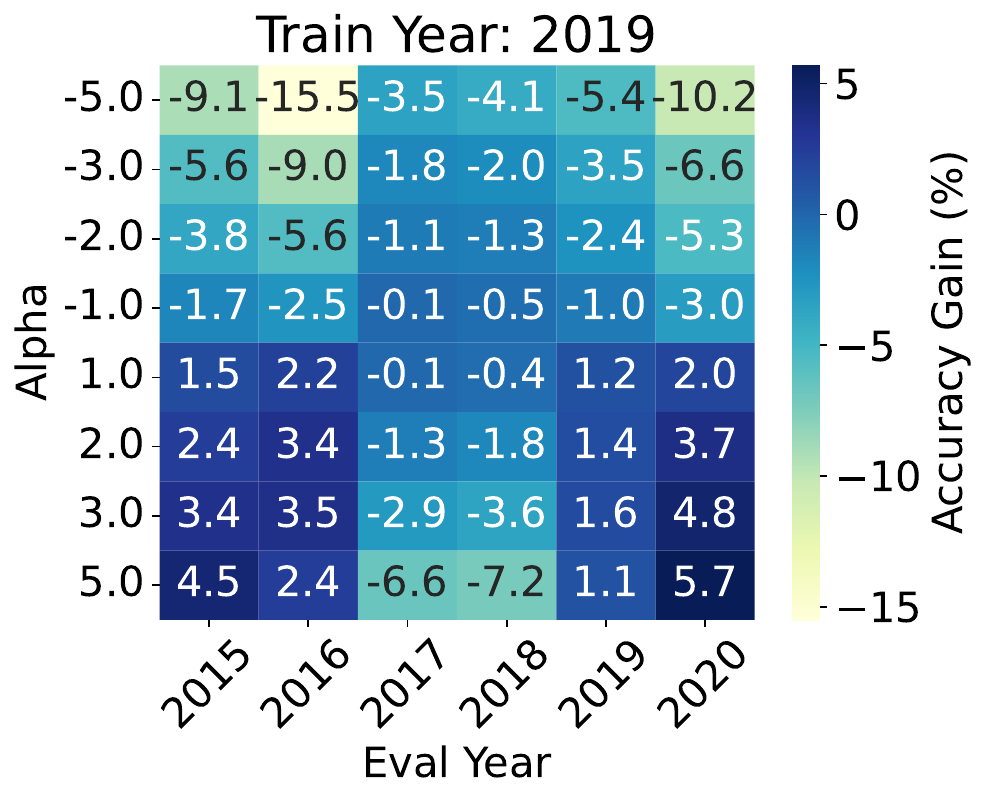}}
\subfigure{\includegraphics[width=0.48\textwidth]{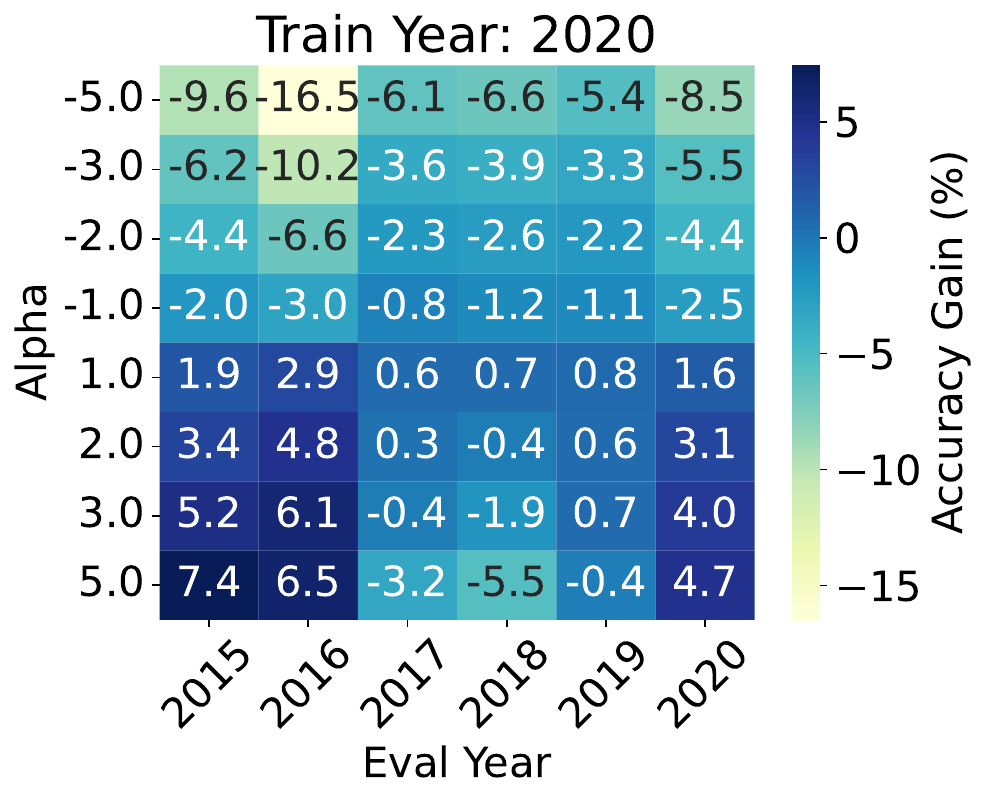}}

\caption{PoliAff: Performance gain by $\SYSNAME$ depending on $\alpha$.}
    \label{fig:steer_alpha_ablation_poli_aff}
\end{figure*}

\begin{figure*}[ht!]
\centering
\subfigure{\includegraphics[width=0.48\textwidth]{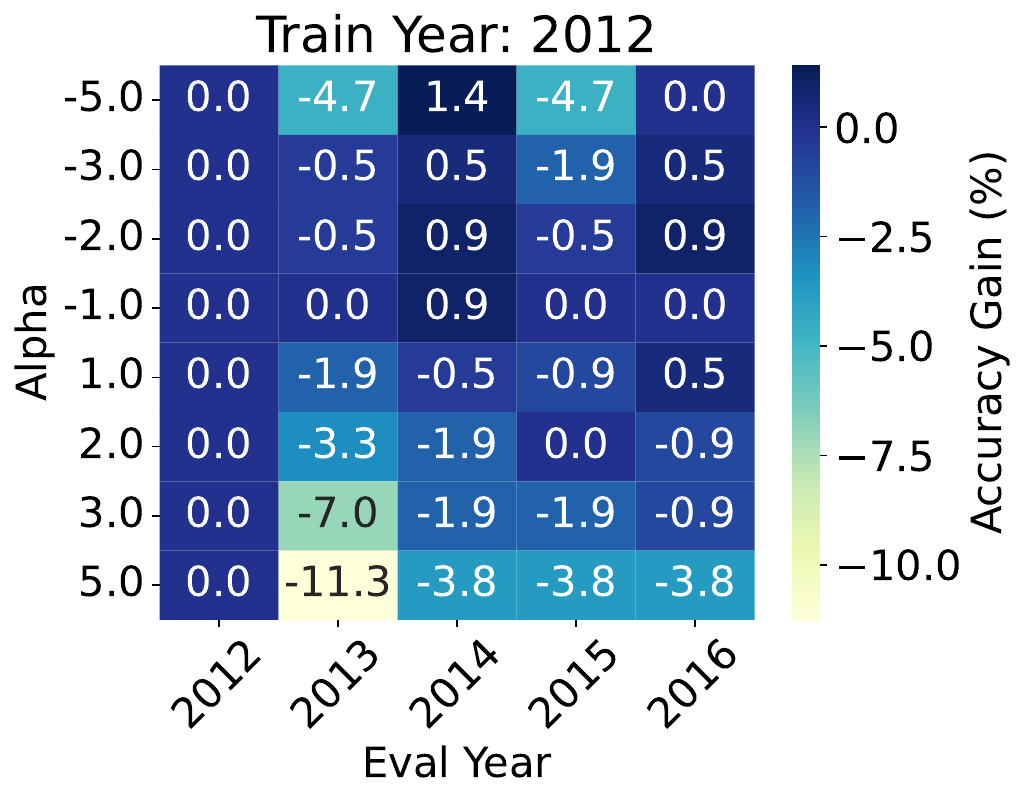}}
\subfigure{\includegraphics[width=0.48\textwidth]{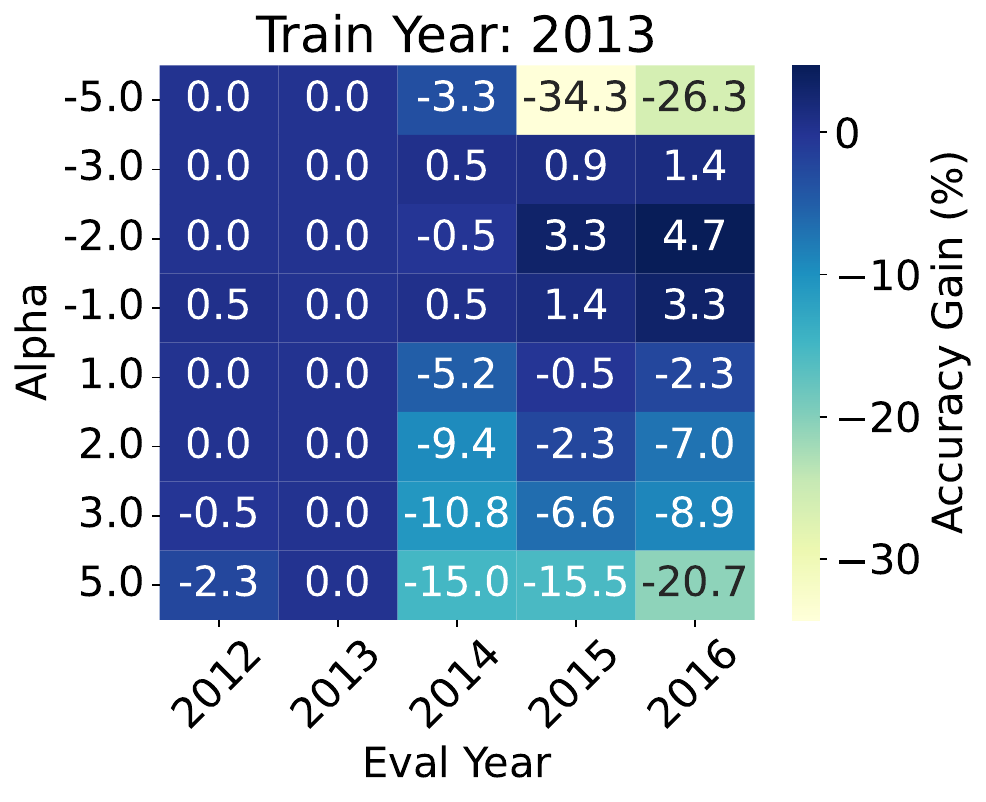}}
\subfigure{\includegraphics[width=0.48\textwidth]{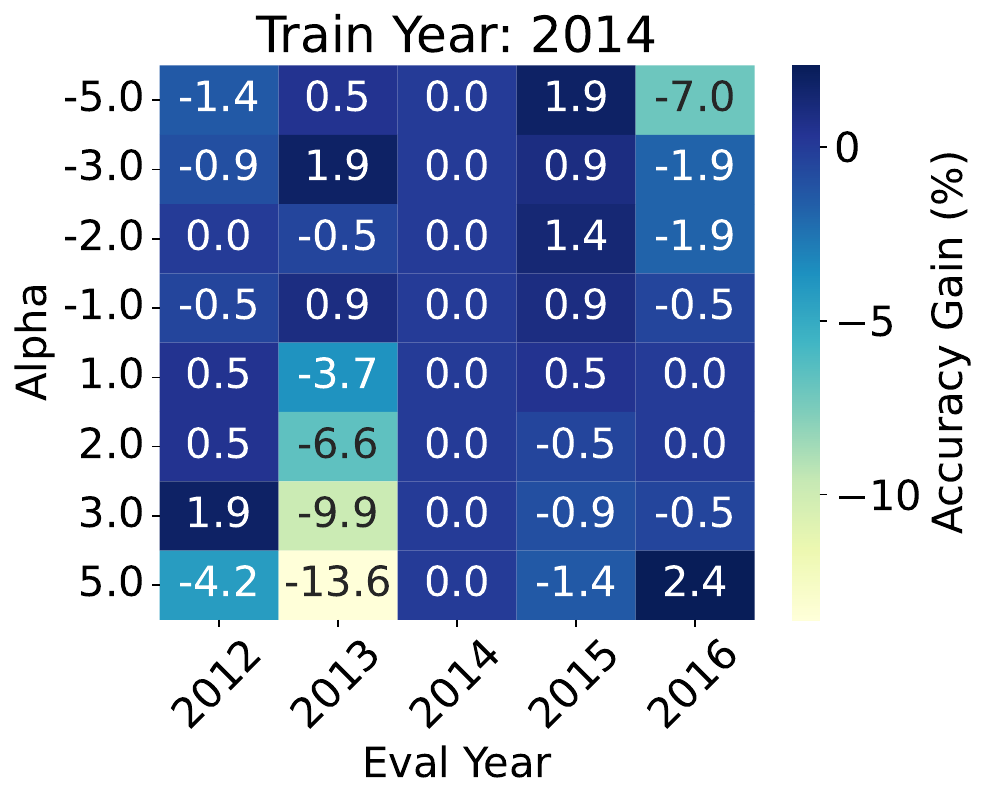}}
\subfigure{\includegraphics[width=0.48\textwidth]{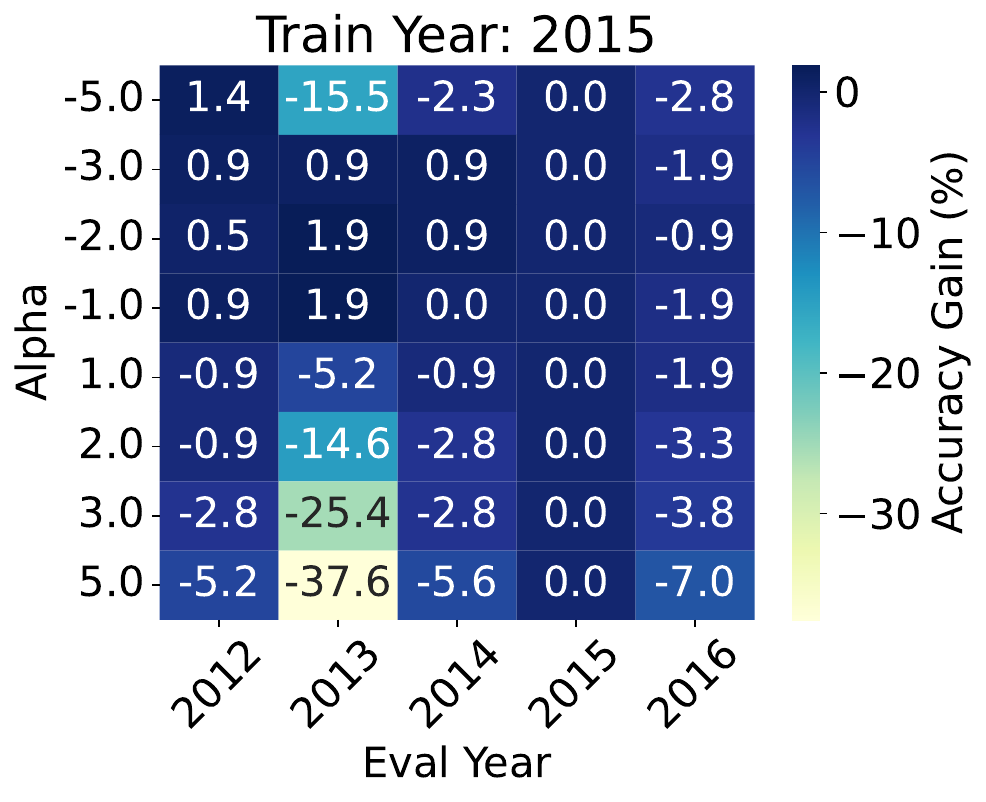}}
\subfigure{\includegraphics[width=0.48\textwidth]{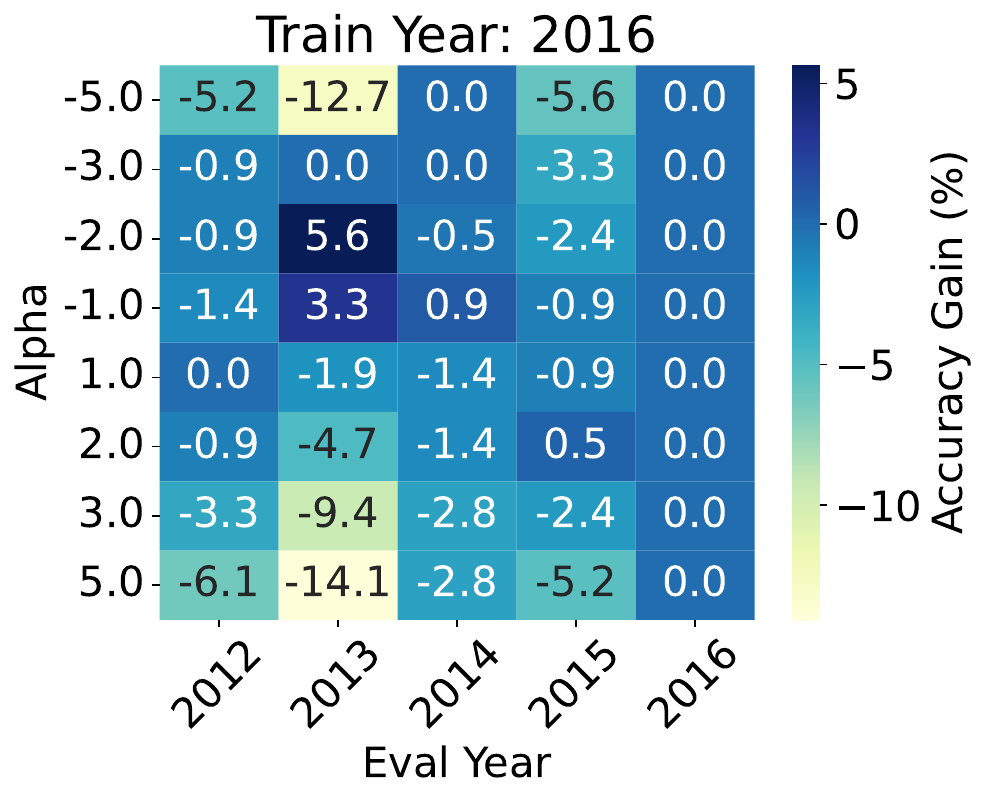}}
\caption{NewsCls: Performance gain by $\SYSNAME$ depending on $\alpha$.}
    \label{fig:steer_alpha_ablation_newscls}
\end{figure*}

\newpage
\paragraph{Ablation test on steering coefficient $\alpha$.} We provided detailed accuracy enhancement/degradation depending on $\alpha$ in Figure \ref{fig:steer_alpha_ablation_aic}, \ref{fig:steer_alpha_ablation_poli_aff}, \ref{fig:steer_alpha_ablation_newscls}.

\begin{figure*}[ht!]
	\centering
	\subfigure [AIC]{
\includegraphics[width=0.33\textwidth]{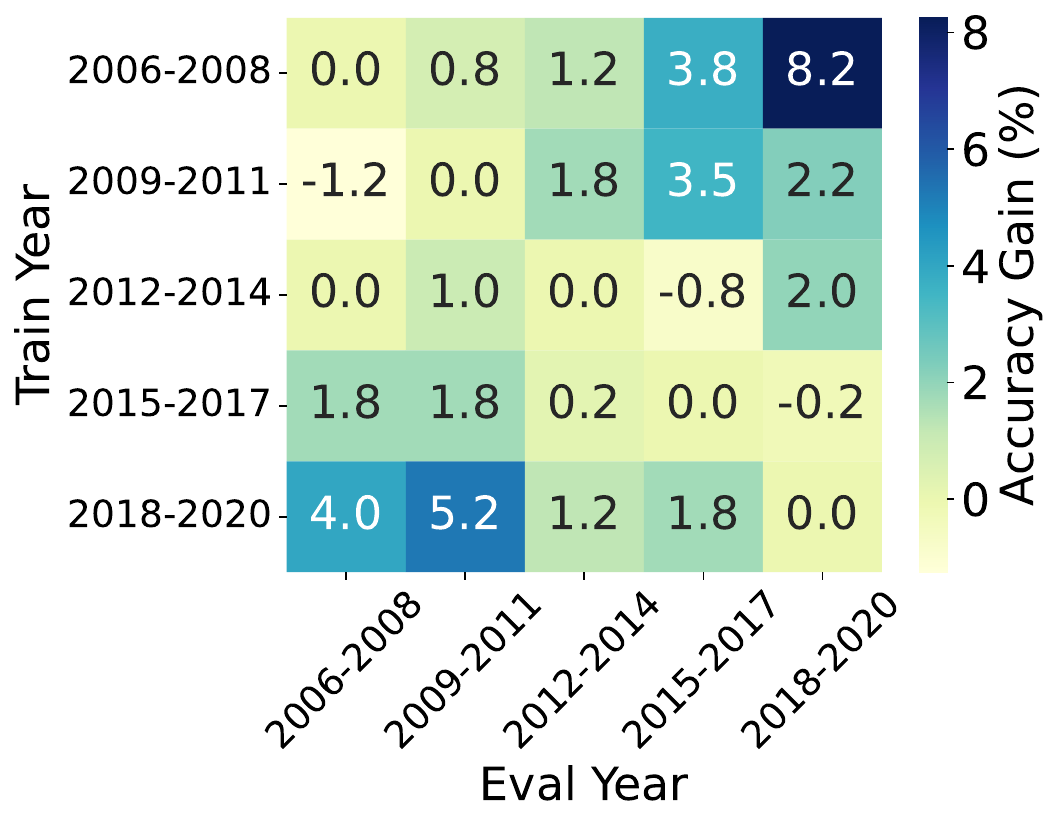}}
\subfigure [PoliAff]{
\includegraphics[width=0.33\textwidth]{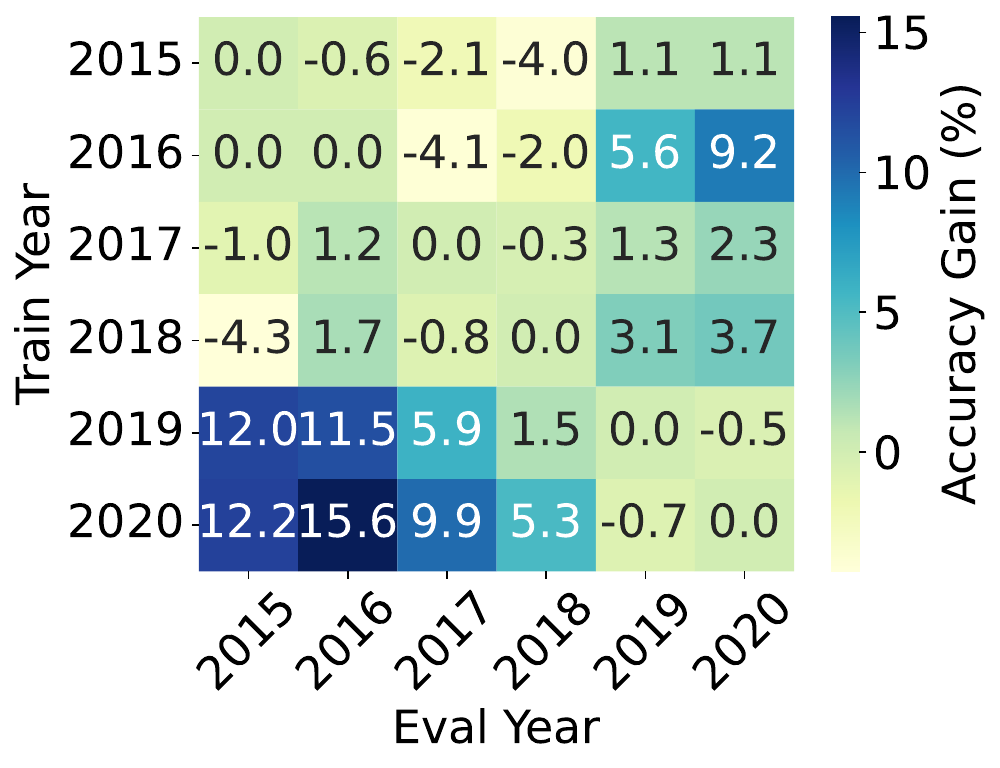}}
\subfigure [NewsCls]{
\includegraphics[width=0.31\textwidth]{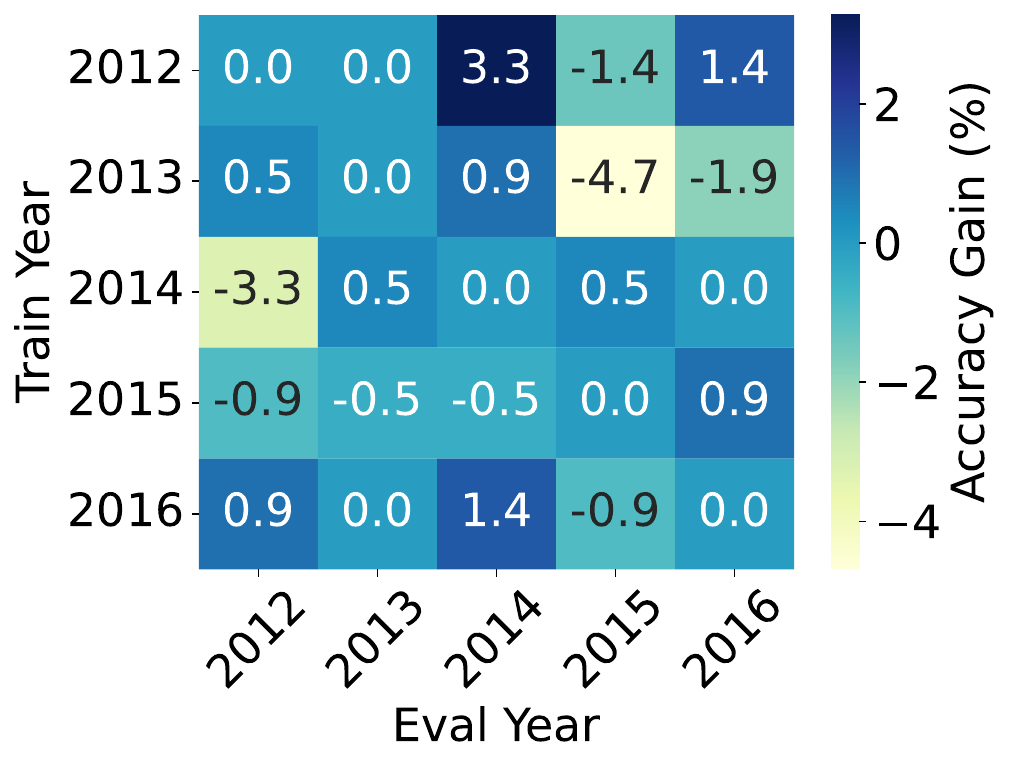}}

\caption{Performance gains when using $\SYSNAME$ on t5-770M. We observe that $\SYSNAME$ can improve accuracy up to 15.6\% without any fine-tuning.}
    \label{fig:steer_770m_results}
\end{figure*}

\newpage
\paragraph{T5-770M Experiment Results.} We repeat the experiments in Section \ref{subsec:exp1} with a larger model (T5-770M) and visualize the accuracy improvements in Figure \ref{fig:steer_770m_results}. We observe result similar to Section \ref{subsec:exp1}.

\begin{figure*}[ht!]
	\centering
	\subfigure [AIC]{
\includegraphics[width=0.33\textwidth]{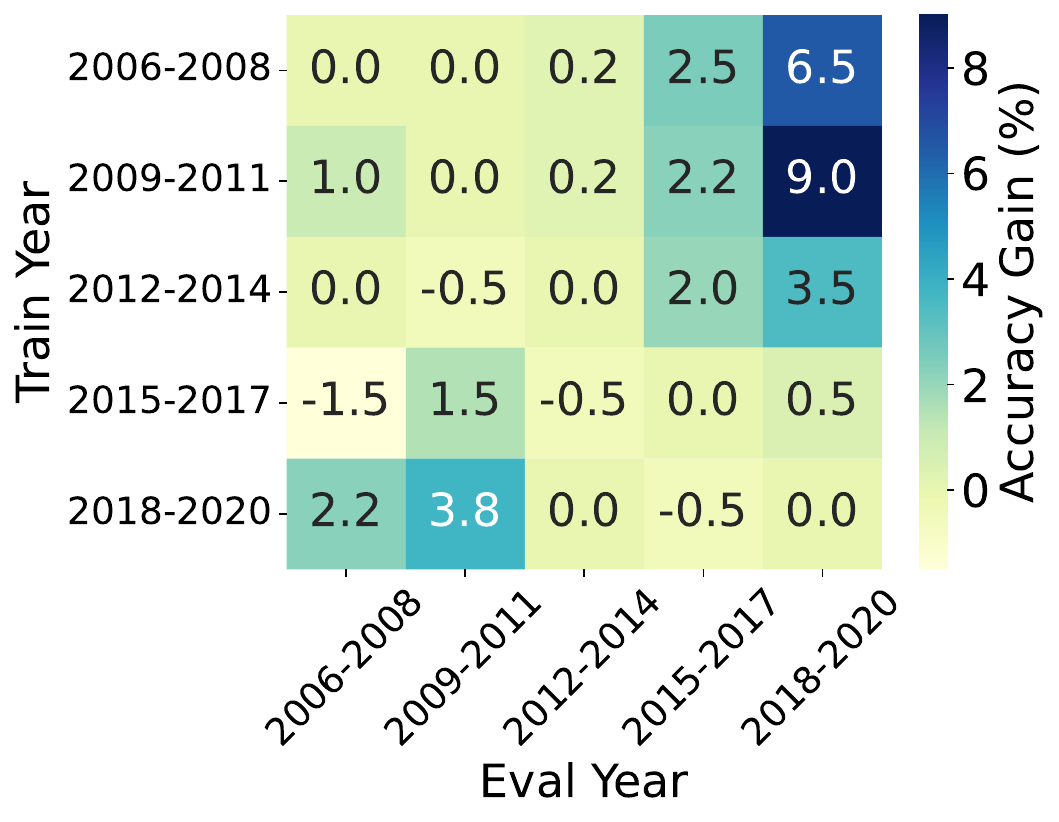}}
\subfigure [PoliAff]{
\includegraphics[width=0.33\textwidth]{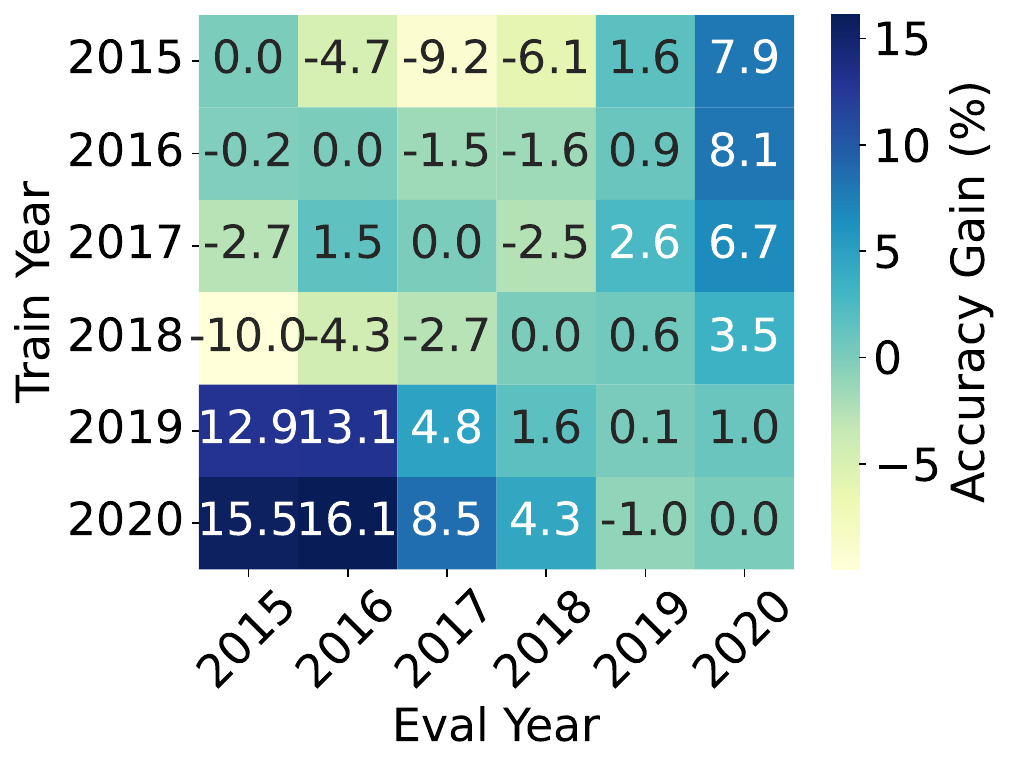}}
\subfigure [NewsCls]{
\includegraphics[width=0.31\textwidth]{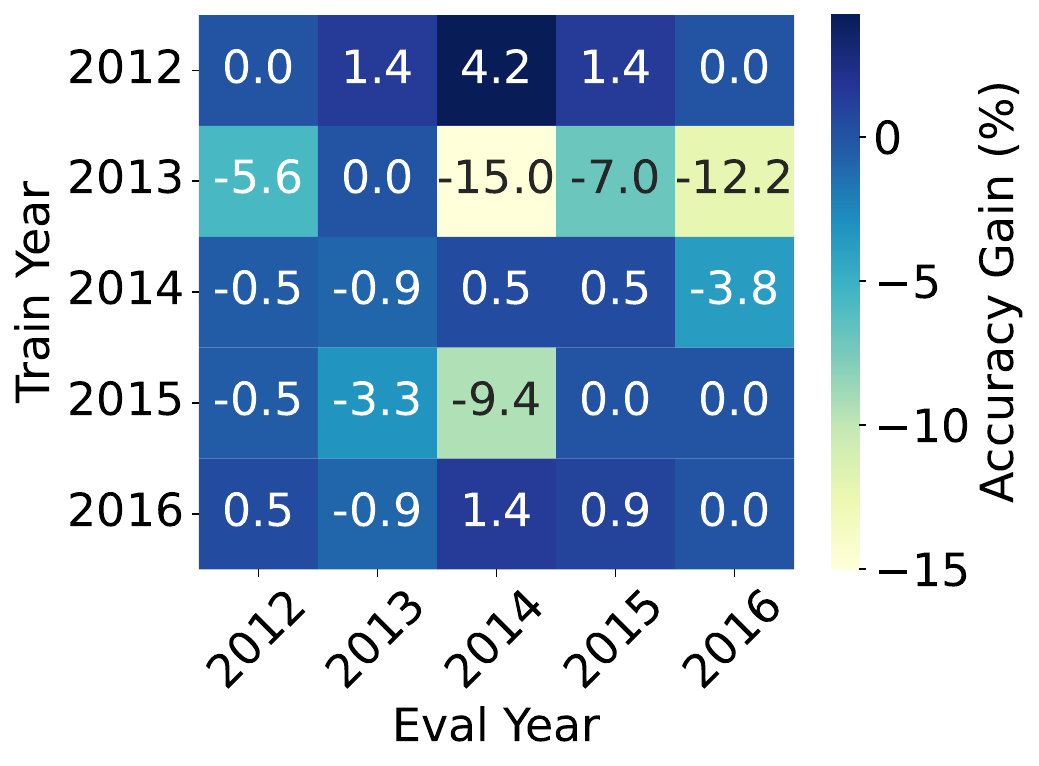}}

\caption{Performance gains when using $\SYSNAME$ on GPT2. We observe that $\SYSNAME$ can improve accuracy up to 16.1\% without any fine-tuning.}
    \label{fig:steer_gpt2_results}
\end{figure*}

\paragraph{GPT2 Experiment Results.} We repeat the experiments in Section \ref{subsec:exp1} with GPT2, a decoder-only architecture. In GPT2, we intervene all MLP layers with $\SYSNAME$ since there is no encoder part. Figure \ref{fig:steer_gpt2_results} shows the results. While the observations in AIC, PoliAff are similar, NewsCls result is quite different. It seems that adapting to vocabulary shift is more challenging in a decoder-only architecture. We leave this for future work.

\subsection{Details of Section \ref{subsec:exp2}} \label{appsec:exp2}
\paragraph{Label Shift Experiment Setup.} We chose the first time period as the target time period and used the finetuned model on the training set with the same time period. To control the label distribution, we iteratively remove 30, 100, 10 examples from one fixed class until there is no remaining data points in that fixed class. Steering vectors are computed with the original full data and the shrunk data.

\begin{figure*}[ht!]
\centering
\subfigure[AIC]{\includegraphics[width=0.32\textwidth]{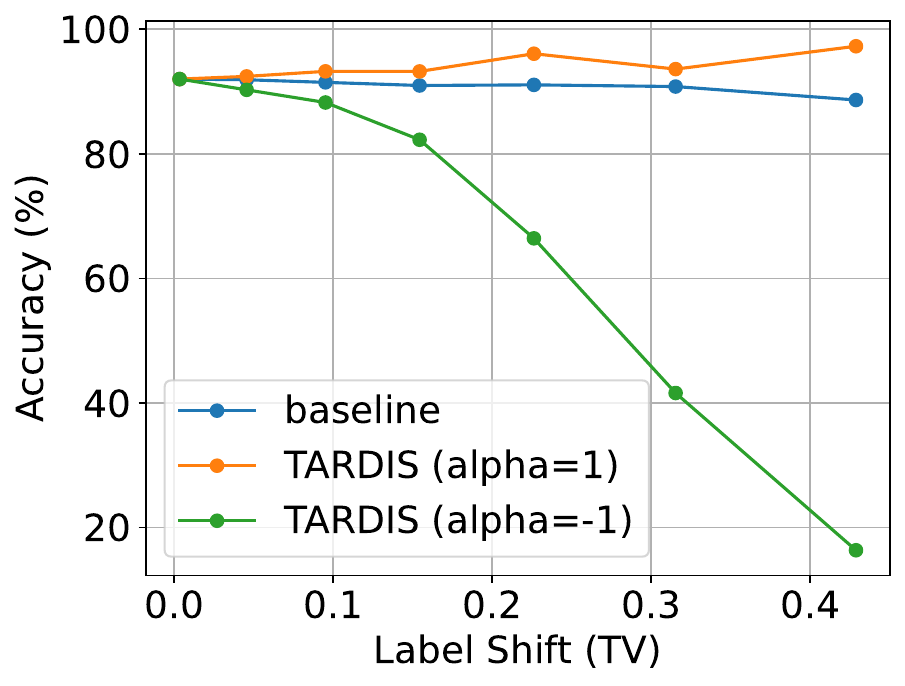}}
\subfigure[PoliAff]{\includegraphics[width=0.32\textwidth]{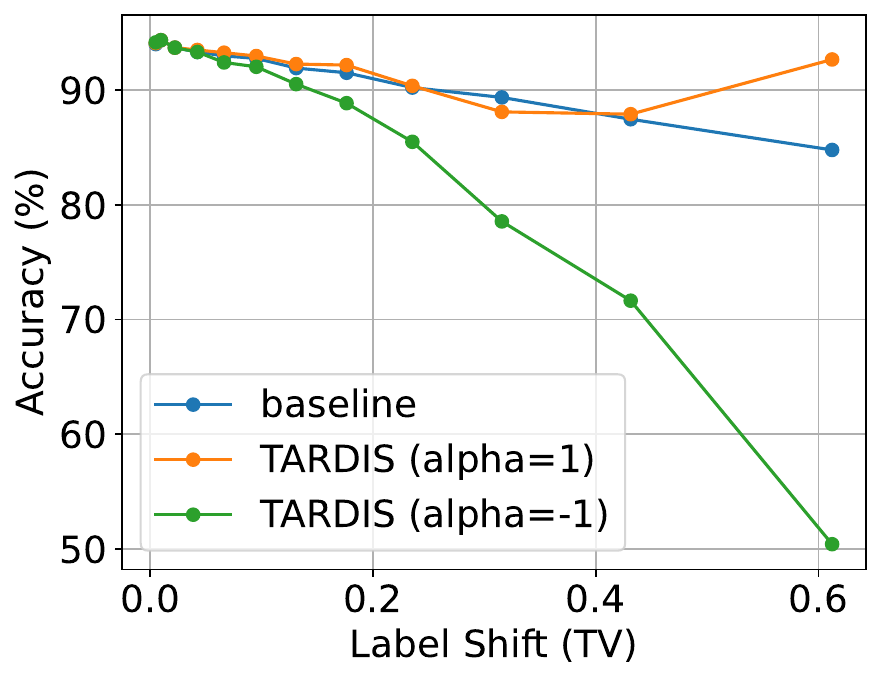}}
\subfigure[NewsCls]{\includegraphics[width=0.32\textwidth]{figures/label-shift_news_cls.pdf}}
\caption{Semi-synthetic Label Shift Experiment Results.}
    \label{fig:label_shift_full}
\end{figure*}

\paragraph{Label Shift Experiment Results.} Figure \ref{fig:label_shift_full} shows the experiment results for all datasets. We can see that the positive alpha leads to performance improvement in label shift setup.

\paragraph{Vocabulary Shift Experiment Setup.} We used the finetuned model on the training set with the first time period. Varying the test time period, we resample data to make the label distribution same with the training time period. Steering vectors are computed using the training time period data and the sampled target period data.

\begin{figure*}[ht!]
\centering
\subfigure[AIC]{\includegraphics[width=0.32\textwidth]{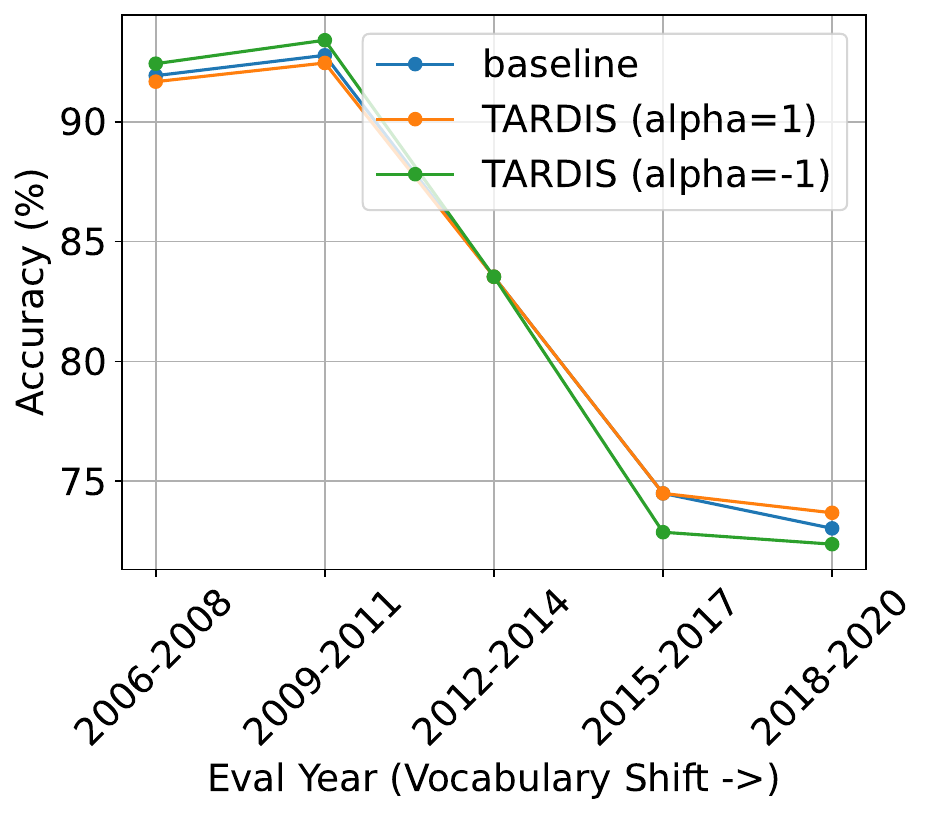}}
\subfigure[PoliAff]{\includegraphics[width=0.32\textwidth]{figures/vocabulary-shift_poli_aff.pdf}}
\subfigure[NewsCls]{\includegraphics[width=0.32\textwidth]{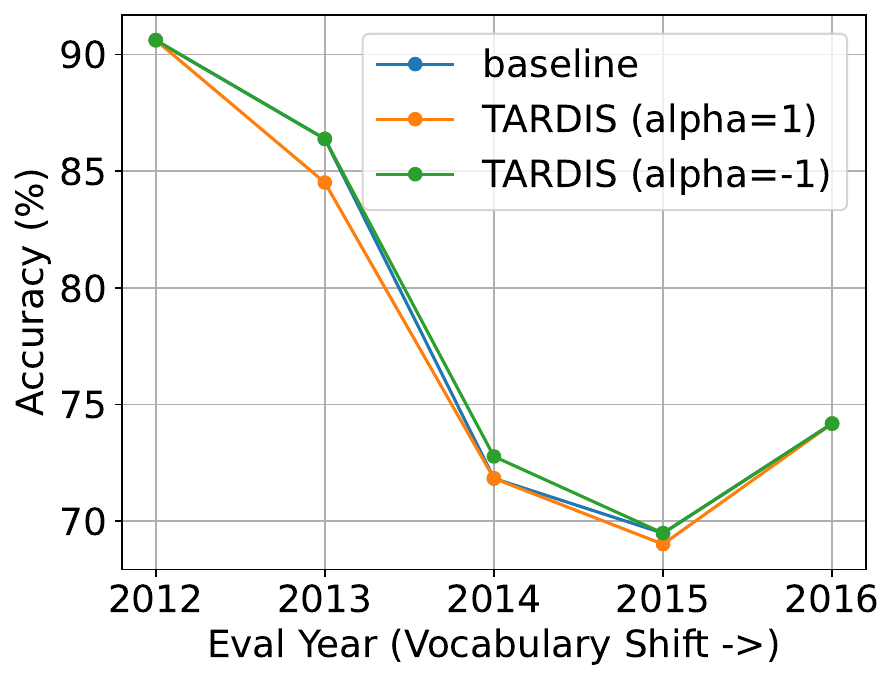}}
\caption{Semi-synthetic Vocabulary Shift Experiment Results.}
    \label{fig:vocab_shift_full}
\end{figure*}

\paragraph{Vocabulary Shift Experiment Results.} Figure \ref{fig:vocab_shift_full} shows the experiment results for all datasets. We can see that the negative alpha can lead to performance improvement in vocabulary shift setup.

\newpage
\subsection{Details of Section \ref{subsec:exp3}}\label{appsec:exp3}
\begin{figure*}[ht!]
\centering
\subfigure[AIC]{\includegraphics[width=0.32\textwidth]{figures/time-on-the-line_forward_aic.pdf}}
\subfigure[PoliAff]{\includegraphics[width=0.32\textwidth]{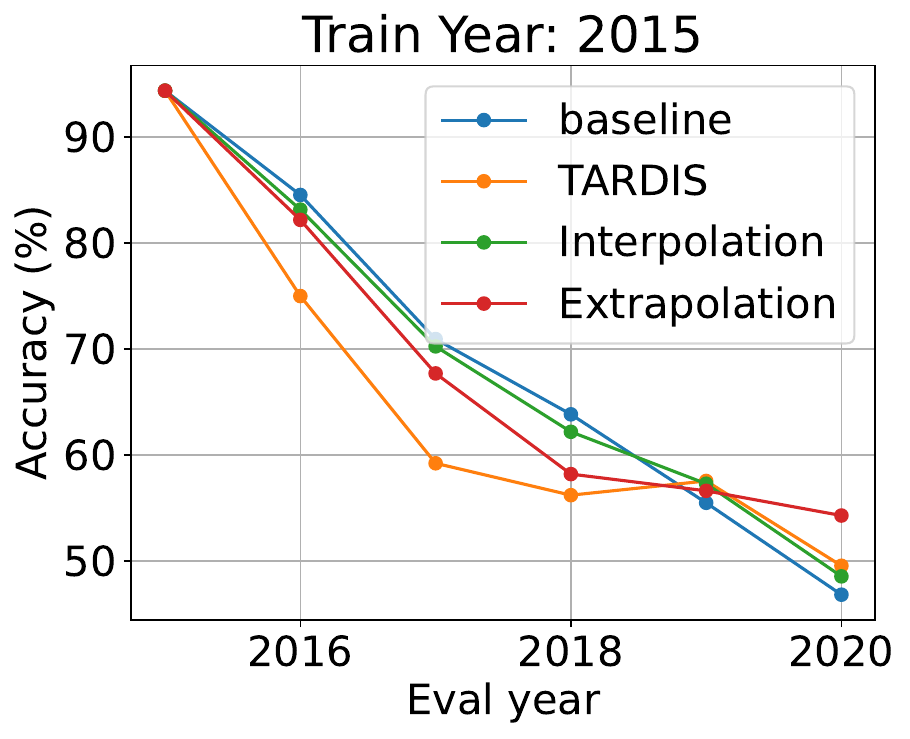}}
\subfigure[NewsCls]{\includegraphics[width=0.32\textwidth]{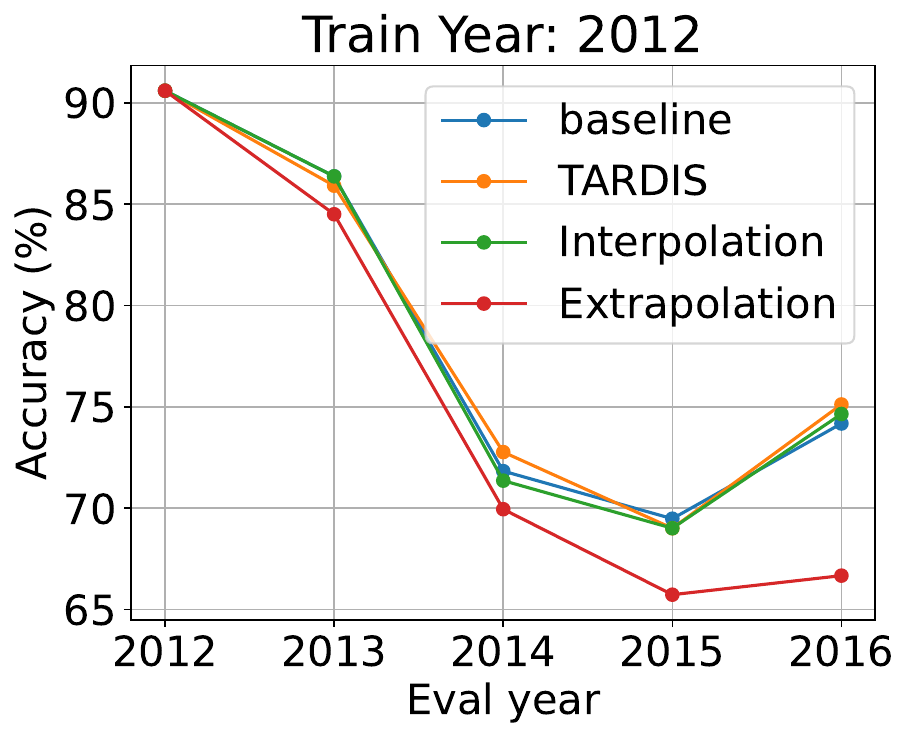}}
\caption{Semi-synthetic Vocabulary Shift Experiment Results.}
    \label{fig:forward_full}
\end{figure*}

\begin{figure*}[ht!]
\centering
\subfigure[AIC]{\includegraphics[width=0.32\textwidth]{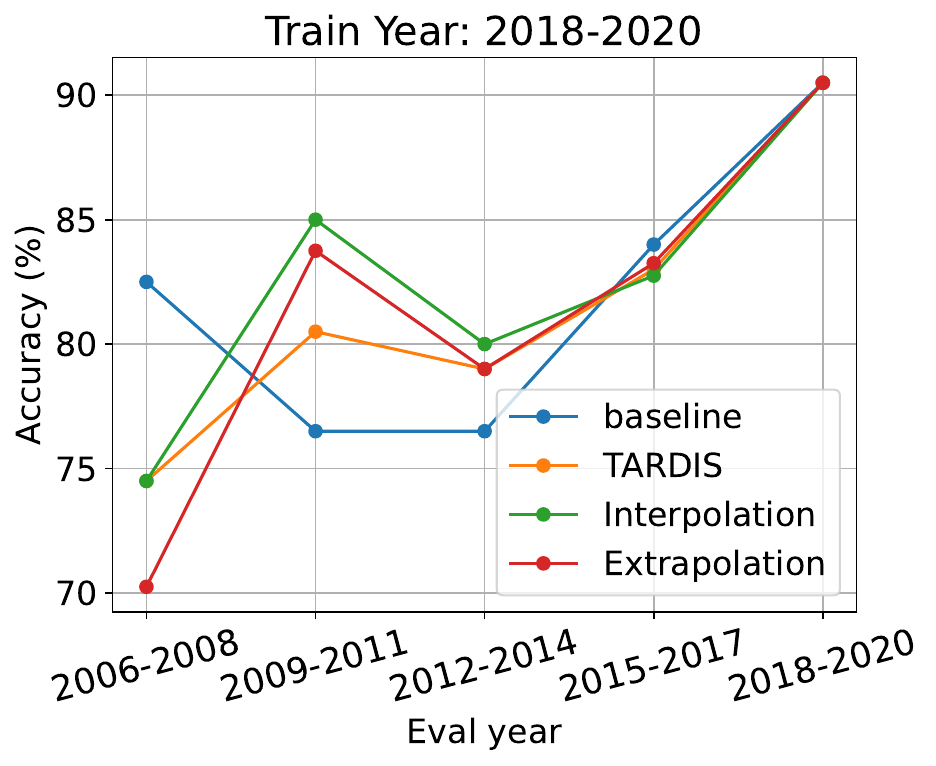}}
\subfigure[PoliAff]{\includegraphics[width=0.32\textwidth]{figures/time-on-the-line_backward_poli_aff.pdf}}
\subfigure[NewsCls]{\includegraphics[width=0.32\textwidth]{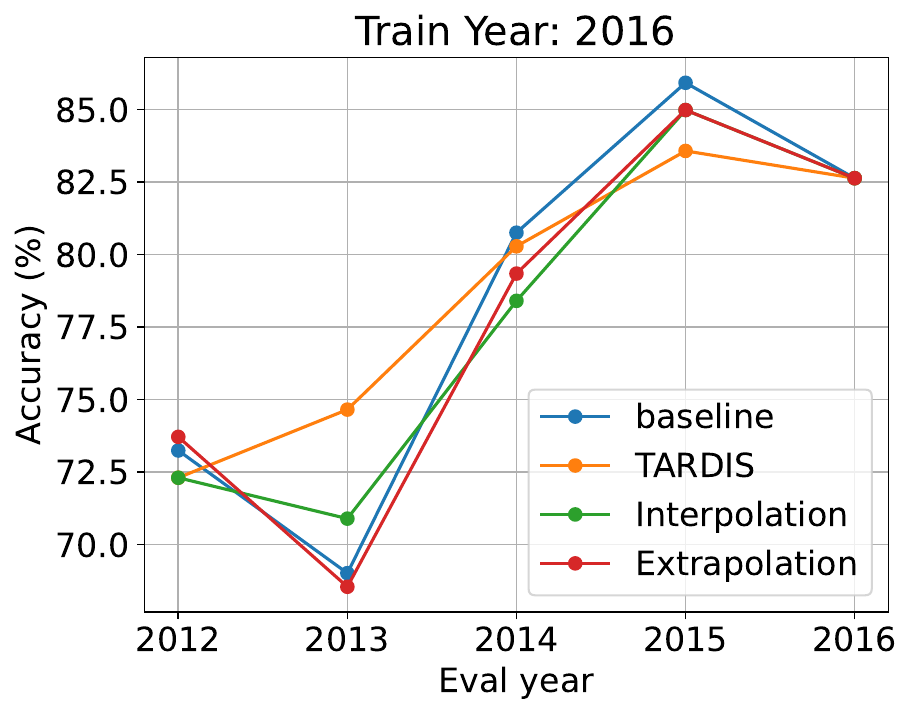}}
\caption{Semi-synthetic Vocabulary Shift Experiment Results.}
    \label{fig:backward_full}
\end{figure*}

We report forward interpolation/extrapolation experiment results in Figure \ref{fig:forward_full} and backward interpolation/extrapolation experiment results in Figure \ref{fig:backward_full}

\subsection{Details of Section \ref{subsec:exp4}}\label{appsec:exp4}
\paragraph{Time Period Classifier Training Details} We trained DistillBERT on the validation set, which is again split into time period classifier training, test datasets with 7:3 ratio. We used learning rate 2e-5, batch size 8, epochs 10. Obtained best accuracies were 38.6\%, 45.4\%, 45.1\% in AIC, PoliAff, NewsCls.

\begin{figure*}
    \centering
    \includegraphics[width=.9\textwidth]{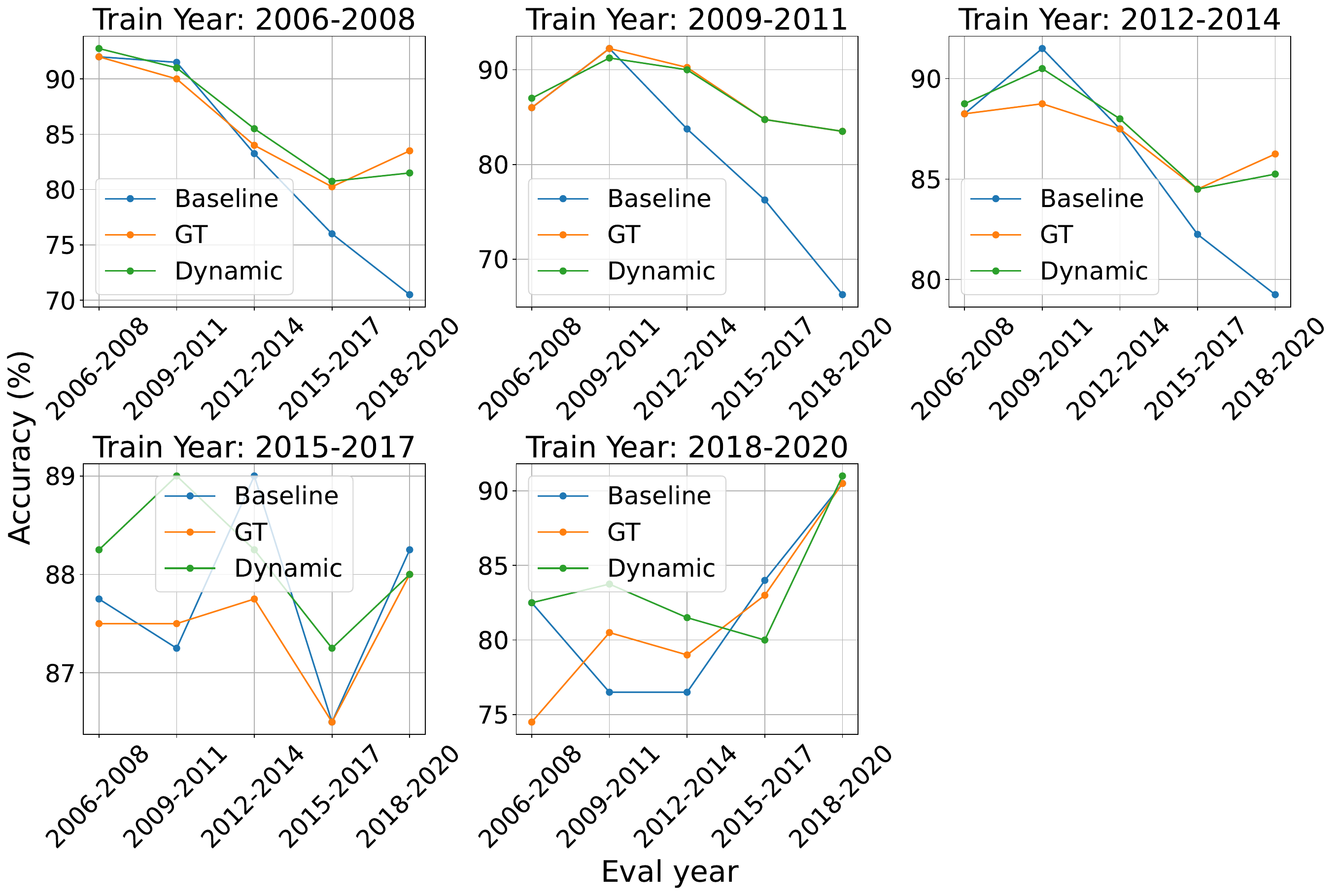}
    \caption{Dynamic $\SYSNAME$ AIC Yearly Accuracy}
    \label{fig:dynamic_aic_yearly}
\end{figure*}

\begin{figure*}
    \centering
    \includegraphics[width=.9\textwidth]{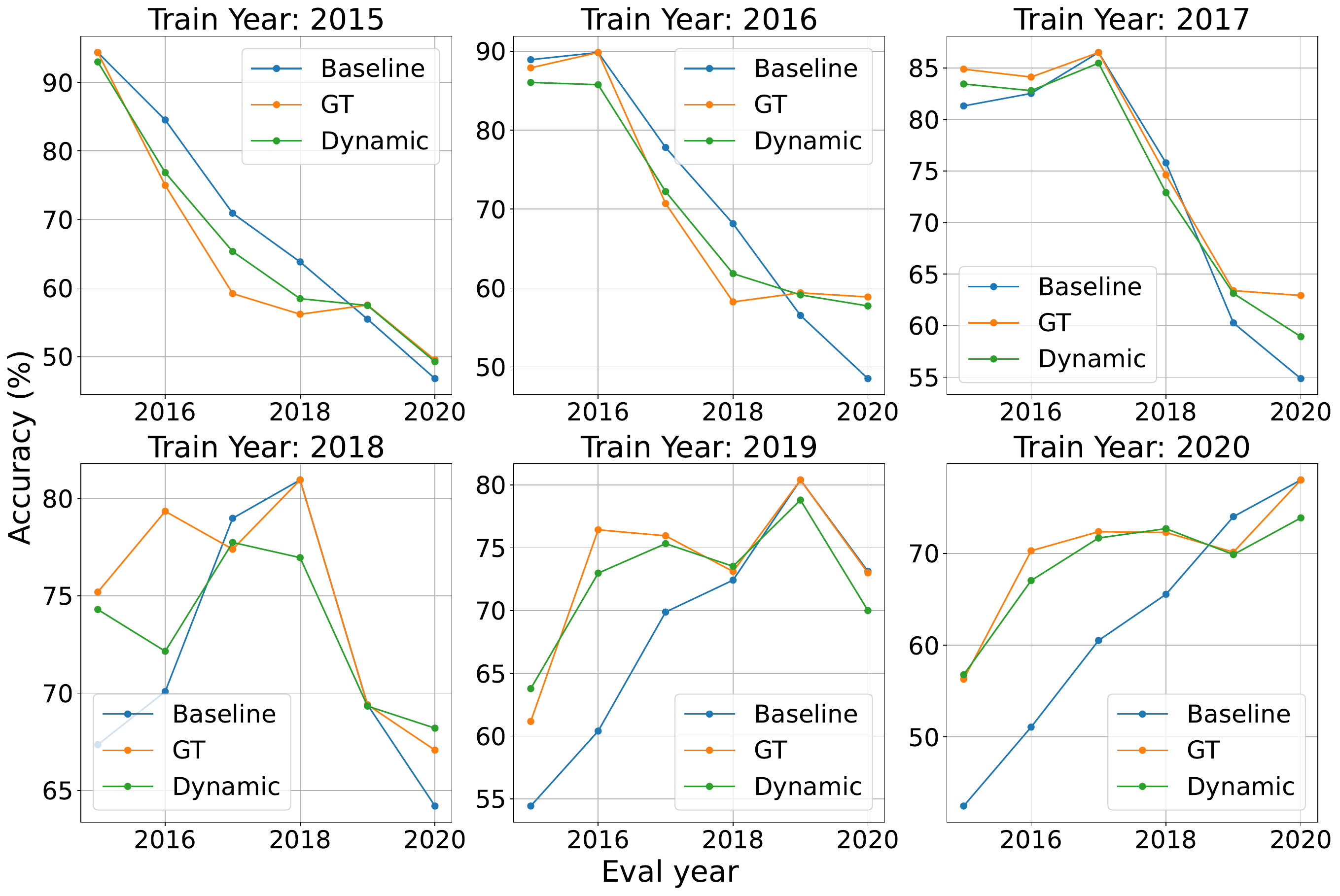}
    \caption{Dynamic $\SYSNAME$ PoliAff Yearly Accuracy}
    \label{fig:dynamic_poli_aff_yearly}
\end{figure*}

\begin{figure*}
    \centering
    \includegraphics[width=.9\textwidth]{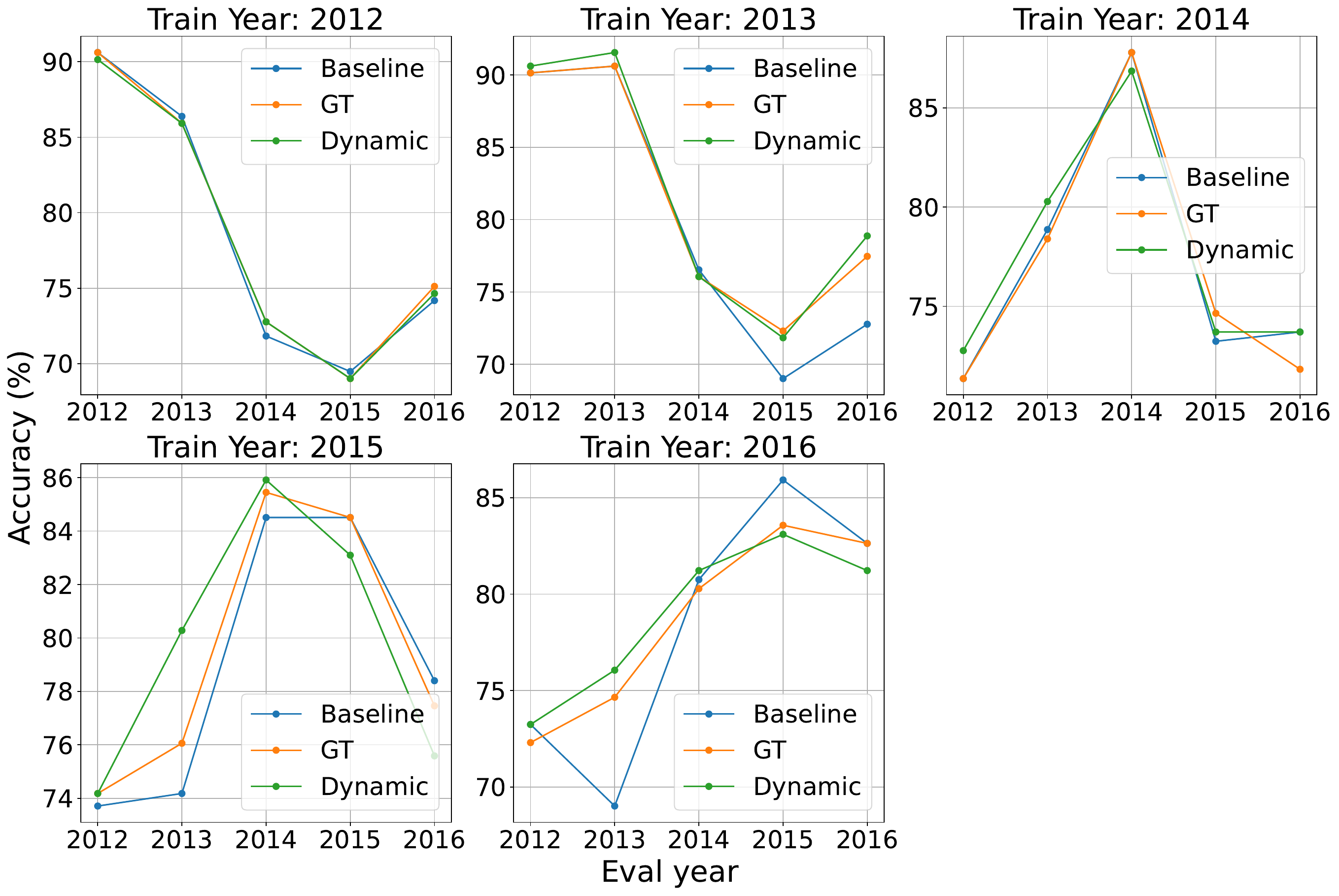}
    \caption{Dynamic $\SYSNAME$ NewsCls Yearly Accuracy}
    \label{fig:dynamic_news_cls_yearly}
\end{figure*}

\paragraph{Yearly Results} We visualize yearly accuracy comparison in Figure \ref{fig:dynamic_aic_yearly}, \ref{fig:dynamic_poli_aff_yearly}, \ref{fig:dynamic_news_cls_yearly}. They show that Dynamic $\SYSNAME$ can mitigate degradation by temporal misalignment well, even without access to the true time period of the target data points.

\paragraph{Computing Resources} We used a machine equipped with an Intel® Core™ i7-11700K @ 3.60GHz processor, 64GB RAM, and NVIDIA GPU RTX-4090.

\clearpage
\section{Additional Experiments}\label{app:additional_exp}

\subsection{Denoising Steering Vectors via Low-Rank Approximation}\label{subsec:exp_low_rank_approx}

Note that TARDIS formulates steering vectors by averaging across all representations from a certain time period. However, not all data points equally exhibit time-specific information, and thus taking the average likely introduces noise that hinders proper temporal alignment. How can we \textit{denoise} the representations towards isolating out the direction of temporal shift? In this experiment, we hypothesize that temporal shift can be accurately represented in low-dimensions and perform low-rank approximation on the set of representations for denoising.

\paragraph{Setup.} Let $\mathbf{H}_s^l = [h_1^{s,l}; \cdots; h_{n_s}^{s,l}] \in \mathbb{R}^{d \times n_s}$ denote the column-wise concatenation of representations from time period $s$ ($\mathbf{H}_t^l \in \mathbb{R}^{d \times n_t}$ for time period $t$, respectively). Then, we perform a rank-$k$ approximation via truncated SVD and compute the steering vectors by averaging the rank-$k$ projections as follows:
\begin{gather*}
    \mathbf{U}_s, \mathbf{S}_s, \mathbf{V}_s = \text{rank-$k$ SVD}(\mathbf{H}_s^l)\\
    \mathbf{U}_t, \mathbf{S}_t, \mathbf{V}_t = \text{rank-$k$ SVD}(\mathbf{H}_t^l)\\
    \hat{v}^l_{s\rightarrow t} = \frac{1}{n_t}\sum_{i=1}^{n_t} [\mathbf{U}_t \mathbf{S}_t \mathbf{V}^T_t]_i - \frac{1}{n_s}\sum_{i=1}^{n_s} [\mathbf{U}_s \mathbf{S}_s \mathbf{V}^T_s]_i
\end{gather*}
where $[\mathbf{X}]_i$ denotes the $i$-th column of matrix $\mathbf{X}$. We consider ranks $k \in [1, 4, 16, 64]$ and compare with full-rank TARDIS.

\begin{figure*}[ht!]
\centering
\subfigure{\includegraphics[width=0.48\textwidth]{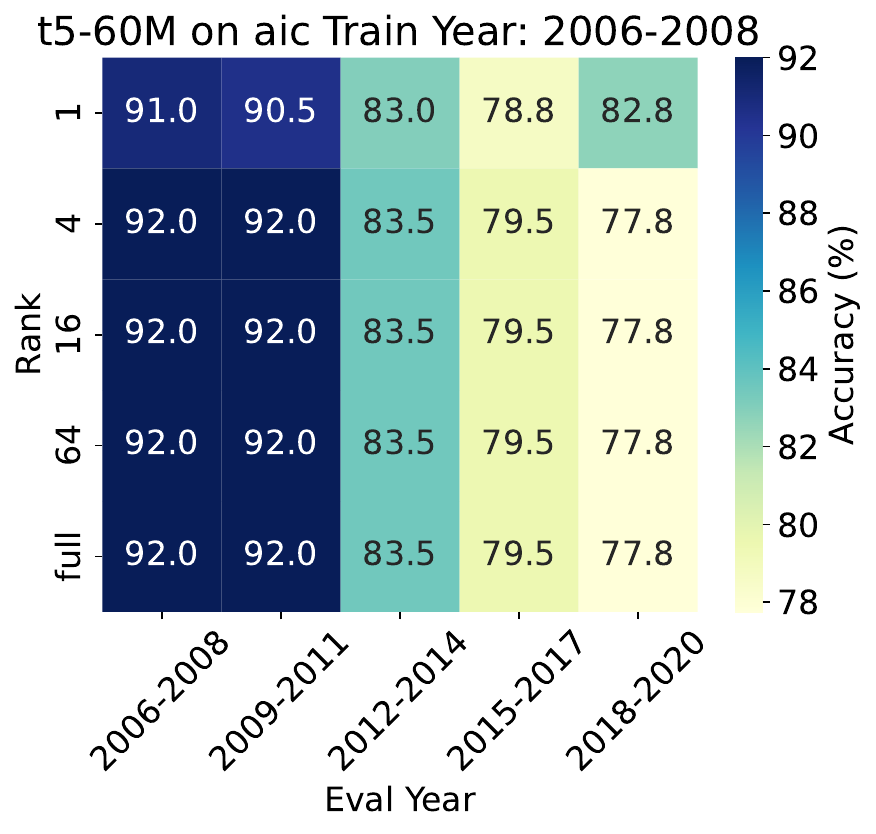}}
\subfigure{\includegraphics[width=0.48\textwidth]{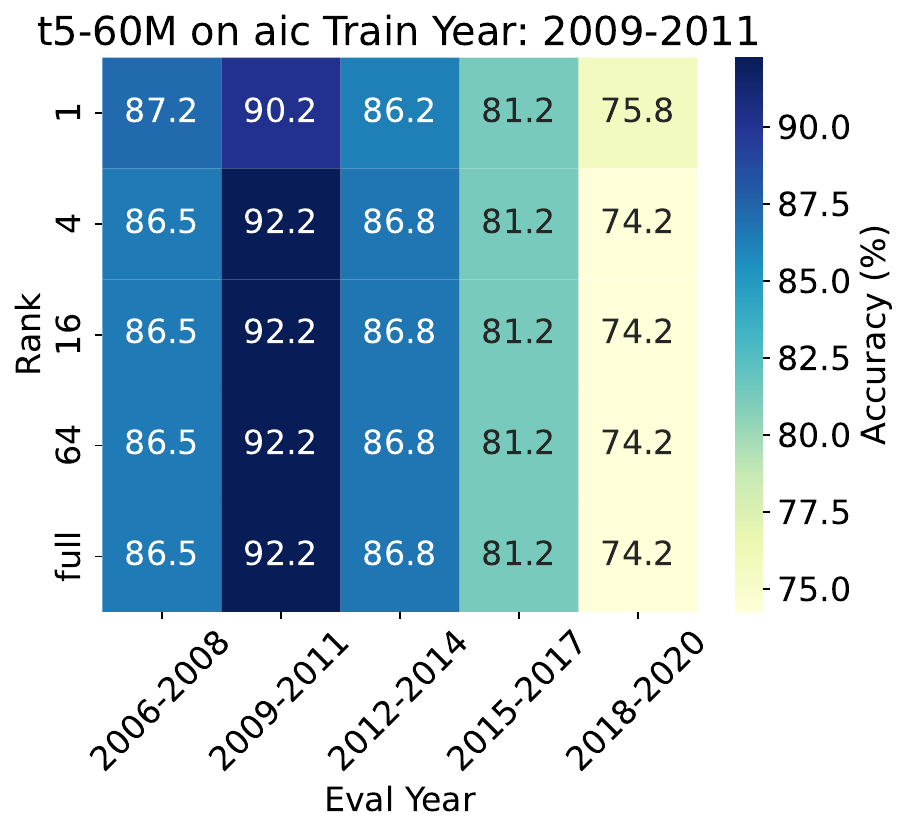}}
\subfigure{\includegraphics[width=0.48\textwidth]{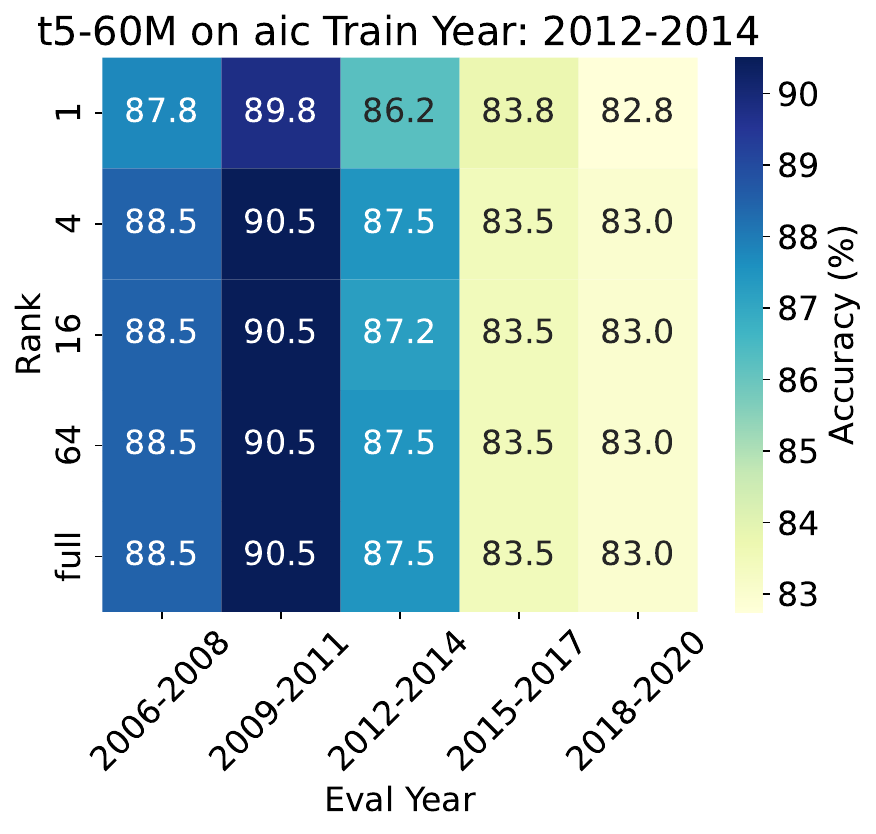}}
\subfigure{\includegraphics[width=0.48\textwidth]{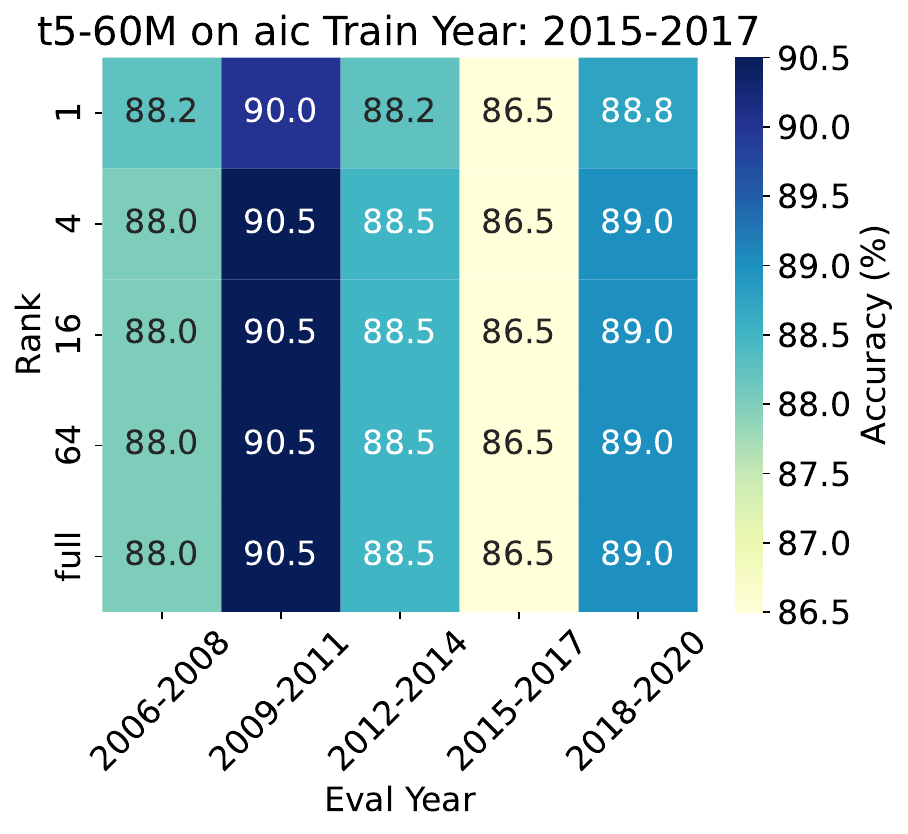}}
\subfigure{\includegraphics[width=0.48\textwidth]{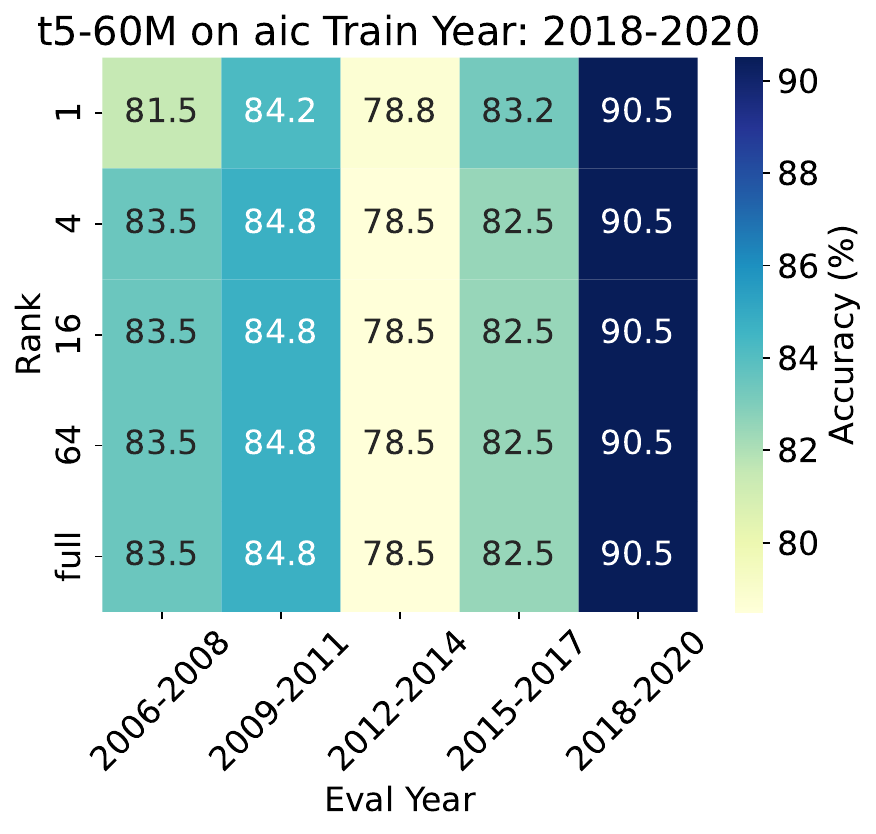}}
\caption{AIC results with or without low-rank projections with various ranks $k$.}\label{fig:rank_ablation_aic}
\end{figure*}

\begin{figure*}[ht!]
\centering
\subfigure{\includegraphics[width=0.48\textwidth]{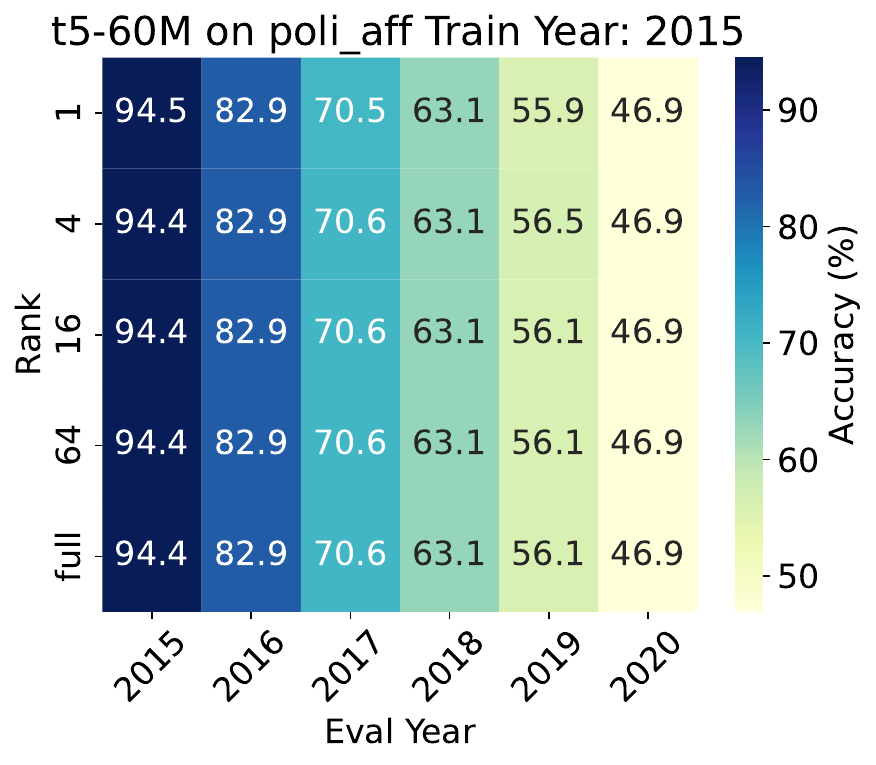}}
\subfigure{\includegraphics[width=0.48\textwidth]{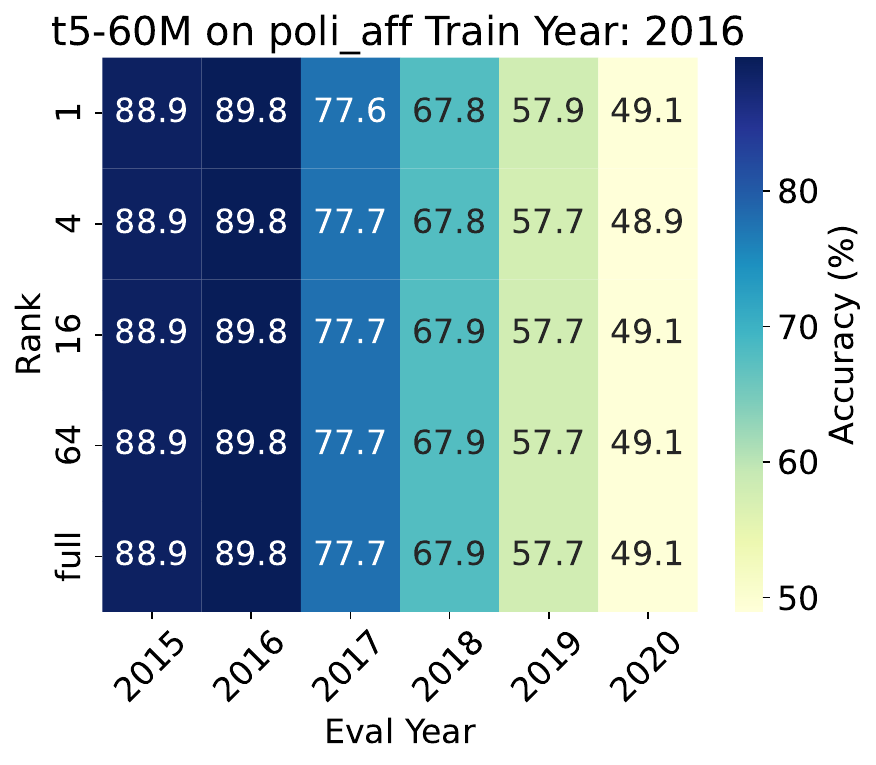}}
\subfigure{\includegraphics[width=0.48\textwidth]{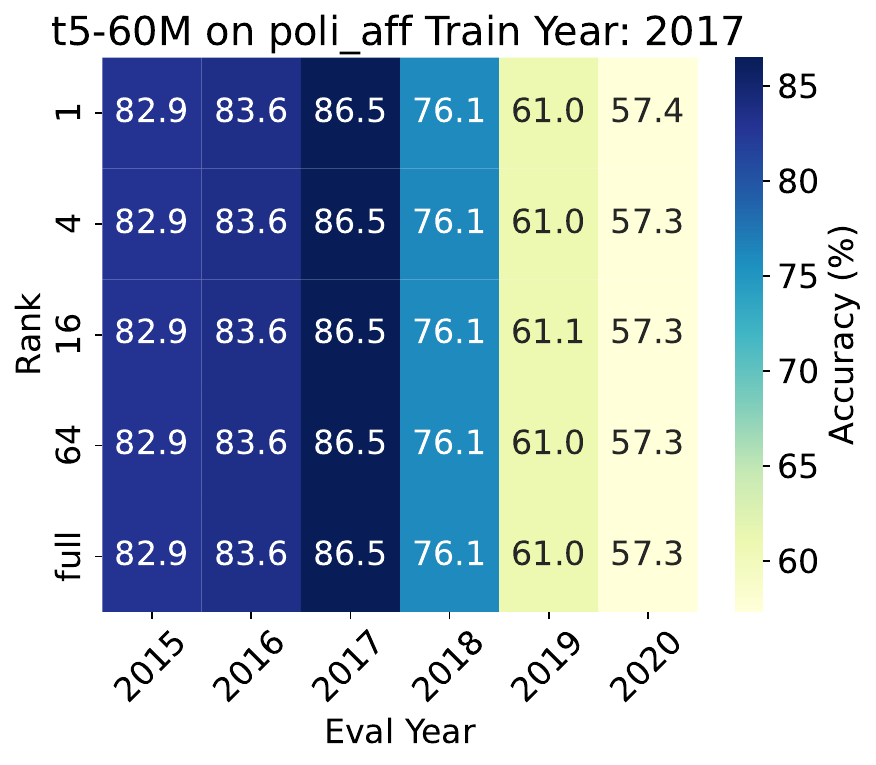}}
\subfigure{\includegraphics[width=0.48\textwidth]{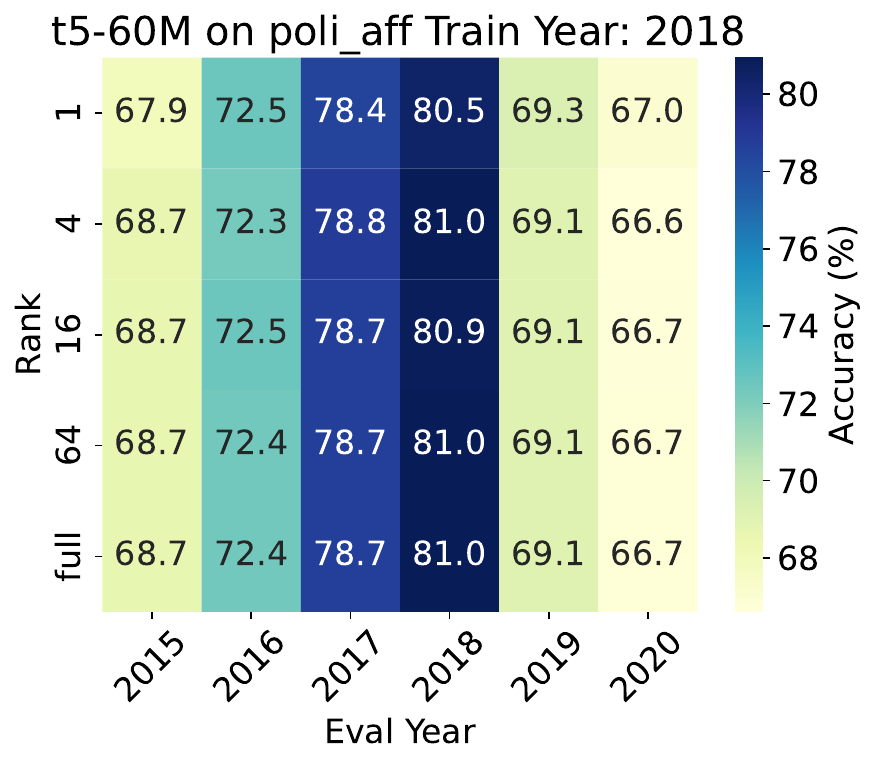}}
\subfigure{\includegraphics[width=0.48\textwidth]{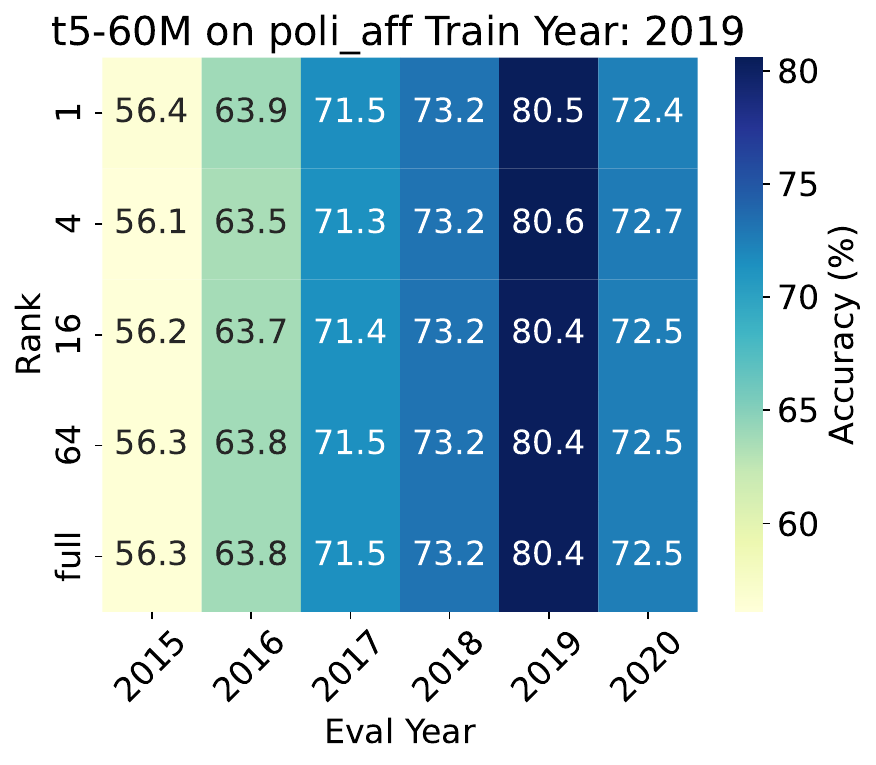}}
\subfigure{\includegraphics[width=0.48\textwidth]{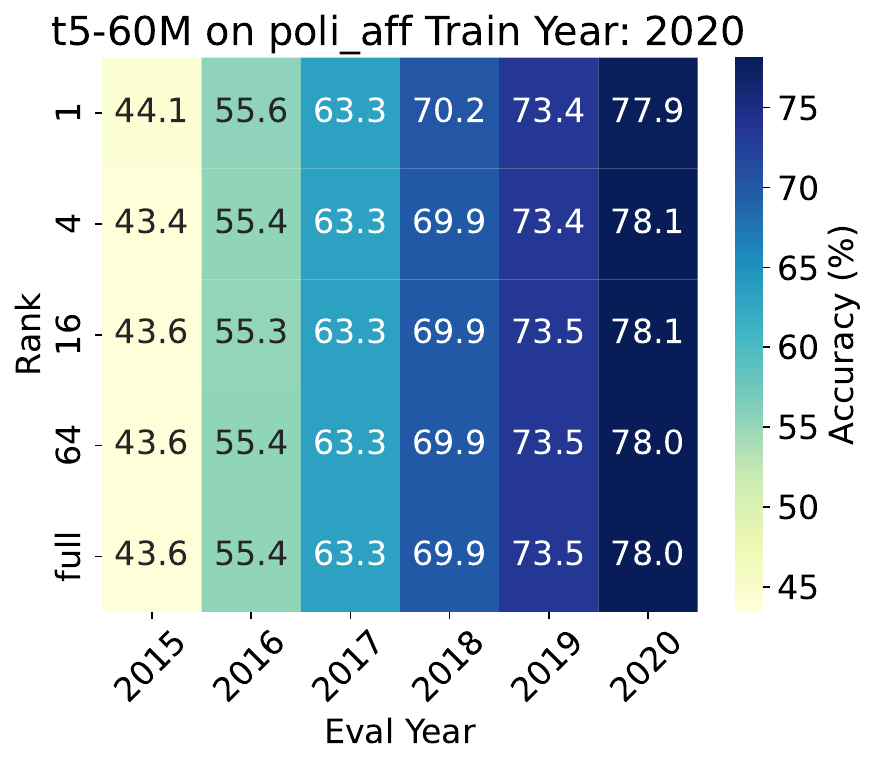}}
\caption{PoliAff results with or without low-rank projections with various ranks $k$.}\label{fig:rank_ablation_poli_aff}
\end{figure*}

\begin{figure*}[ht!]
\centering
\subfigure{\includegraphics[width=0.48\textwidth]{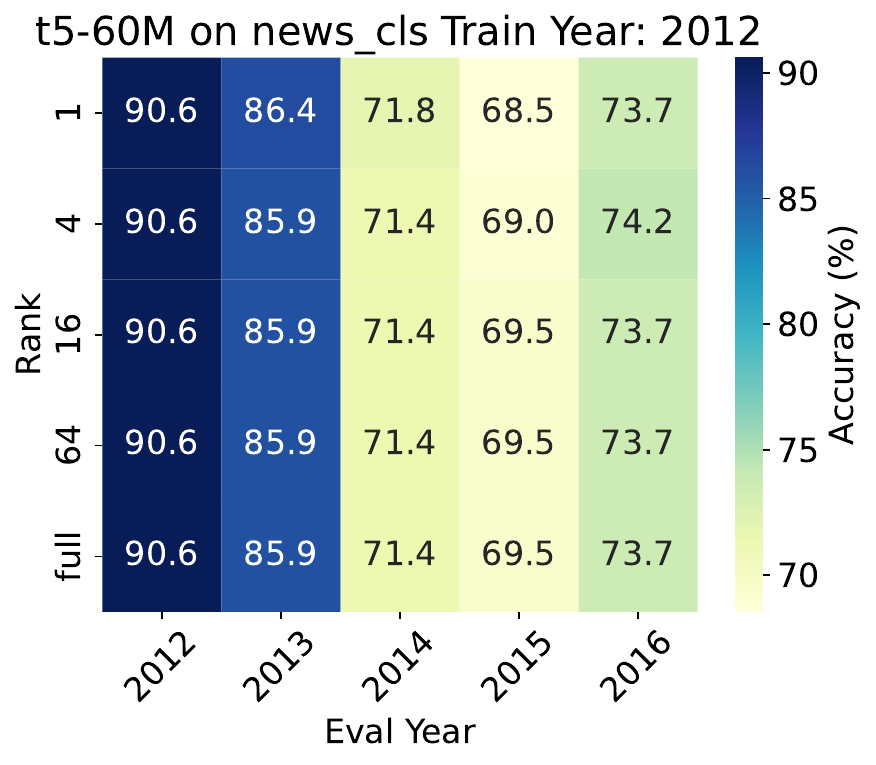}}
\subfigure{\includegraphics[width=0.48\textwidth]{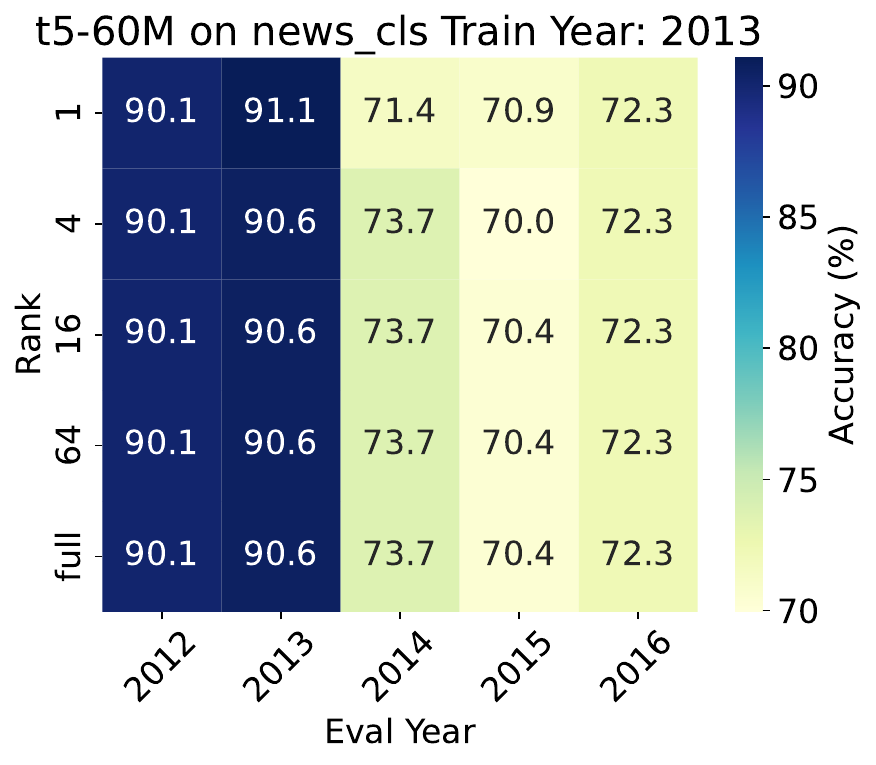}}
\subfigure{\includegraphics[width=0.48\textwidth]{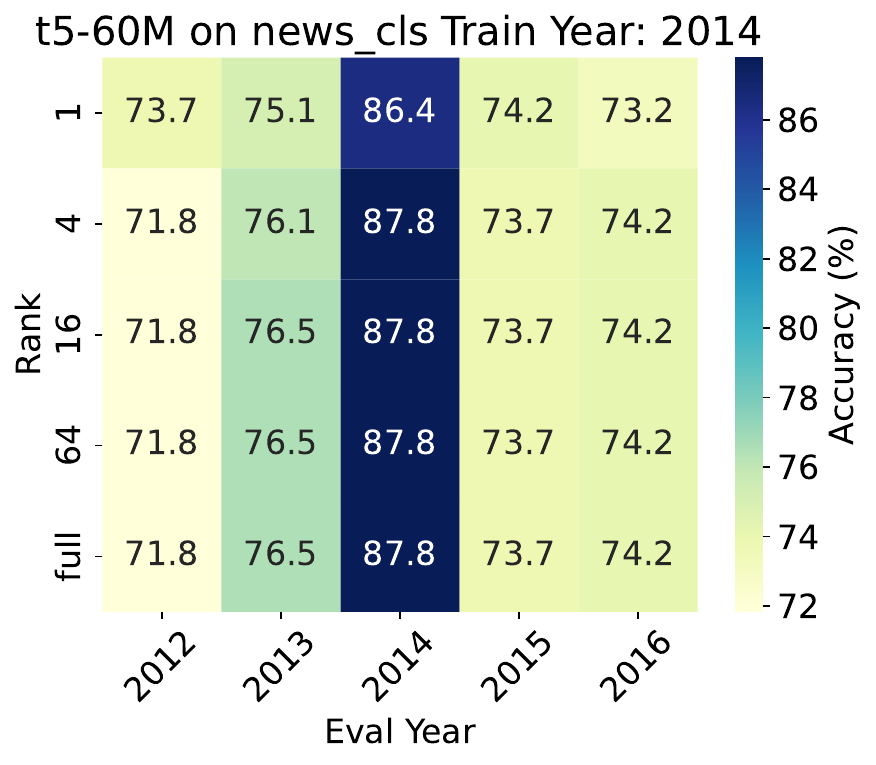}}
\subfigure{\includegraphics[width=0.48\textwidth]{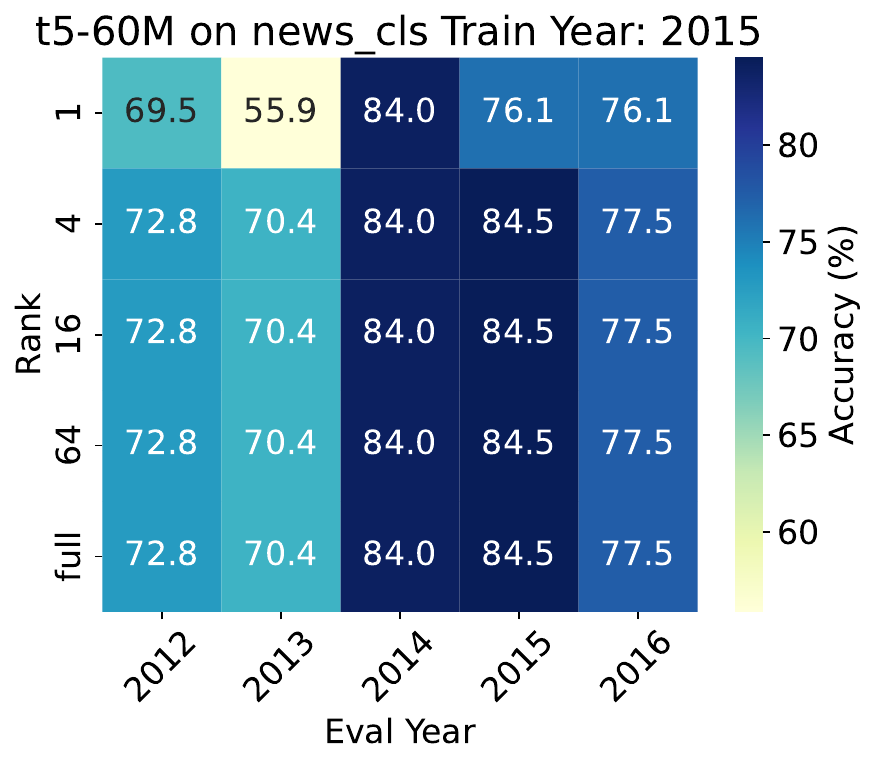}}
\subfigure{\includegraphics[width=0.48\textwidth]{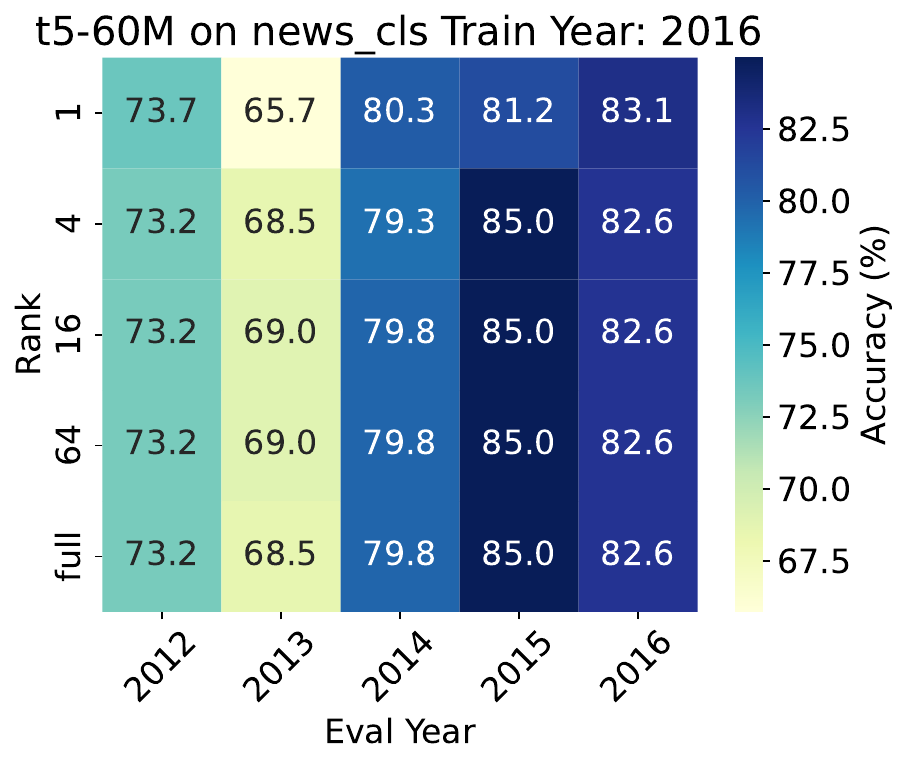}}
\caption{NewsCls results with or without low-rank projections with various ranks $k$.}\label{fig:rank_ablation_news_cls}
\end{figure*}

\paragraph{Results.} Figures~\ref{fig:rank_ablation_aic}, \ref{fig:rank_ablation_poli_aff}, \ref{fig:rank_ablation_news_cls} shows the results. In general, we find that using a rank-4 approximation is sufficient to perform comparably to full-rank TARDIS, which verifies our conjecture that temporal shifts can be fully captured in low-dimensions. More surprisingly, there exist cases (e.g., AIC 2018-2020 trained on 2006-2008, PoliAff 2018 trained on 2020, and NewsCls 2012 trained on 2014) where we see that a rank-1 approximation outperforms higher ranks, demonstrating downstream efficacy of filtering vectors into the low-rank space. Nonetheless, higher ranks lead to better predictive performances in general which also indicates a tradeoff between denoising vs. feature capacity in utilizing low-rank projections.

\clearpage

\subsection{Ablation Study on Steering Components: Attention vs. Feedforward}\label{subsec:exp_layer_ablation}

We deviate from steering representations from the final encoder layer of T5-60M and explore how different choices of the steering layer can affect performance.

\paragraph{Setup.} Instead of steering the final encoder layer as TARDIS, we test steering intermediate attention or feed-forward layers in the encoder, which leads to 12 possible layers to steer as T5-60M uses 6 encoder layers. 

\begin{figure*}[ht!]
\centering
\subfigure{\includegraphics[width=0.48\textwidth]{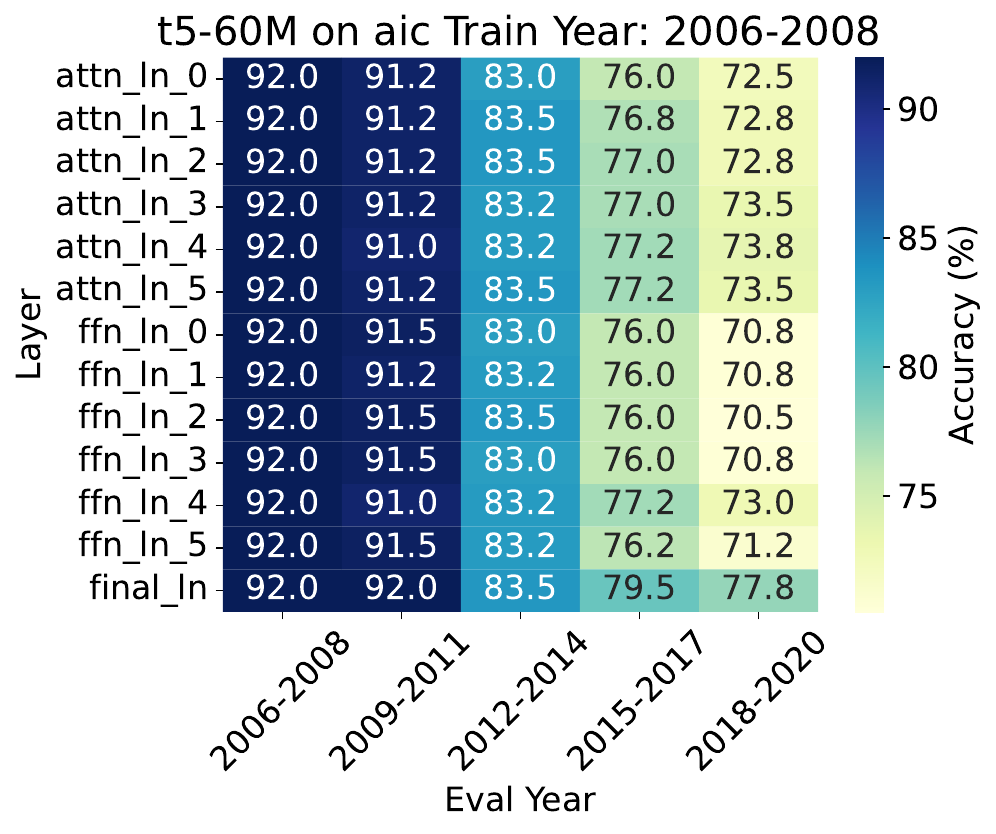}}
\subfigure{\includegraphics[width=0.48\textwidth]{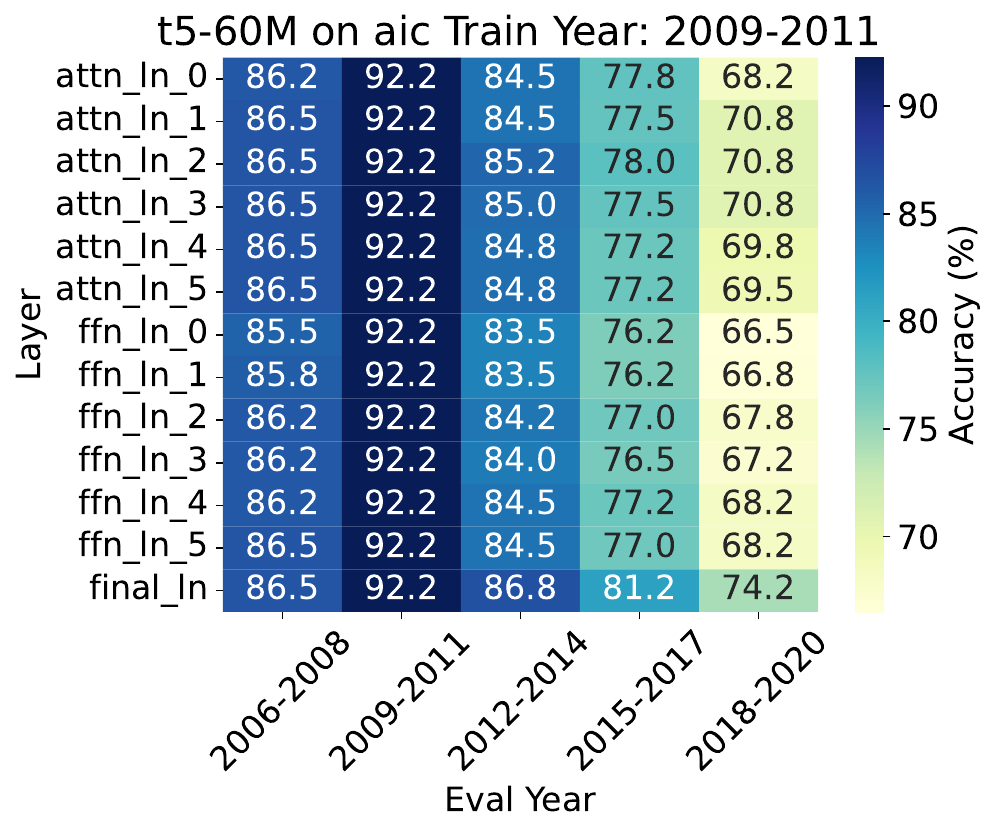}}
\subfigure{\includegraphics[width=0.48\textwidth]{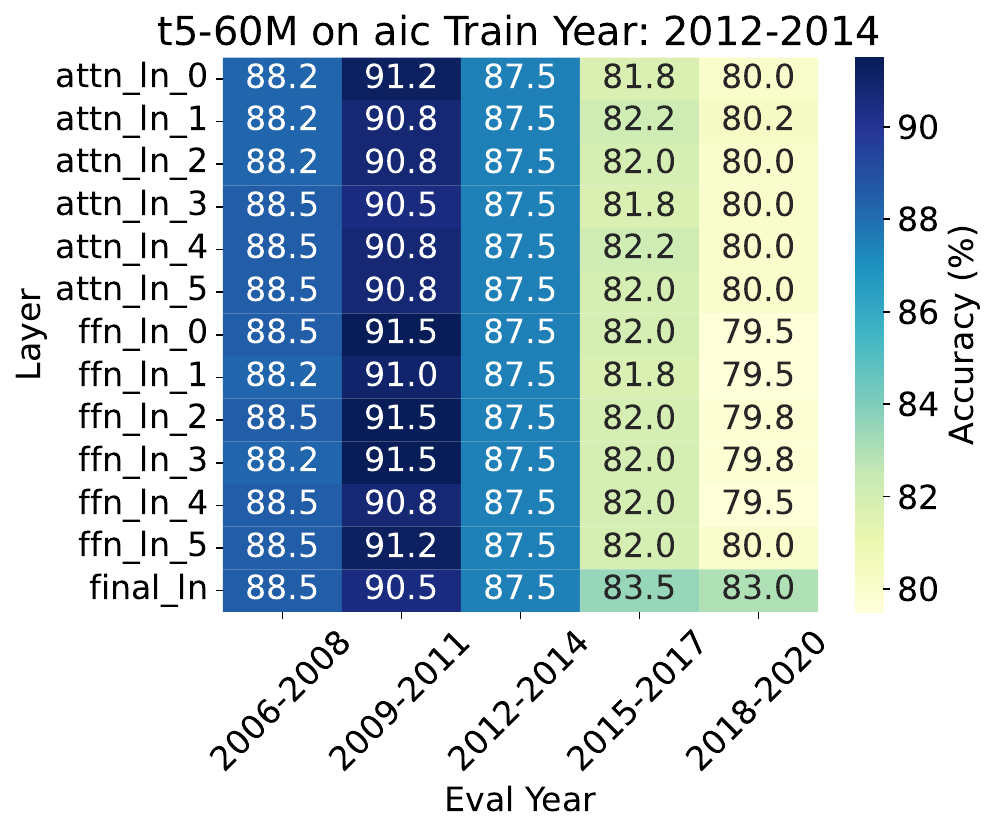}}
\subfigure{\includegraphics[width=0.48\textwidth]{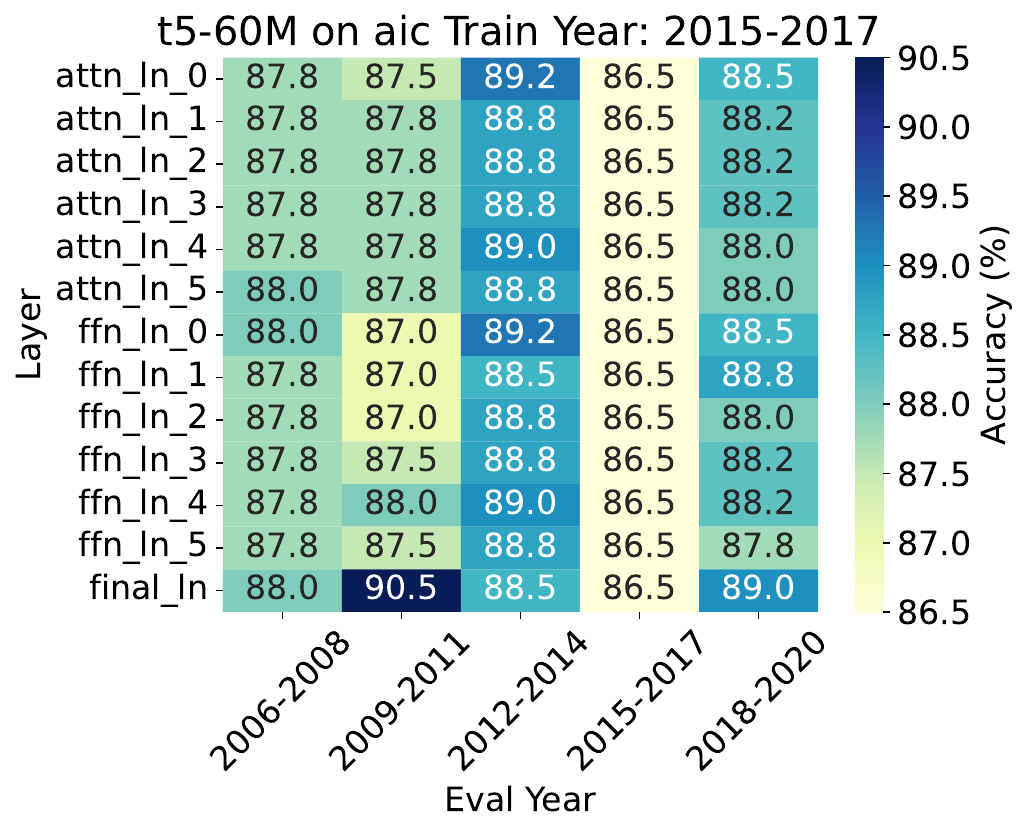}}
\subfigure{\includegraphics[width=0.48\textwidth]{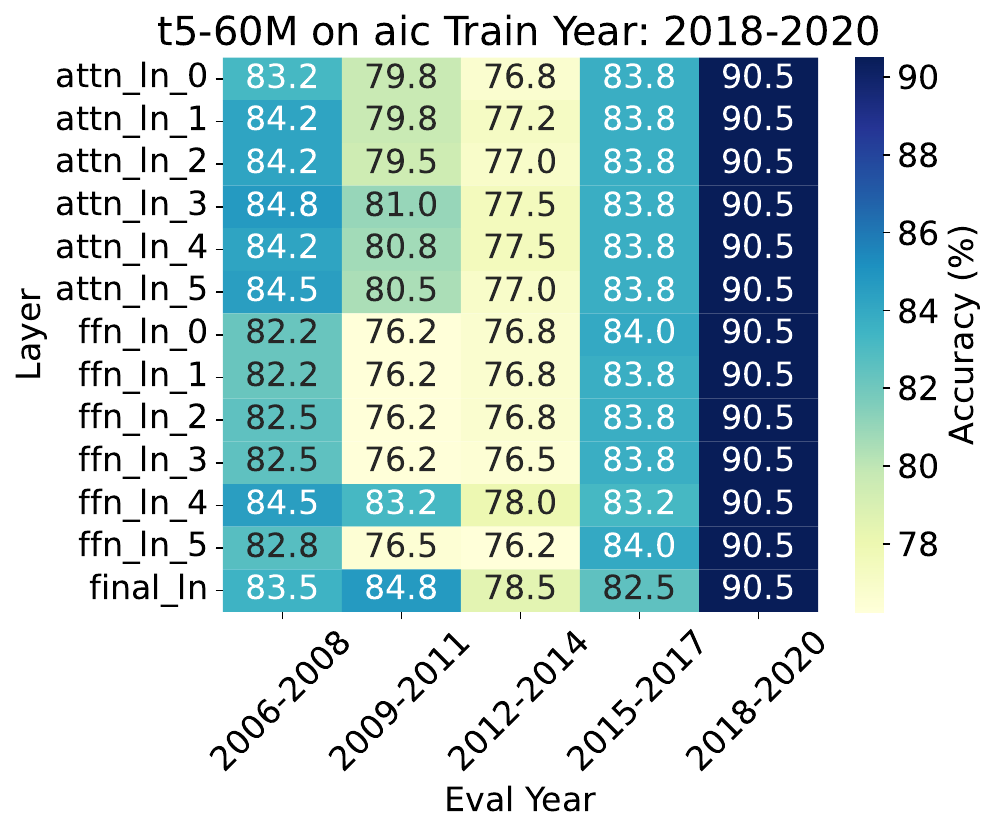}}
\caption{AIC results from using different steering layers.}\label{fig:layer_ablation_aic}
\end{figure*}

\begin{figure*}[ht!]
\centering
\subfigure{\includegraphics[width=0.48\textwidth]{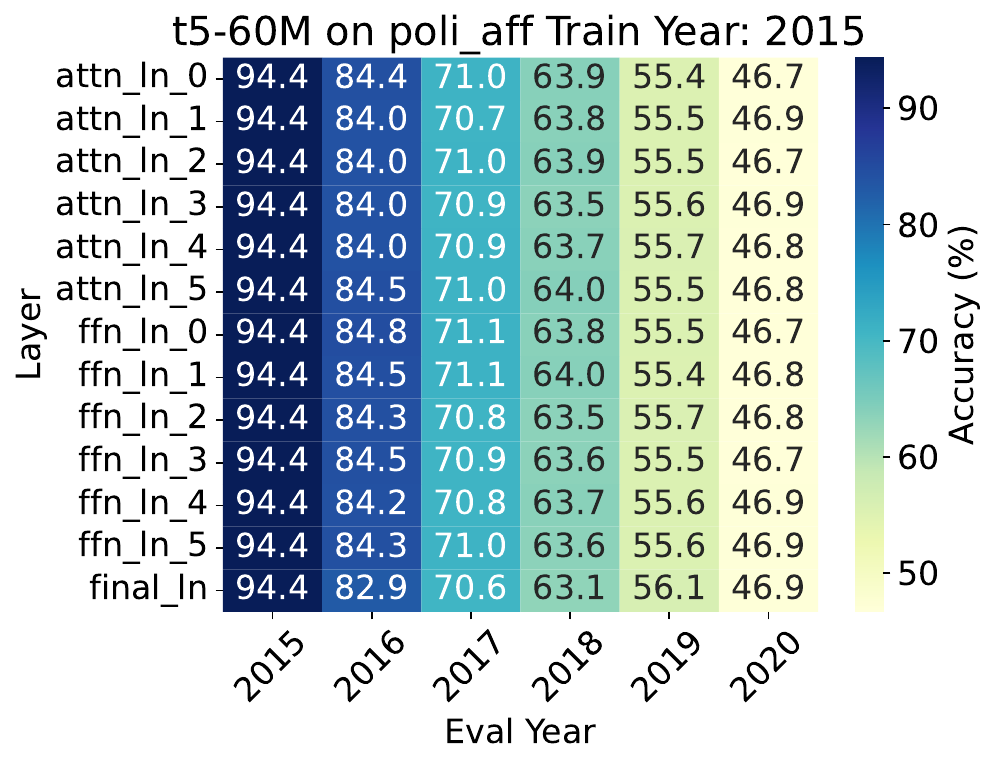}}
\subfigure{\includegraphics[width=0.48\textwidth]{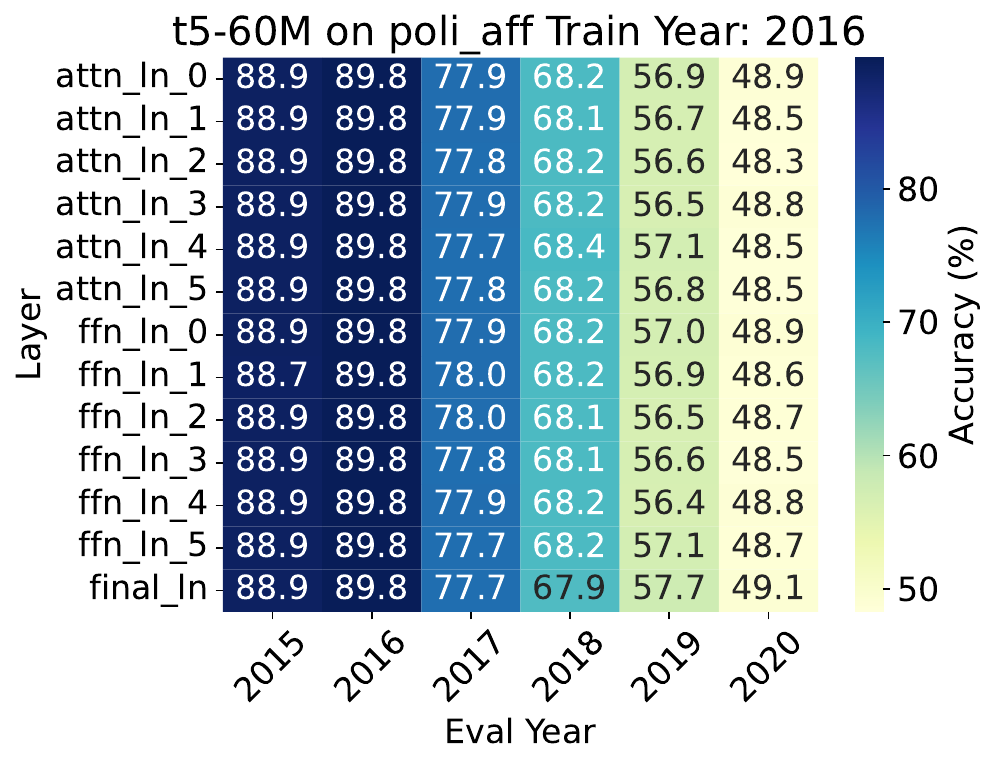}}
\subfigure{\includegraphics[width=0.48\textwidth]{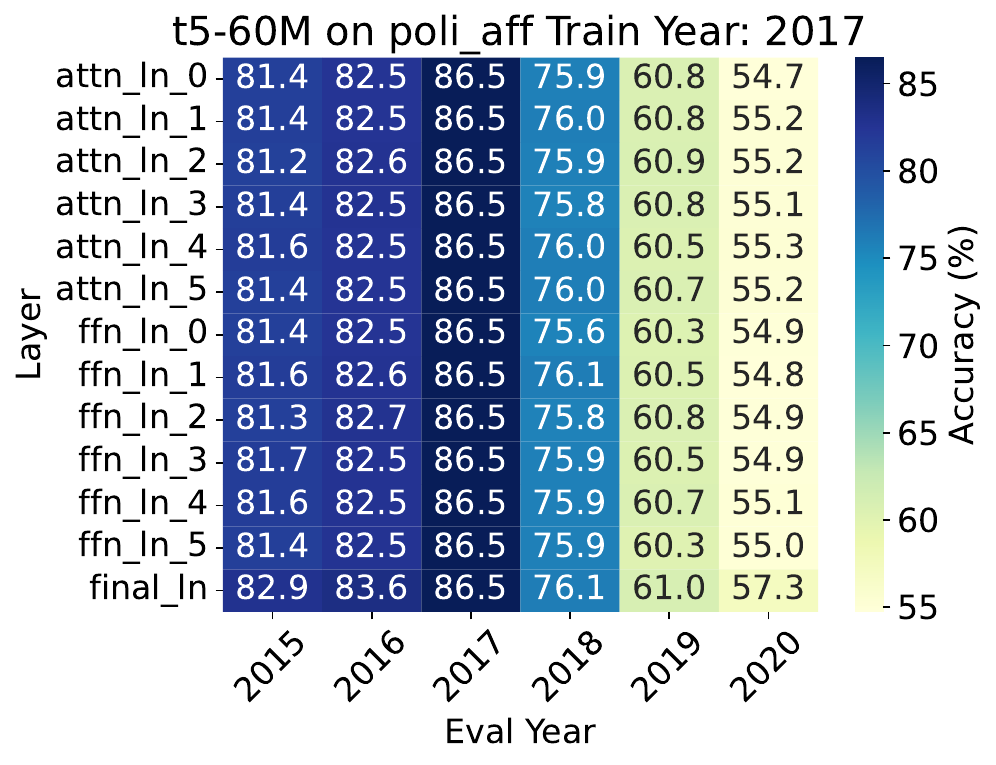}}
\subfigure{\includegraphics[width=0.48\textwidth]{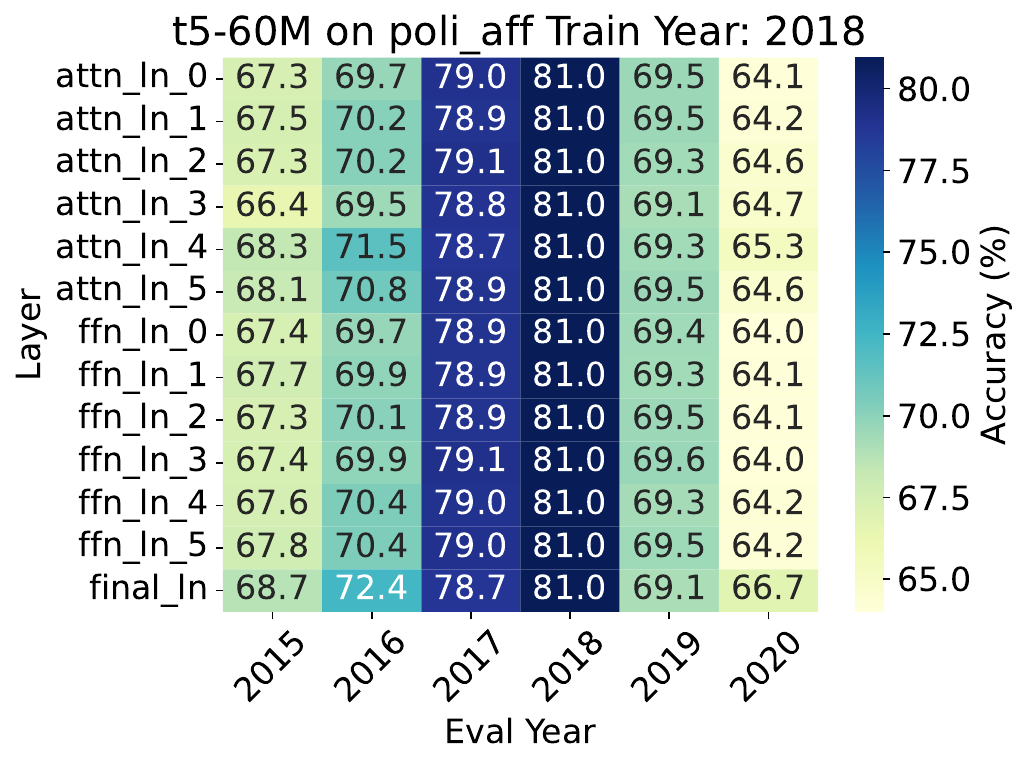}}
\subfigure{\includegraphics[width=0.48\textwidth]{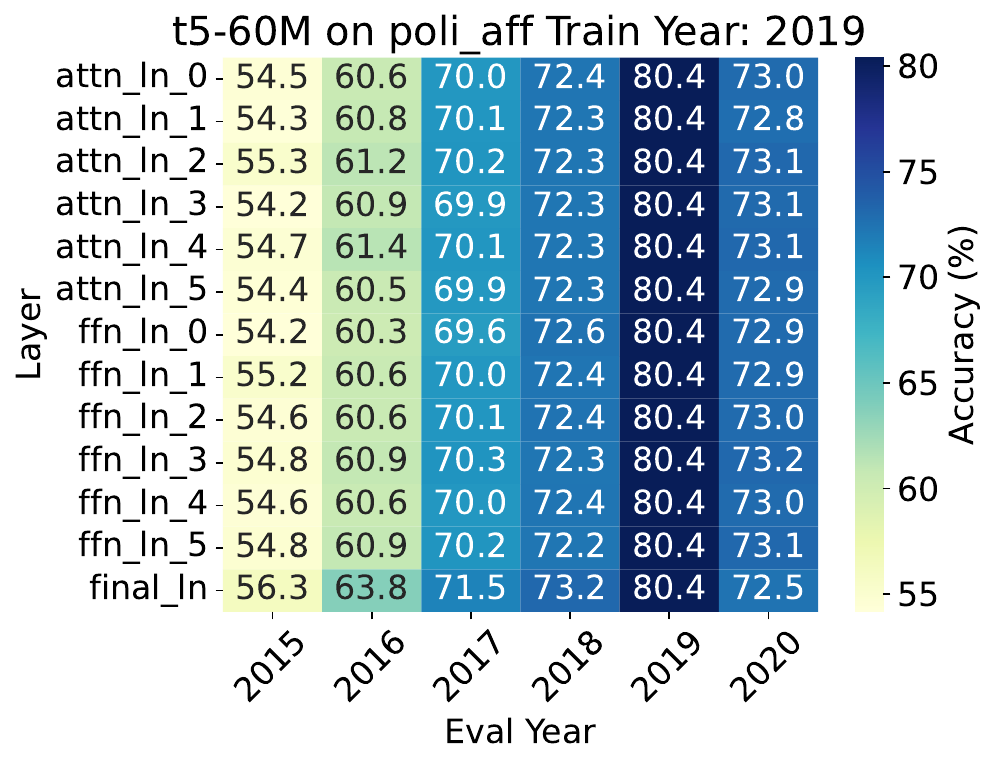}}
\subfigure{\includegraphics[width=0.48\textwidth]{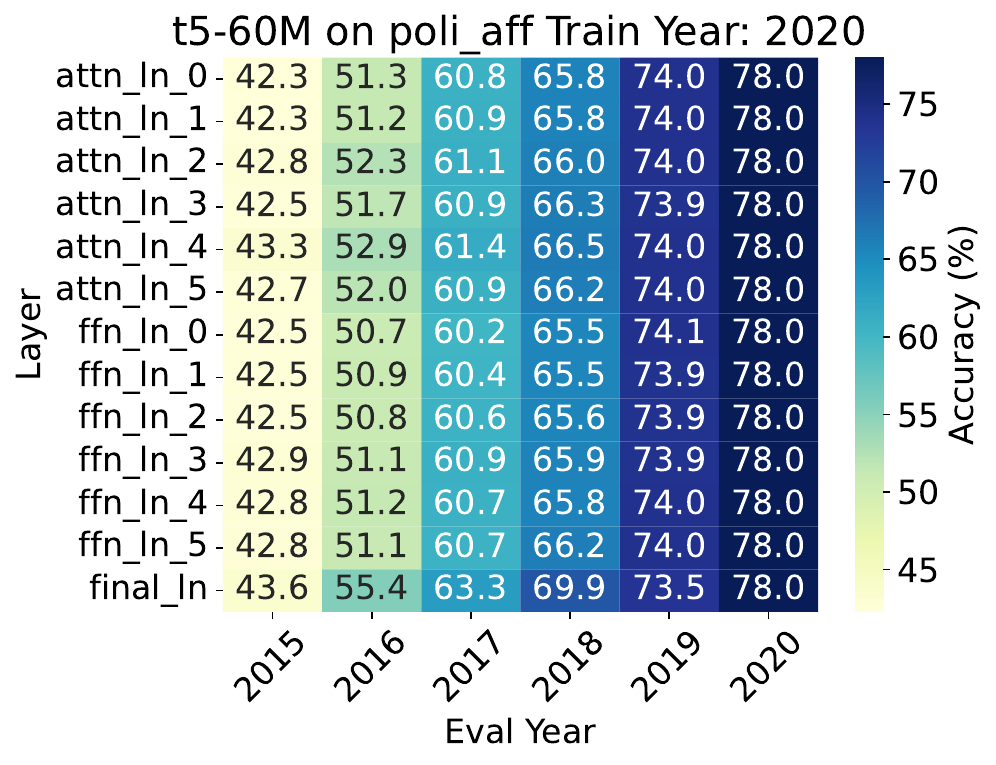}}
\caption{PoliAff results from using different steering layers.}\label{fig:layer_ablation_poli_aff}
\end{figure*}

\begin{figure*}[ht!]
\centering
\subfigure{\includegraphics[width=0.48\textwidth]{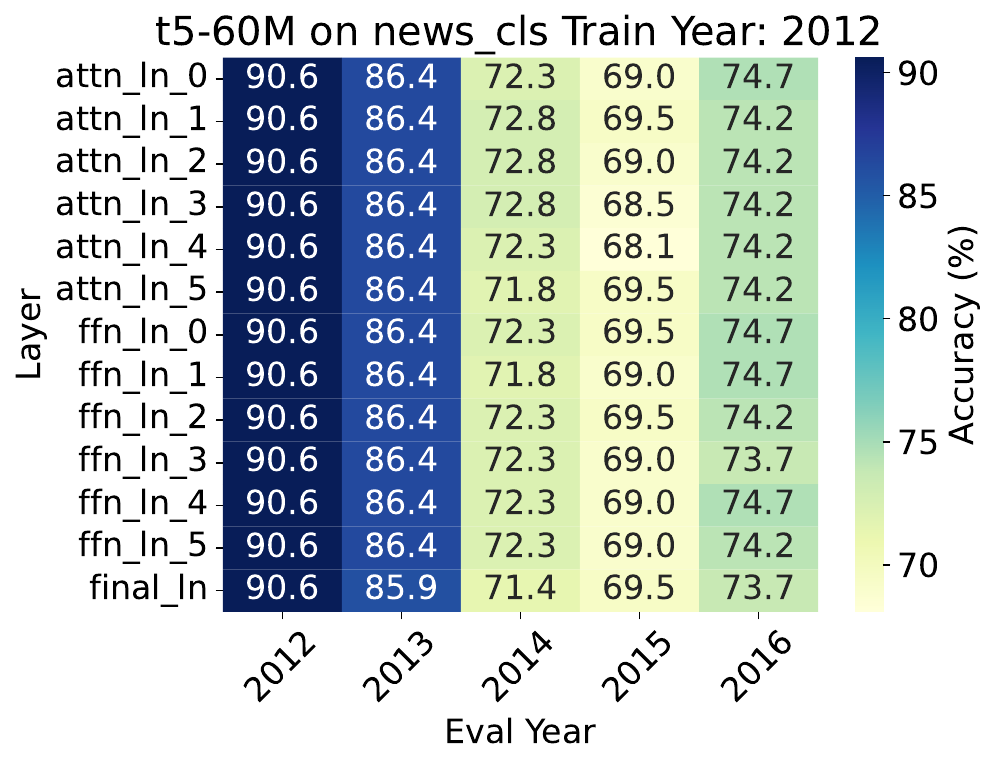}}
\subfigure{\includegraphics[width=0.48\textwidth]{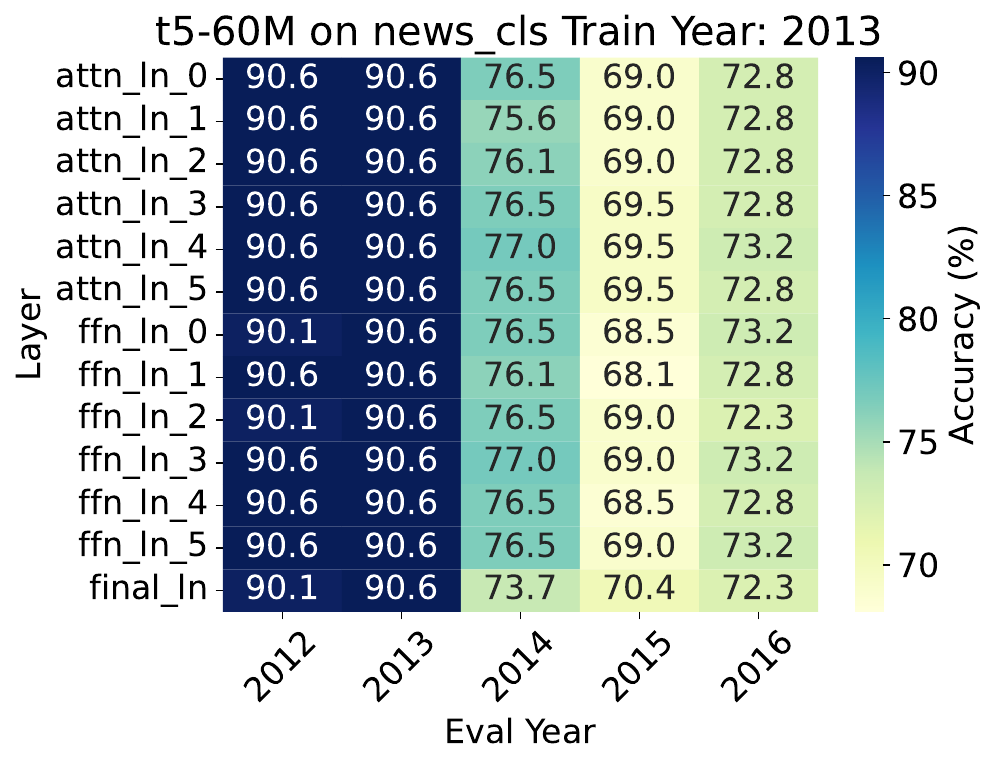}}
\subfigure{\includegraphics[width=0.48\textwidth]{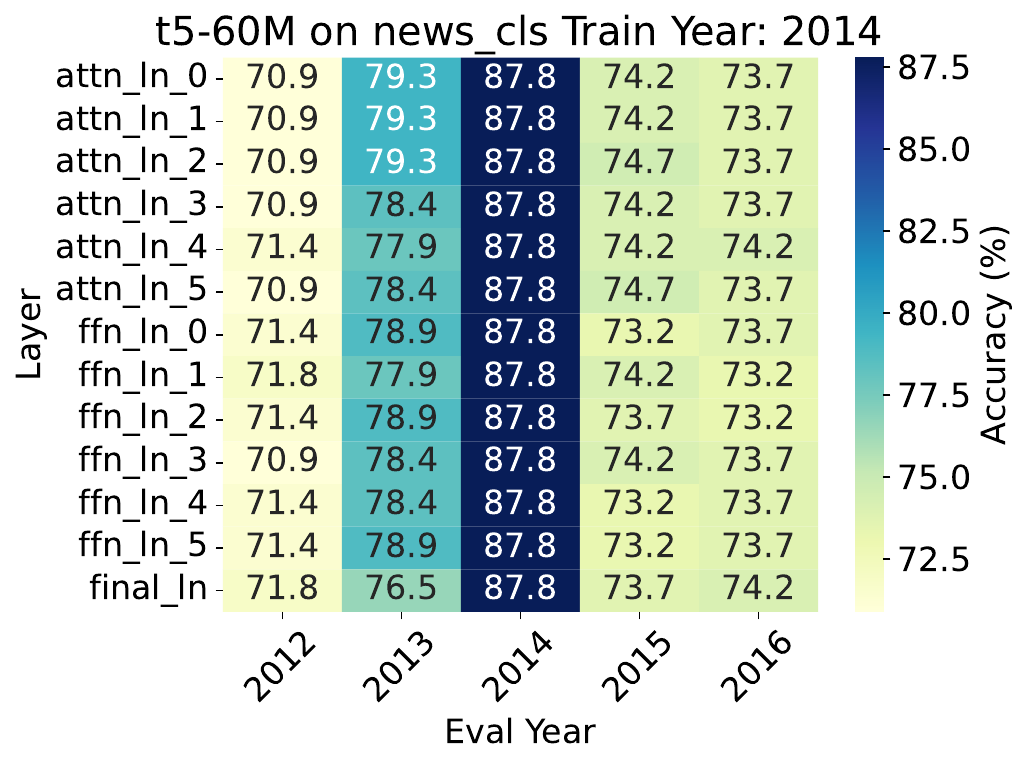}}
\subfigure{\includegraphics[width=0.48\textwidth]{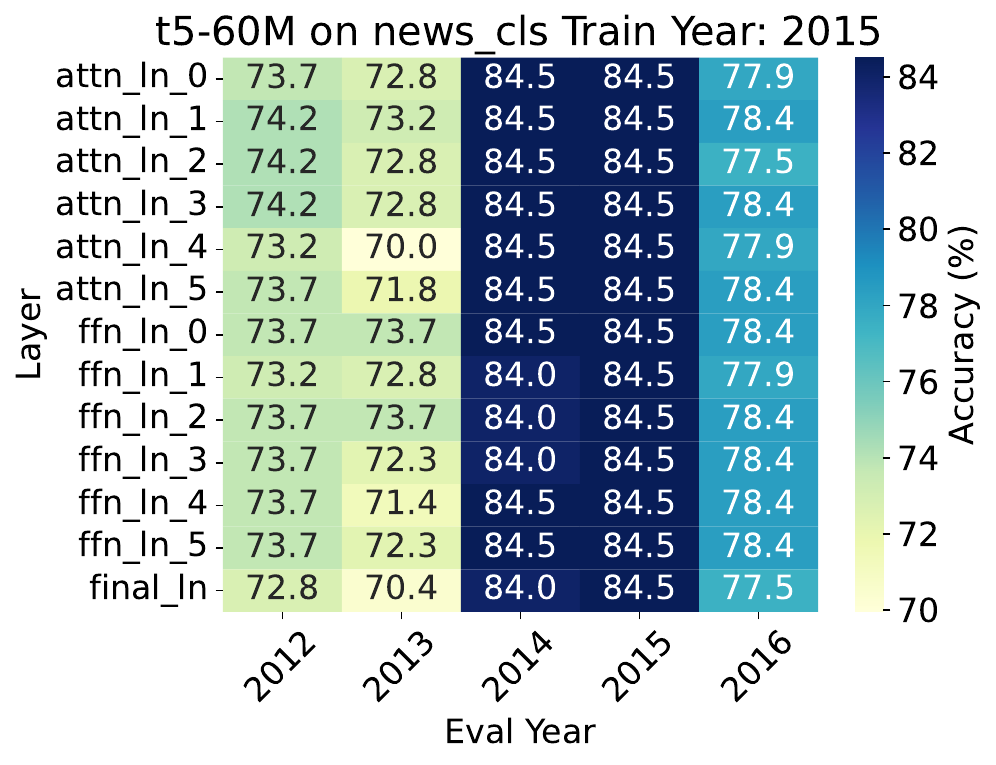}}
\subfigure{\includegraphics[width=0.48\textwidth]{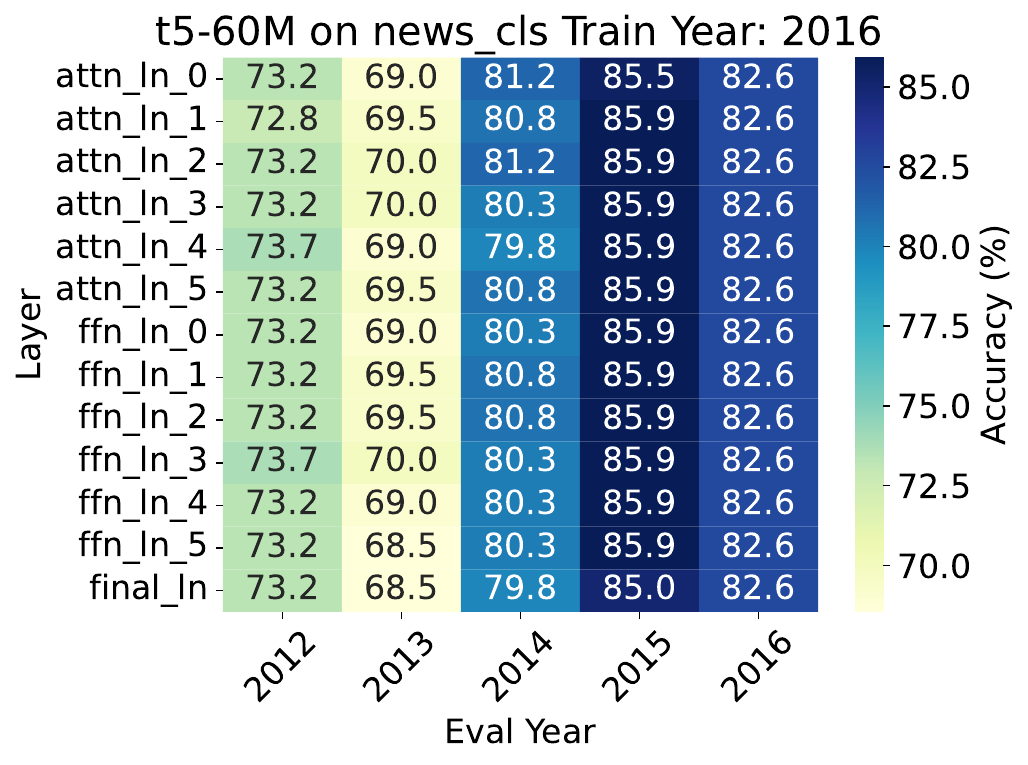}}
\caption{NewsCls results from using different steering layers.}\label{fig:layer_ablation_news_cls}
\end{figure*}

\paragraph{Results.} Figures \ref{fig:layer_ablation_aic}, \ref{fig:layer_ablation_poli_aff}, \ref{fig:layer_ablation_news_cls} in share the full results from steering different encoder layers. For AIC and PoliAff, steering the post-layernorm activations from the final layer consistently leads to best results. For NewsCls, however, we often find cases where steering earlier layers lead to better performance. This indicates that the utility of temporal features residing in each layer can depend upon the task at hand: AIC and PoliAff requires higher-level steering after sufficient mixing of information across tokens whereas NewsCls works better with low-level steering. For future work, we hope to explore methods that can adaptively choose the steering layer based upon the task and data at hand.

\clearpage
\subsection{Ablation Study on Steering Layer: Encoder vs. Decoder vs. Both}\label{subsec:exp_layer_ablation_encoder_decoder}

In our main experiments, we applied steering to both the encoder and decoder layers of the encoder-decoder model (T5). To investigate which layer has a more significant effect on steering performance, we conducted an ablation study by varying the intervention layer.

\paragraph{Setup.}
We followed the same experimental setup as before but additionally examined the effects of steering the encoder and decoder layers separately. Specifically, we applied interventions to the last feedforward output of the encoder, the last feedforward output of the decoder, and both, then observed how the results varied under each condition.

\begin{figure*}
    \centering
    \includegraphics[width=.95\textwidth]{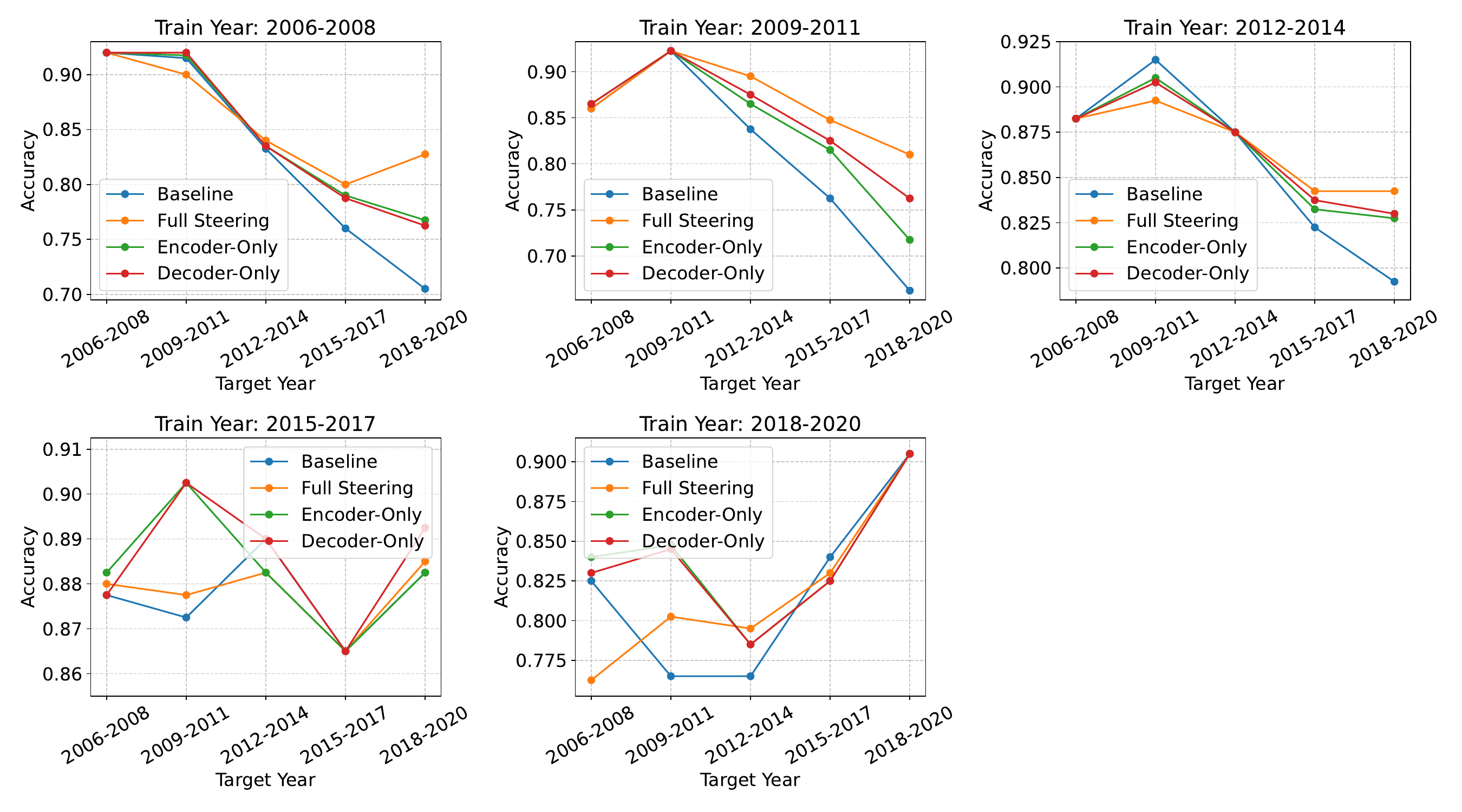}
    \caption{Intervention Layer Ablation: AIC}
    \label{fig:aic_intervention_layer_ablation}
\end{figure*}

\begin{figure*}
    \centering
    \includegraphics[width=.95\textwidth]{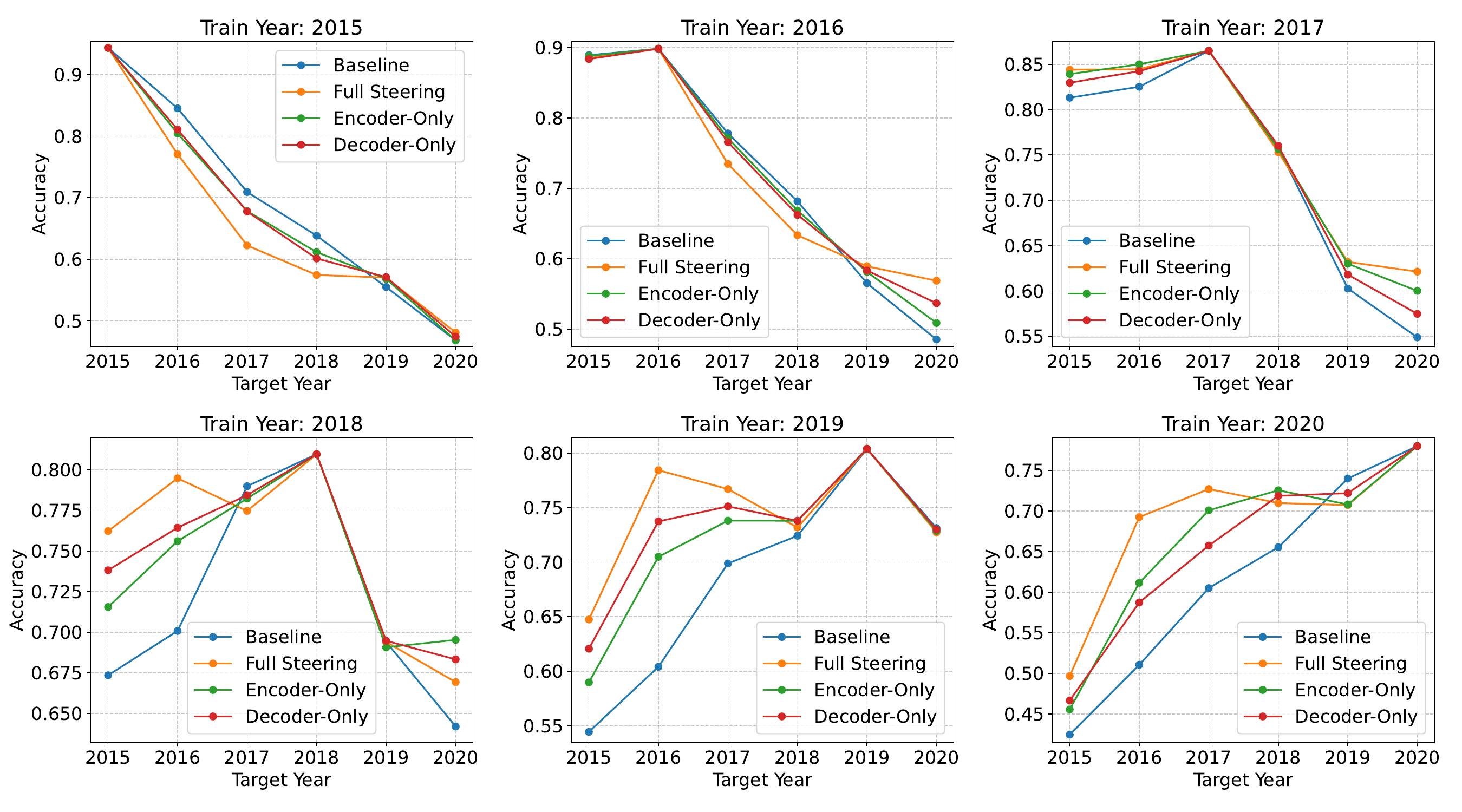}
    \caption{Intervention Layer Ablation: PoliAff}
    \label{fig:poliaff_intervention_layer_ablation}
\end{figure*}

\begin{figure*}
    \centering
    \includegraphics[width=.95\textwidth]{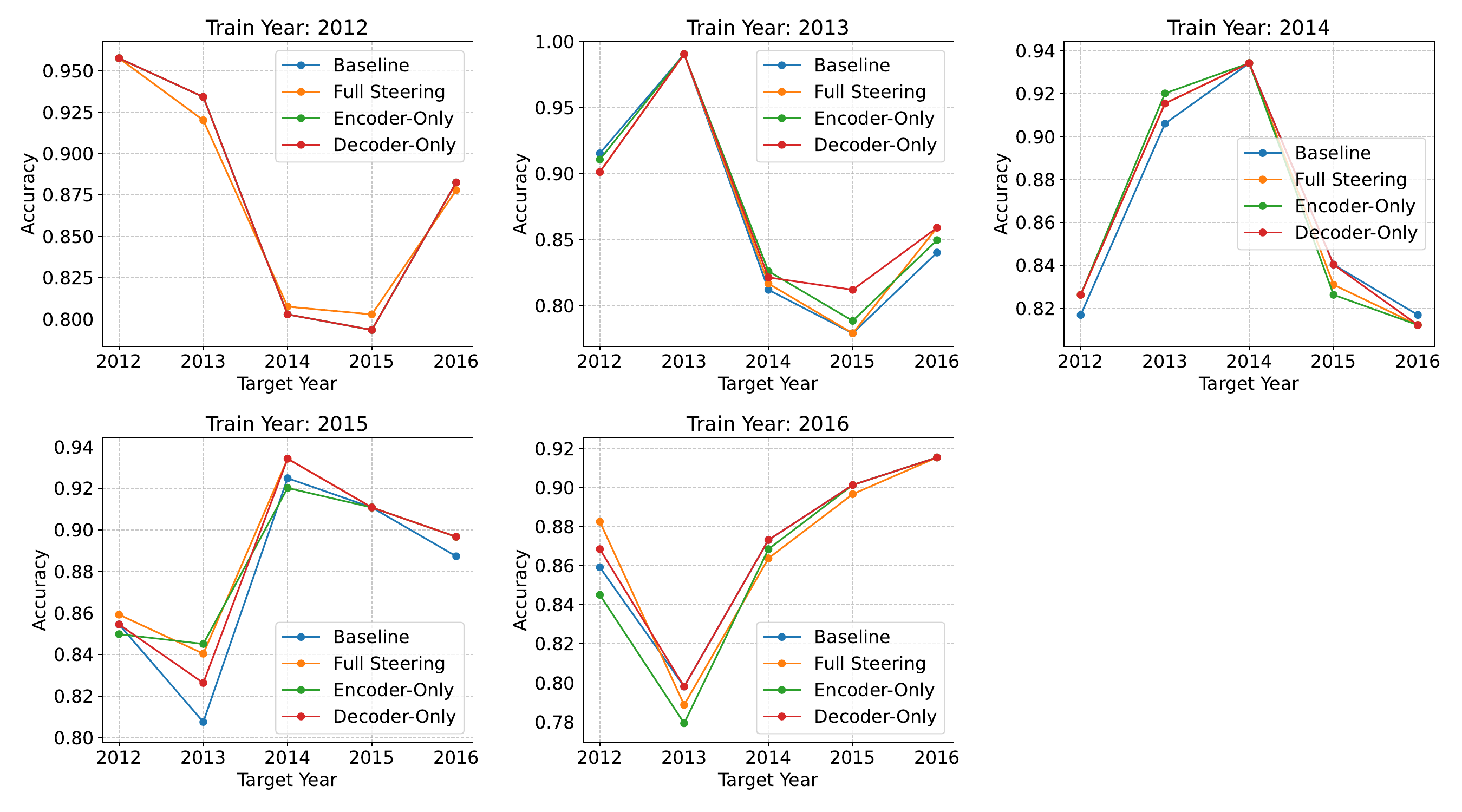}
    \caption{Intervention Layer Ablation: NewsCls}
    \label{fig:news_intervention_layer_ablation}
\end{figure*}

\paragraph{Results.} Figures \ref{fig:aic_intervention_layer_ablation}, \ref{fig:poliaff_intervention_layer_ablation}, \ref{fig:news_intervention_layer_ablation} present the experimental results. While the results show some inconsistency, intervening in both the encoder and decoder layers generally yields the best performance across conditions.

\clearpage

\subsection{Ablation on Data Size for Steering}
We hypothesize that representation steering can be effective with a small number of data points, demonstrating data efficiency.

\paragraph{Setup.}
We follow the same experimental setup as in Section \ref{subsec:exp1}, except that we vary the dataset size by randomly selecting subsets of the test set to compute the steering vectors. We repeat this process 10 times using different random seeds and report the mean and standard deviation. Note that the test set used for evaluation remains unchanged.

\begin{figure*}
    \centering
    \includegraphics[width=.95\textwidth]{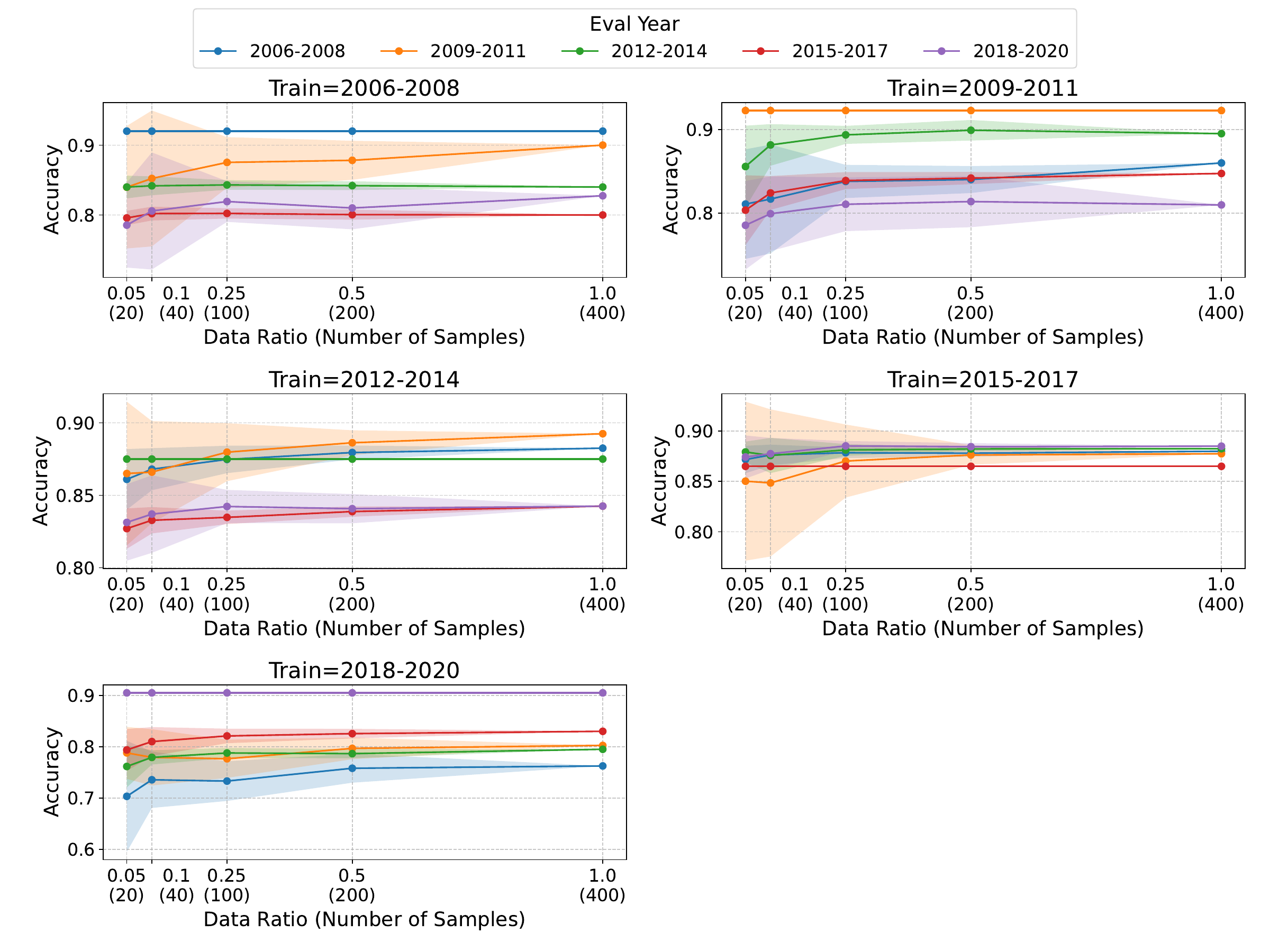}
    \caption{Ablation on Data Size for Steering: AIC}
    \label{fig:aic_data_size_ablation}
\end{figure*}

\begin{figure*}
    \centering
    \includegraphics[width=.95\textwidth]{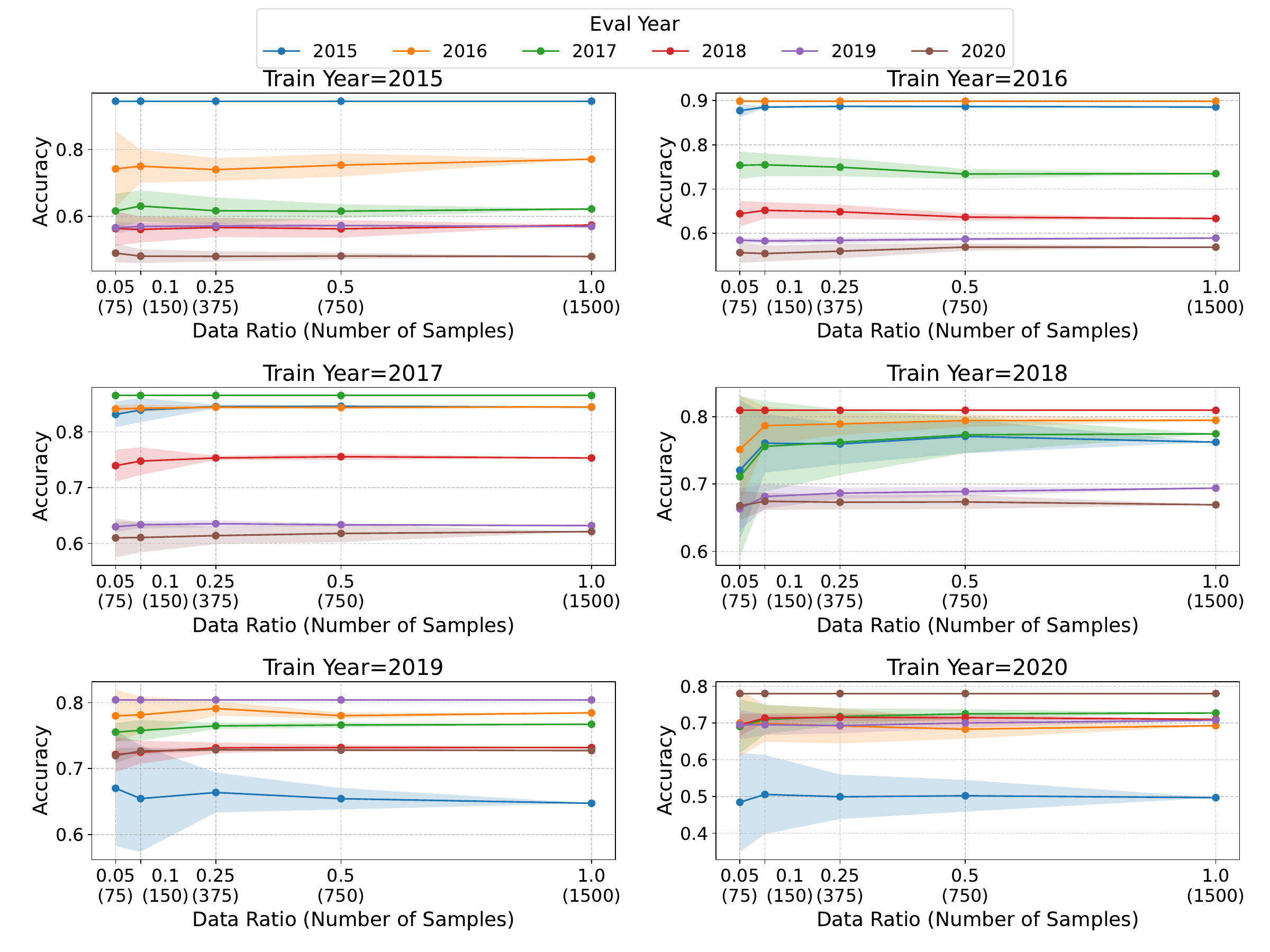}
    \caption{Ablation on Data Size for Steering: PoliAff}
    \label{fig:poliaff_data_size_ablation}
\end{figure*}

\begin{figure*}
    \centering
    \includegraphics[width=.95\textwidth]{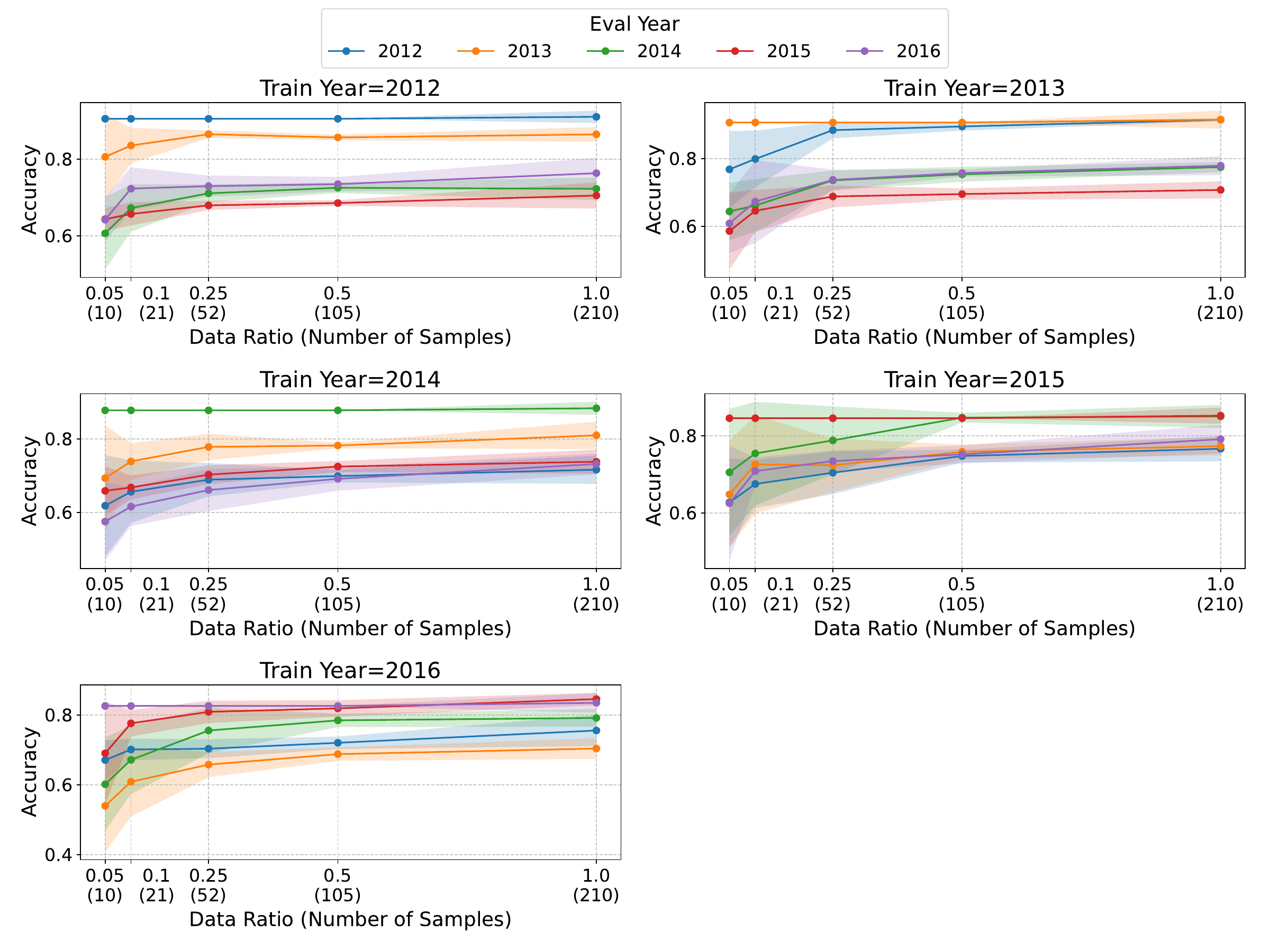}
    \caption{Ablation on Data Size for Steering: NewsCls}
    \label{fig:news_data_size_ablation}
\end{figure*}

\paragraph{Results.}
Figure \ref{fig:aic_data_size_ablation}, \ref{fig:poliaff_data_size_ablation}, \ref{fig:news_data_size_ablation} presents the results. As expected, even a small subset of data (100–200 samples) is sufficient to achieve effective steering across datasets.

\clearpage

\end{document}